\documentclass[final,12pt]{clear2025} 

\usepackage[utf8]{inputenc} 
\usepackage[T1]{fontenc}    
\usepackage{hyperref}       
\usepackage{url}            
\usepackage{booktabs}       
\usepackage{amsfonts}       
\usepackage{nicefrac}       
\usepackage{microtype}      
\usepackage{lipsum}
\usepackage{fancyhdr}       
\usepackage{graphicx}       

\usepackage{multirow}
\usepackage{subcaption}
\usepackage{caption}

\usepackage{array}
\graphicspath{{media/}}     
\usepackage{mathtools}

\usepackage{booktabs}
\usepackage{bbding}
\usepackage{pifont}
\newcommand{\cmark}{\textcolor{teal}{\ding{51}}}%
\newcommand{\xmark}{\textcolor{purple}{\ding{55}}}%

\usepackage{tikz}
\def\halfcheckmark{\tikz\draw[scale=0.3,gray,fill=gray](0,.35) -- (.25,0) -- (1,.7) -- (.25,.15) -- cycle (0.75,0.2) -- (0.77,0.2)  -- (0.6,0.7) -- cycle;}

\usepackage{esvect}

\usepackage{algorithm}

\usepackage{enumitem}
\renewcommand{\labelenumi}{\arabic{enumi}.}

\title[MXMap]{MXMap: A Multivariate Cross Mapping Framework for Causal Discovery in Dynamical Systems}
\usepackage{times}

\clearauthor{%
 \Name{Elise Zhang} \Email{elise.zhang@mail.mcgill.ca}\\
 \addr  McGill University, Montréal, QC Canada
 \AND
 \Name{François Mirallès} \Email{miralles.francois@hydroquebec.com}\\
 \addr Hydro-Québec Research Institute (IREQ), Varennes, QC Canada
 \AND
 \Name{Raphaël Rousseau-Rizzi} \Email{rousseau-rizzi.raphael@hydroquebec.com}\\
 \addr Hydro-Québec Research Institute (IREQ), Varennes, QC Canada
 \AND
 \Name{Di Wu} \Email{di.wu5@mcgill.ca}\\
 \addr McGill University, Montréal, QC Canada
 \AND
 \Name{Arnaud Zinflou} \Email{zinflou.arnaud@hydroquebec.com}\\
 \addr Hydro-Québec Research Institute (IREQ), Varennes, QC Canada
 \AND
 \Name{Benoit Boulet} \Email{benoit.boulet@mcgill.ca}\\
 \addr McGill University, Montréal, QC Canada
}

\begin{document}

\maketitle

\begin{abstract}%

Convergent Cross Mapping (CCM) is a powerful method for detecting causality in coupled nonlinear dynamical systems, providing a model-free approach to capture dynamic causal interactions. Partial Cross Mapping (PCM) was introduced as an extension of CCM to address indirect causality in three-variable systems by comparing cross-mapping quality between direct cause-effect mapping and indirect mapping through an intermediate conditioning variable. However, PCM remains limited to univariate delay embeddings in its cross-mapping processes. In this work, we extend PCM to the multivariate setting, introducing multiPCM, which leverages multivariate embeddings to more effectively distinguish indirect causal relationships. We further propose a multivariate cross-mapping framework (MXMap) for causal discovery in dynamical systems. This two-phase framework combines (1) pairwise CCM tests to establish an initial causal graph and (2) multiPCM to refine the graph by pruning indirect causal connections. Through experiments
\footnote{Implementation at \url{https://github.com/elisejiuqizhang/multiPCM}.}
on simulated data and the \textit{ERA5 Reanalysis} weather dataset, we demonstrate the effectiveness of MXMap. Additionally, MXMap is compared against several baseline methods, showing advantages in accuracy and causal graph refinement.%

\end{abstract}

\begin{keywords}%
  Causal inference, state-space reconstruction, convergent cross-mapping, partial cross mapping, nonlinear dynamical system %
\end{keywords}

\section{Introduction}

Nonlinear dynamical systems are omnipresent across scientific disciplines, and understanding causal relationships in these systems is crucial for unveiling the underlying mechanisms that drive system behaviors. Classic causal inference methods, such as Granger Causality (GC)~\citep{granger1969investigating} and other functional causal models (FCMs), including the Additive Noise Model (ANM)~\citep{hoyer2008nonlinear, liu2024causal} and the Post Nonlinear Model (PNL)~\citep{zhang2015estimation, keropyan2023rank}, struggle with these systems due to their assumption of a predictive relationship from cause to effect, which does not hold in the presence of complex dynamics like coupling and chaos.


Convergent Cross Mapping (CCM)~\citep{sugihara2012detecting, barraquand2021inferring} was proposed as a model-free approach for bivariate causal inference in coupled dynamical systems. CCM addresses these limitations by leveraging state-space manifold reconstructions and cross mapping between reconstructed embeddings. Since its introduction, CCM has inspired further developments, including Partial Cross Mapping (PCM)~\citep{leng2020partial}, which aims to distinguish indirect from direct causalities in three-variable systems. However, PCM is limited to mapping operations between univariate delay embeddings, which can be less effective or even fail when dealing with complex systems with multiple interconnected variables~\citep{chen2022causation}.

To overcome this limitation, we propose \textbf{multiPCM}, an extension of PCM to the multivariate setting that allows for more effective causal inference by utilizing cross mapping via multivariate embeddings. We further integrate multiPCM with bivariate CCM into a two-phase framework named \textbf{MXMap} (Multivariate Cross Mapping for Causal Discovery). The proposed framework is designed for multivariate causal discovery, and is not only confined to assumptions of directed acyclic graphs (DAGs) but can also handle cycles. In the first phase, bivariate CCM generates an initial, potentially dense causal graph; In the second phase, multiPCM prunes indirect connections, refining the graph to isolate direct causal relationships. We systematically evaluate multiPCM and MXMap on benchmark datasets, including both simulated ecosystems and real-world meteorological data.

The contributions of this work are summarized as follows:

\begin{itemize}
    \item \textbf{Extension of PCM to multivariate settings}: We introduce multiPCM, which extends PCM to utilize multivariate delay embeddings for more robust causal inference in high-dimensional dynamical systems.
    \item \textbf{Two-phase causal discovery framework}: We propose MXMap, combining bivariate CCM with multiPCM to generate and refine causal graphs in nonlinear dynamical systems, which can also detect cycles.
    \item \textbf{Comprehensive evaluation on nonlinear dynamical systems}: We validate multiPCM and MXMap on simulated and real-world datasets. MXMap is compared against multiple baseline methods — including tsFCI, VAR-LiNGAM, PCMCI, Granger Causality, DYNOTEARS, SLARAC — demonstrating advantages in accuracy and refinement capabilities.
\end{itemize}
\section{Preliminaries}
\label{sec:prelim}

\subsection{State-Space Reconstruction (SSR)}
\label{sec:ssr}

In physical continuous-time dynamical systems, the interplay between driving forces and dissipation leads systems to settle into characteristic behaviors, represented by attractor manifolds in state space~\citep{milnor1985concept}. Understanding these attractors is crucial for interpreting the system's dynamics and predicting future behavior. However, real-world measurements are often limited, making it infeasible to observe all variables required to fully characterize the state-space attractor.


State-Space Reconstruction (SSR)~\citep{vlachos2008state} addresses this challenge: \textit{Given an $n$-dimensional observed time series from an $N$-dimensional dynamical system ($n < N$), can we recover the attractor manifold, and thereby the higher-dimensional dynamics, from the lower-dimensional observations?} A common approach is to use sequences of lagged observations to reconstruct a delay embedding (DE) that approximates the system's attractor. Whitney's Embedding Theorem~\citep{whitney1936differentiable} and Takens' Embedding Theorem~\citep{takens2006detecting} establish that this reconstruction is diffeomorphic (i.e., a continuously differentiable and invertible mapping) to the true attractor under certain conditions~\citep{sauer1991j}. When these conditions are satisfied, such delay embeddings are termed "shadow manifolds" and serve as low-dimensional approximations of the system~\citep{sugihara2012detecting}.


Following~\citep{vlachos2010nonuniform, butler2023causal}, we introduce the univariate delay-coordinate embedding used in Takens' Theorem. Suppose an attractor $\mathcal{A}$ exists for the dynamical system, and a time series $\{x_t\}_{t=0}^{T}$ is observed from one state variable $\textbf{x}$ sampled at a constant rate. Given delay $\tau$ and embedding dimension $E$, where $\tau$ and $E$ are positive integers, the vector signal $\vv{\textbf{m}}_{x}(t)$ of lagged values is defined as:

\begin{equation}
    \vv{\textbf{m}}_{x}(t) \vcentcolon= \left[x_{t},  x_{t-\tau},  x_{t-2\tau},  x_{t-3\tau},  \ldots,  x_{t-(E-2)\tau},  x_{t-(E-1)\tau}\right]
\end{equation}

As time progresses, these vectors form an $E$-dimensional delay embedding $\textbf{M}_x$. The lag $\tau$ determines the observation time scale for reconstruction, while the embedding dimension $E$ defines the complexity of the embedding. Takens' theorem suggests that $E$ should be greater than twice the fractal dimension of the attractor $\mathcal{A}$, i.e., $E > 2 \cdot \text{dim}(\mathcal{A})$~\citep{sauer1991j, kugiumtzis1996state}. In practice, $\tau$ and $E$ are determined empirically. Time-delayed autocorrelation~\citep{kugiumtzis1996state} and delay mutual information~\citep{fraser1986independent, klikova2011reconstruction} are commonly used to select an optimal $\tau$, while the \textit{false nearest neighbors} (FNN) method~\citep{kennel1992determining} is typically used to determine $E$, by tracking changes in nearest neighbors as embedding dimensions increase.

\subsection{Convergent Cross Mapping (CCM)}
\label{sec:uni-ccm}
Causality in a discrete-time dynamical system~\citep{butler2023causal,cummins2015efficacy} can be defined as follows: given two state variables $\textbf{x}$ and $\textbf{y}$, if the future evolution of $\textbf{y}$ depends on $\textbf{x}$, then $\textbf{x}$ is said to cause $\textbf{y}$, denoted as $\textbf{x} \Rightarrow \textbf{y}$. This causal influence can be represented in a state-space equation, as shown in Eq.~\ref{eq:def_cause}:

\begin{equation}
\label{eq:def_cause}
    \textbf{y}_{t+1}=\mathcal{F}_y \left(\textbf{y}_{t}, \textbf{x}_{t} \right)
\end{equation}

The relationship between $\textbf{x}$ and $\textbf{y}$ can be unidirectional ($\textbf{x} \Rightarrow \textbf{y}$ or $\textbf{y} \Rightarrow \textbf{x}$), bidirectional ($\textbf{x} \Leftrightarrow \textbf{y}$), or there may be no causal link at all.


CCM~\citep{sugihara2012detecting} leverages the diffeomorphism between reconstructed shadow manifolds, as stated in Takens' Theorem. Cross mapping measures how well local neighborhoods in one reconstructed manifold map to the corresponding neighborhoods in another. For delay embeddings $\textbf{M}_x$ and $\textbf{M}_y$ reconstructed from $\textbf{x}$ and $\textbf{y}$, if $\textbf{M}_x$ and $\textbf{M}_y$ are both valid shadow manifolds of the attractor $\mathcal{A}$, they are diffeomorphic to each other via their relationship to $\mathcal{A}$.

If $\textbf{x}$ causes $\textbf{y}$ ($\textbf{x} \Rightarrow \textbf{y}$), observations of $\textbf{y}$ should contain information about $\textbf{x}$. This allows the reconstruction of the dynamics of $\textbf{x}$ from $\textbf{y}$, but not necessarily vice versa. In this case, the quality of mapping from $\textbf{M}_y$ to $\textbf{M}_x$ should be better compared to the reverse direction, indicating a causal link from $\textbf{x}$ to $\textbf{y}$.

CCM uses a $k$-nearest neighbor ($k$NN) regression approach (also known as \textit{simplex projection}) to evaluate the quality of cross mapping. Given time series $\{x_t\}_{t=0}^{T}$ and $\{y_t\}_{t=0}^{T}$, to verify whether $\textbf{x} \Rightarrow \textbf{y}$, the procedure is as follows:

\begin{enumerate}[label={[\arabic*]}]
    \item Construct delay embeddings $\textbf{M}_x, \textbf{M}_y$ with appropriate delay $\tau$ and embedding dimension $E$.
    \item For each point in $\textbf{M}_y$, identify the $k$-nearest neighbors $\mathcal{N}_y$.
    \item Use the timestamps of $\mathcal{N}_y$ to find corresponding points $\hat{\mathcal{N}}_x$ on $\textbf{M}_x$ and compute a weighted average to form a projected reconstruction $\hat{\textbf{M}}_x$, hence the reconstructed $\hat{\mathbf{x}}$.
    \item Calculate the correlation score $\rho_{x\Rightarrow y}$ between the true $\mathbf{x}$ and the reconstructed $\hat{\mathbf{x}}$.
    \item Repeat these steps with increasing sequence length; if $\textbf{x} \Rightarrow \textbf{y}$, the correlation score should converge, indicating a valid cross map.
\end{enumerate}

The same procedure is repeated for the reverse causal assumption to yield another correlation score $\rho_{y\Rightarrow x}$. The correlation scores $\rho_{x\Rightarrow y}$ and $\rho_{y\Rightarrow x}$ quantify the cross mapping quality, where a higher score in one direction suggests a stronger causal link. In practice, if the input length $L$ is large enough, we consider that the yielded correlation scores are already in the convergence zone, and can be used as final correlation estimates.

\subsection{Partial Cross Mapping (PCM)}
\label{sec:uni-pcm}
The original CCM does not distinguish between direct and indirect causality. For three variables $\textbf{x}$, $\textbf{y}$, and $\textbf{z}$ in a causal chain ($\textbf{x}\Rightarrow\textbf{y}\Rightarrow\textbf{z}$), CCM may incorrectly identify a direct causal link between $\textbf{x}$ and $\textbf{z}$ due to transitivity through $\textbf{y}$. Partial Cross Mapping (PCM)~\citep{leng2020partial,jiang2023partial} was proposed to distinguish between direct and indirect causal relationships. In a causal chain like $\textbf{x}\Rightarrow\textbf{y}\Rightarrow\textbf{z}$, PCM aims to determine whether the causal link between $\textbf{x}$ and $\textbf{z}$ is direct or mediated by $\textbf{y}$.

A PCM test considers three variables: the \textit{potential cause} $\textbf{x}$, the \textit{condition} $\textbf{y}$, and the \textit{potential effect} $\textbf{z}$. The goal is to assess whether there is a direct link between $\textbf{x}$ and $\textbf{z}$. This is done by performing cross mapping between the shadow manifolds of each variable to obtain a reconstruction of $\textbf{x}$, denoted by $\hat{\textbf{x}}^{\textbf{z}}$ (from $\textbf{z}$ to $\textbf{x}$), and another reconstruction of $\textbf{x}$ via $\textbf{y}$, denoted by $\hat{\textbf{x}}^{\hat{\textbf{y}}^{\textbf{z}}}$ (first from $\textbf{z}$ to $\textbf{y}$, then from $\textbf{y}$ to $\textbf{x}$). The correlation scores are defined as follows:

\begin{align}
    \rho_{All}=\left| \text{Corr}(\textbf{x}, \hat{\textbf{x}}^{\textbf{z}}) \right|   &&    \rho_{Direct}=\left| \text{ParCorr}(\textbf{x}, \hat{\textbf{x}}^{\textbf{z}} | \hat{\textbf{x}}^{\hat{\textbf{y}}^{\textbf{z}}} ) \right|
\end{align}

$\rho_{All}$ represents the correlation between the original $\textbf{x}$ and the reconstruction $\hat{\textbf{x}}^{\textbf{z}}$, capturing apparent information transfer through all paths. $\rho_{Direct}$ represents the partial correlation, conditioning on the intermediate variable $\textbf{y}$ to assess direct information transfer between $\textbf{x}$ and $\textbf{z}$. If no direct causal link exists, the direct information transfer should be significantly reduced after conditioning on $\textbf{y}$.

PCM uses an empirical threshold $H \in [0, 1)$ to determine causality:
\begin{itemize}
    \item If $\rho_{All} \geq \rho_{Direct} \geq H$, a direct causal link from $\textbf{x}$ to $\textbf{z}$ is inferred.
    \item If $\rho_{All} \geq H \gg \rho_{Direct}$, only indirect causality is suggested.
    \item If $H > \rho_{All} \geq \rho_{Direct}$, no causal relationship is inferred.
\end{itemize}

To distinguish direct from indirect links, we propose an adaptation in our work, to use the correlation ratio $\gamma$ as an alternative:
\begin{equation}
    \gamma=\frac{\rho_{Direct}}{\rho_{All}}
\end{equation}
A smaller ratio $\gamma$ implies negligible direct information transfer after conditioning, suggesting indirect causality via $\textbf{y}$. Conversely, a larger $\gamma$ indicates strong direct information transfer, suggesting a direct causal link. An empirical ratio threshold $\gamma^* \in (0, 1)$ is used to decide whether to retain or eliminate the direct link based on how important such causal influence is.

\section{Methodology}
\label{sec:method}
We present key components of our approach: multivariate state space reconstruction for capturing coupled system dynamics; multivariate partial cross mapping to distinguishes between direct and indirect causality in multivariate settings; MXMap, the proposed framework leveraging the two for multivariate causal discovery.

\subsection{Multivariate State Space Reconstruction (multiSSR)}
\label{sec:multi-ssr}

Consider a set of $K$ observed time series denoted as $\left\{ x^{[k]}_{t} \right\}_{t=0}^{T},  k=1, 2, 3, ..., K$ from a dynamical system. Building upon the univariate delay embedding introduced in Section~\ref{sec:ssr}, each time series generates a univariate embedding $\mathcal{M}_{X}^{[k]}$. The vector $\vv{\textbf{m}}_{x}^{[k]}(t)$ on $\mathcal{M}_{x}^{[k]}$ is given by:

\begin{equation}
    \vv{\textbf{m}}_{x}^{[k]}(t)= \left[x_{t}^{[k]},  x^{[k]}_{t-\tau},  x^{[k]}_{t-\tau\cdot2},  x^{[k]}_{t-\tau\cdot3},  ...,  x^{[k]}_{t-\tau\cdot\left(E-2\right)},  x^{[k]}_{t-\tau\cdot\left(E-1\right)}\right]
\end{equation}

Multivariate State-Space Reconstruction (multiSSR) constructs a multivariate embedding by stacking all $K$ univariate delay embeddings. A vector $\textbf{m}_{x}(t)$ on $\mathcal{M}_{x}(t)$ is represented as:

\begin{equation}
    \vv{\textbf{m}}_{x}(t)= \left[ \vv{\textbf{m}}_{x}^{[1]}(t) \ \ \   \vv{\textbf{m}}_{x}^{[2]}(t) \ \ \     \vv{\textbf{m}}_{x}^{[3]}(t) \ \ \    ... \ \ \   \vv{\textbf{m}}_{x}^{[K]}(t)\right]
\end{equation}




For simplicity and consistency, we adopt a uniform multiSSR scheme, where the delay values $\tau$ and embedding dimensions $E$ are the same for all $K$ time series, as suggested by \citep{vlachos2010nonuniform}.

\begin{algorithm}[htb]
    \caption{MXMap Workflow}
    \label{alg:mxmap}
    \scriptsize
    \KwData{Delay $\tau$, dimension $E$; Time series of a multivariate dynamical system $G = \{X_1, X_2, \ldots, X_n\}$; Variable indices $S := \{1, \ldots, K\}$; PCM threshold of correlation ratio $\gamma^\star$.}
    \KwResult{Adjacency matrix $\mathcal{A}$; Final children dictionary $CH$} 

    \textbf{Phase 1: Initial Causal Graph} 
    
    \For{$i \in S$}{
        \For{$j \in S$ and $j \neq i$}{
            Embed $X_i$ and $X_j$ with delay parameters $\tau$ and $E$; 
            Perform cross mappings to reconstruct $X_i$ from $X_j$ and vice versa; 

            Compute scores $\beta_{X_i\Rightarrow X_j}$ and $\beta_{X_j\Rightarrow X_i}$;

            \If{$\beta_{X_i\Rightarrow X_j} > \beta_{X_j\Rightarrow X_i}$ and both scores $\geq 0.5$}{
                Establish link $X_i\Rightarrow X_j$ (set $\mathcal{A}_{i,j}=1$ and add $j$ to $CH[i]$);
            }
            \ElseIf{$\beta_{X_j\Rightarrow X_i} > \beta_{X_i\Rightarrow X_j}$ and both scores $\geq 0.5$}{
                Establish link $X_j\Rightarrow X_i$ (set $\mathcal{A}_{j,i}=1$ and add $i$ to $CH[j]$);
            }
        }
    }

    \textbf{Phase 2: Prune Indirect Connections}

    \For{$i \in S$}{
        \For{$j \in CH[i]$}{
            \If{Longest path between $X_i$ and $X_j$ has more than 2 nodes}{
                Identify all intermediate nodes between $X_i$ and $X_j$ as set $\textit{Conds}$;
                
                Embed $X_i$, $X_j$, and $\textit{Conds}$; perform multiPCM to compute reconstructions;

                Calculate scores $\rho_{All}$, $\rho_{Direct}$, and ratio $\gamma = \rho_{Direct} / \rho_{All}$;

                \If{$\gamma < \gamma^\star$}{
                    Remove link $X_i\Rightarrow X_j$ (set $\mathcal{A}_{i,j}=0$ and remove $j$ from $CH[i]$);
                }
            }
        }
    }
    
\end{algorithm}

\subsection{Multivariate Partial Cross Mapping (multiPCM)}
\label{sec:multi-pcm}

The original PCM method considers three univariate inputs: the potential cause, effect, and condition variables. Multivariate Partial Cross Mapping (multiPCM) extends this by allowing the condition set to be multivariate, making it better suited for high-dimensional systems.

Consider a set of variables $\mathcal{G}$, with a quasi-chain structure $X_1 \Rightarrow \mathcal{G} \setminus {\{X_1, X_2\}} \Rightarrow X_2$. Here, $X_1$ is the alleged cause, $X_2$ is the alleged effect, and the rest of the variables, denoted as $\textit{Conds} := \mathcal{G} \setminus {\{X_1, X_2\}}$, serve as the condition set.

The two univariate inputs, $X_1$ and $X_2$, form univariate delay embeddings, while the multivariate condition set forms a multivariate embedding as described in Section~\ref{sec:multi-ssr}.

Similar to the original PCM as described in Section~\ref{sec:uni-pcm}, we conduct the apparent (univariate) and conditioned (multivariate) cross mappings to obtain the two correlation scores. 

\begin{itemize}
    \item \textbf{Apparent cross mapping:} Perform univariate cross mapping to reconstruct $X_1$ from $X_2$, denoted as $\hat{X_1}^{X_2}$.
    \item \textbf{Conditioned cross mapping:} Reconstruct the condition set from $X_2$, denoted by $\widehat{\textit{Conds}}^{X_2}$, and then reconstruct $X_1$ from $\widehat{\textit{Conds}}^{X_2}$, resulting in $\hat{X_1}^{\widehat{\textit{Conds}}^{X_2}}$.
\end{itemize}

The two correlation scores are then computed as follows:


\begin{align}
    \rho_{All}=\left| Corr(X_1, \hat{X_1}^{X_2}) \right| &&  \rho_{Direct}=\left| ParCorr(X_1, \hat{X_1}^{X_2} | {\hat{X_1}^{\widehat{\textit{Conds}}^{X_2}}} ) \right|
\end{align}

Following a similar reasoning as in Section~\ref{sec:uni-pcm}, we determine whether a direct causal link exists between $X_1$ and $X_2$ by calculating the correlation ratio $\gamma = \frac{\rho_{Direct}}{\rho_{All}}$ and comparing it against an empirical threshold.

\subsection{Multivariate Cross Mapping (MXMap) for Multivariate Causal Discovery in Dynamical Systems}
\label{sec:mxmap}

We propose MXMap as a two-phase framework for multivariate causal discovery in dynamical systems. Inspired by the structure of RESIT (Regression with Subsequent Independence Test)~\citep{peters2014causal} algorithm, MXMap first establishes an initial causal graph and then prunes indirect connections (Algorithm~\ref{alg:mxmap}). 

The details of each phase are as follows:
\begin{itemize}
    \item \textbf{Phase 1 (Establish Initial Causal Graph):} 
    In the first phase, MXMap applies exhaustive bivariate CCM tests between all pairs of variables to establish an initial causal graph. For each pair of variables, we embed their time series using delay embedding parameters ($\tau$, $E$) and compute delay embeddings $\mathcal{M}_{X_i}$ and $\mathcal{M}_{X_j}$. Cross mapping is then performed between these embeddings to reconstruct each variable, yielding reconstructed series $\hat{X_i}$ and $\hat{X_j}$. Correlation scores are calculated for both directions—$\beta_{X_i \Rightarrow X_j}$ and $\beta_{X_j \Rightarrow X_i}$. The direction with the higher score is chosen as the causal direction, and a link is added to the initial graph accordingly. For this CCM phase, if both scores are below a correlation threshold (0.5), no link is established.
    \item \textbf{Phase 2 (Prune Indirect Causal Connections):} The initial causal graph generated in Phase 1 may contain indirect connections, which can result in an overly dense graph. In the second phase, MXMap uses multiPCM to refine the graph by pruning indirect links. For each parent-child pair in the initial graph, we identify all intermediate variables forming paths between them. We then apply multiPCM to assess whether the causal link between the parent and child is direct or goes through intermediate variables. Specifically, we embed the parent, child, and intermediate variables, and compute two correlation scores: $\rho_{All}$, representing overall correlation, and $\rho_{Direct}$, representing partial correlation conditioned on the intermediate variables and assessing the direct information transfer between parent and child variables. We calculate the correlation ratio $\gamma = \frac{\rho_{Direct}}{\rho_{All}}$ and compare it to a predefined ratio threshold $\gamma^\star$. If ratio $\gamma$ is below $\gamma^\star$, the direct information transfer isn't considered important enough, and the parent-child link is considered indirect and is removed from the graph.
\end{itemize}

\section{Experiments}
\label{sec:exp}


We evaluate the proposed multiPCM and MXMap on simulated and real-world dynamical systems as described in the Section~\ref{sec:data} below. MXMap performance is then compared with several established multivariate causal inference methods), including RESIT~\citep{peters2014causal}, tsFCI~\citep{entner2010causal}, VAR-LiNGAM~\citep{hyvarinen2010estimation}, PCMCI~\citep{runge2019detecting}, Granger Causality~\citep{granger1969investigating}, DYNOTEARS~\citep{pamfil2020dynotears}, and SLARAC~\citep{weichwald2020causal}. Overview of these methods and how the outputs of each model are interpreted can be found in Appendix~\ref{appsec:baseline_methods}. 

For RESIT and VAR-LiNGAM, we adopt python implementations in the \texttt{LiNGAM} library~\citep{shimizu2014lingam}; tsFCI is adapted based on the implementation of FCI from the \texttt{causal-learn} library~\citep{zheng2024causal}; PCMCI is from the \texttt{tigramite} library~\citep{runge2023causal}; Granger Causality is also from \texttt{causal-learn}; DYNOTEARS is implemented using the \texttt{Causalnex}~\citep{Beaumont_CausalNex_2021} library; SLARAC is from the \texttt{tidybench} repository~\citep{weichwald2020causal}. Experimental setup is provided in Appendix~\ref{appsec:exp_setup}.

\subsection{Data}
\label{sec:data}

\subsubsection{Simulated Data: Species Interaction Systems}
\label{sec:genData}

Following similar data generation scheme as in~\cite{leng2020partial}, we generate the following systems of varying complexity, from 3-variable to 7-variable. These systems are derived and adapted from the Lotka-Volterra competition models~\citep{volterra1931theorie, lotka1925elements, roques2011probing} that characterize species interactions and exhibit chaotic behaviors. Examples of 3-species and 4-species systems are as Eq.~\ref{eq:3var} and Eq.~\ref{eq:4var}.

\begin{equation}
\label{eq:3var}
\begin{aligned}
x_t & =x_{t-1}\left(\alpha_x-\alpha_x x_{t-1}-\beta_{yx} y_{t-1}-\beta_{zx} z_{t-1}\right)\cdot\eta_{x}+\epsilon_{x} \\
y_t & =y_{t-1}\left(\alpha_y-\alpha_y y_{t-1}-\beta_{xy} x_{t-1}-\beta_{zy} z_{t-1}\right)\cdot\eta_{y}+\epsilon_{y} \\
z_t & =z_{t-1}\left(\alpha_z-\alpha_z z_{t-1}-\beta_{xz} x_{t-1}-\beta_{yz} y_{t-1}\right)\cdot\eta_{z}+\epsilon_{z}
\end{aligned}
\end{equation}
For the 3-species system (Eq.~\ref{eq:3var}), the coefficients $\alpha$ for autonomous dynamics are set respectively as $\alpha_x=3.70, \alpha_y=3.78, \alpha_z=3.72$.

\begin{equation}
\label{eq:4var}
\begin{aligned}
w_t & =w_{t-1}\left(\alpha_w-\alpha_w w_{t-1}-\beta_{xw} x_{t-1}-\beta_{yw} y_{t-1}-\beta_{zw} z_{t-1}\right)\cdot\eta_{w}+\epsilon_{w} \\
x_t & =x_{t-1}\left(\alpha_x-\alpha_x x_{t-1}-\beta_{wx} w_{t-1}-\beta_{yx} y_{t-1}-\beta_{zx} z_{t-1}\right)\cdot\eta_{x}+\epsilon_{x} \\
y_t & =y_{t-1}\left(\alpha_y-\alpha_y y_{t-1}-\beta_{wy} w_{t-1}-\beta_{xy} x_{t-1}-\beta_{zy} z_{t-1}\right)\cdot\eta_{y}+\epsilon_{y} \\
z_t & =z_{t-1}\left(\alpha_z-\alpha_z z_{t-1}-\beta_{wz} w_{t-1}-\beta_{xz} x_{t-1}-\beta_{yz} y_{t-1}\right)\cdot\eta_{z}+\epsilon_{z}
\end{aligned}
\end{equation}
For the 4-species system (Eq.~\ref{eq:4var}), the coefficients $\alpha_w=3.70, \alpha_x=3.78, \alpha_y=3.72, \alpha_z=3.70$ are used. For each system, the coupling coefficient $\beta_{ij}$ is either 0 or 0.35, depending on whether the causal interaction is present or absent. 



For higher dimensional systems, we follow a similar logic, where the autonomous dynamics coefficients $\alpha$ are sampled in range $[3.70, 3.80)$, and the coupling coefficients $\beta$ are 0.35 when there is causal interaction between a variable pair (0 if no causal interaction). These systems are used to evaluate the effectiveness of multiPCM and MXMap in capturing both direct and indirect causal links in controlled scenarios. The ground truth causal structures of all used systems (from 3-variable to 7-variable) are demonstrated in Appendix~\ref{appsec:complet}.

\subsubsection{ERA5 Reanalysis Meteorological Data}

We also evaluate MXMap on real-world meteorological data from the ERA5 Global Reanalysis dataset~\citep{hersbach2020era5}, provided by the Copernicus program by ECMWF. The ERA5 dataset offers hourly climate variables over a global scale, allowing us to investigate causality in a practical environmental setting.


We extract hourly winter data (December to February) for the Montreal region from 1981 to 2023. Two experimental setups were designed to assess the effectiveness of causal discovery methods:

\begin{itemize}
    \item \textbf{3V Chain}: 
    A simplified 3-variable system capturing the causal chain: $tcw \Rightarrow rad \Rightarrow  T_{2m}$. 
    This causal chain is particularly pronounced in winter when cloud coverage strongly affects radiation levels, which subsequently modulate ground-level temperatures.
    \item \textbf{5V System}: This system includes solar radiation ($rad_{solar}$), terrestrial radiation ($rad_{terr}$), near-ground temperature advection ($T_{adv950}$), total cloud water ($tcw$) and near-ground temperature ($T_{2m}$). This setup examines whether well-established causal relationships can be detected, namely $rad_{solar} \Rightarrow T_{2m}$, $rad_{terr} \Rightarrow T_{2m}$, $T_{adv950} \Rightarrow T_{2m}$, $tcw\Rightarrow rad_{solar}$. 
    The 5V system builds on the 3V chain by introducing two radiation components as well as a temperature advection term. While the complete causal graph of this system is not fully established due to its complexity, the bivariate relationships outlined above are well-supported in meteorological literature and provide a robust benchmark for evaluation.
\end{itemize}

Detailed explanations of these mechanisms and meteorological contexts are provided in Appendix~\ref{appsec:era5}. By focusing on winter data, when these causal mechanisms are most pronounced, these setups serve as an ideal test case for assessing MXMap’s ability to uncover causal relationships in complex, real-world environmental systems.

\subsection{Validation of multiPCM}
\label{sec:valid_multiPCM}

To validate the effectiveness of multiPCM in distinguishing between direct and indirect causal relationships, we perform experiments on simulated four-variable systems generated without noise. Specifically, we evaluate three scenarios: purely direct causality, purely indirect causality, and combined direct and indirect causality, as illustrated in Table~\ref{tab:multiPCM}. The causal relationships of interest are highlighted in color, while the other variables are shown in gray and grouped together to form a multivariate embedding, used as the condition set ($Conds$) for multiPCM.

We use an input length of $L = 3500$ time steps for all tests and conduct multiPCM on a range of lag and embedding dimension values ($\tau, E \in \{1, 2, 3, \ldots, 8\}$). The results of the grid search are presented in Table~\ref{tab:multiPCM} (complete table with more cases in Appendix~\ref{appsec:valid_multiPCM}), where we analyze the correlation ratio $\gamma=\rho_{Direct}/\rho_{All}$ and the predicted labels indicating whether direct causality between colored nodes is rejected and to be removed based on a PCM threshold of 0.45 (this empirical threshold selection is discussed in Appendix~\ref{appsec:thres_multiPCM}). \textcolor{red}{Red} label indicates rejection, since only indirect causality exists; while \textcolor{blue}{blue} label indicates the existence direct causality, hence the link between colored nodes should be kept.

\begin{table}[htb]
\centering

\makebox[\linewidth]{%

\resizebox{0.85\textwidth}{!}{

\begin{tabular}{c|c|c|c}
Type         & $Direct$ & $Indirect$ & $Both$ \\ \hline
Causality         &\begin{minipage}{.15\linewidth} \centering \includegraphics[width=0.3\linewidth]{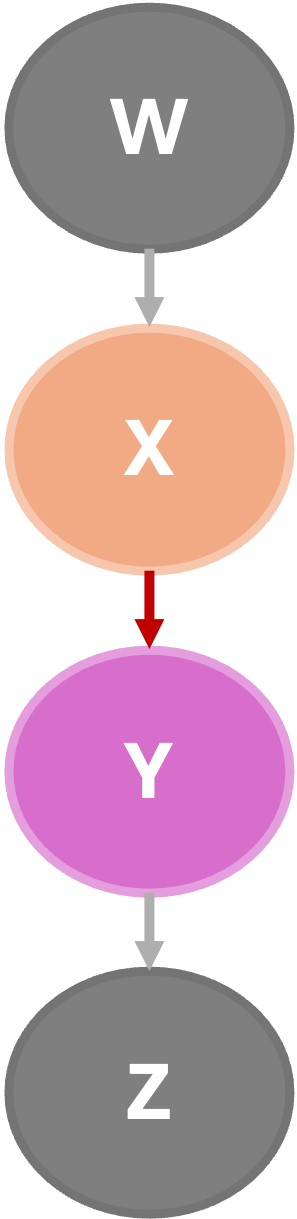} \end{minipage}& \begin{minipage}{.15\linewidth} \centering \includegraphics[width=0.3\linewidth]{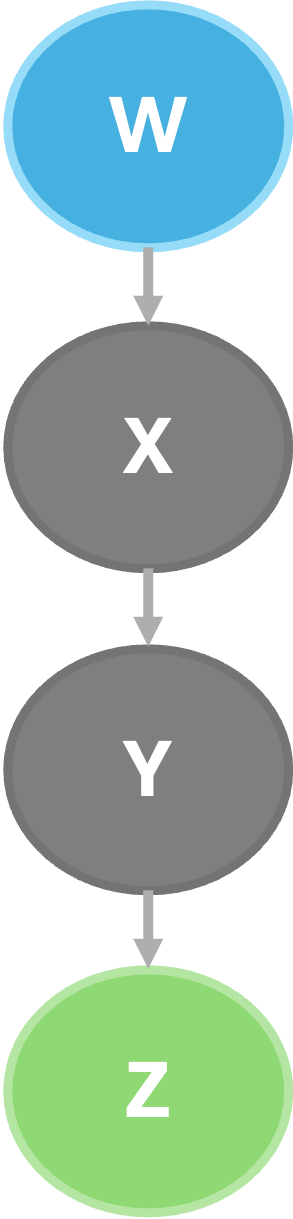} \end{minipage}   & \begin{minipage}{.15\linewidth} \centering \includegraphics[width=0.5\linewidth]{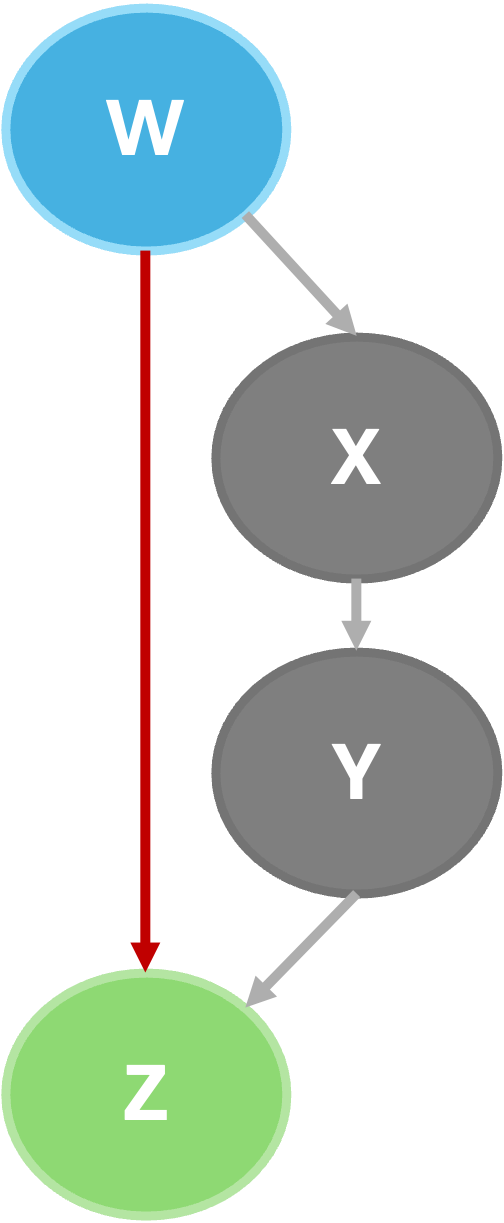} \end{minipage}   \\ \hline
$\rho_{All}$    &\begin{minipage}{.3\linewidth} \centering \includegraphics[width=\linewidth]{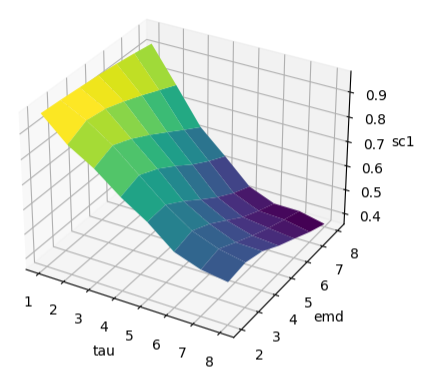} \end{minipage}& \begin{minipage}{.3\linewidth} \centering \includegraphics[width=\linewidth]{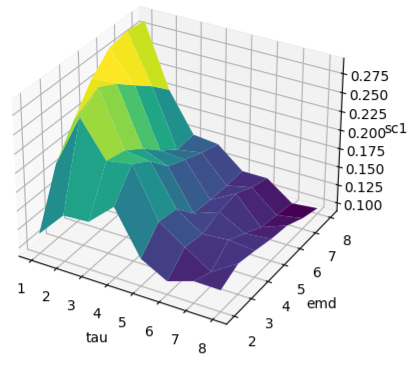} \end{minipage}    &  \begin{minipage}{.3\linewidth} \centering \includegraphics[width=\linewidth]{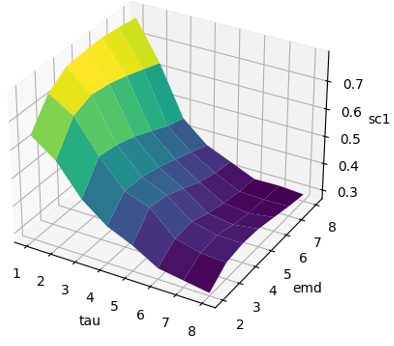} \end{minipage}  \\ \hline
$\rho_{Direct}$ &\begin{minipage}{.3\linewidth} \centering \includegraphics[width=\linewidth]{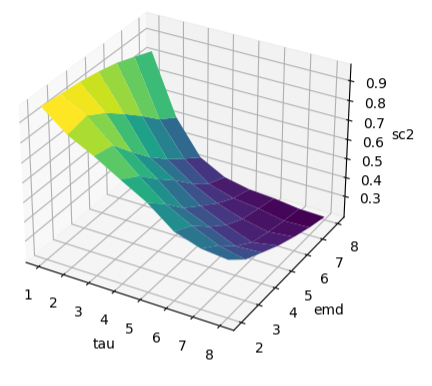} \end{minipage}& \begin{minipage}{.3\linewidth} \centering \includegraphics[width=\linewidth]{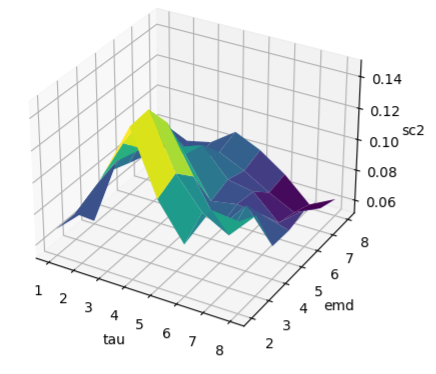} \end{minipage}    & \begin{minipage}{.3\linewidth} \centering \includegraphics[width=\linewidth]{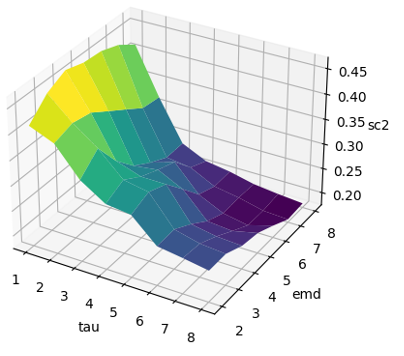} \end{minipage} \\ \hline
Ratio             &\begin{minipage}{.3\linewidth} \centering \includegraphics[width=\linewidth]{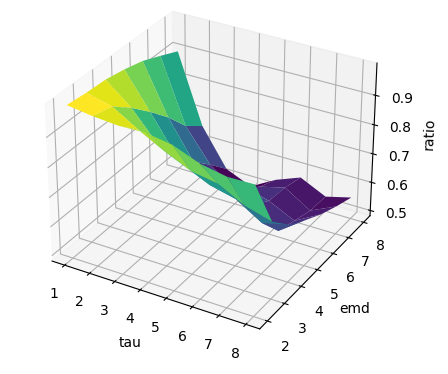} \end{minipage}& \begin{minipage}{.3\linewidth} \centering \includegraphics[width=\linewidth]{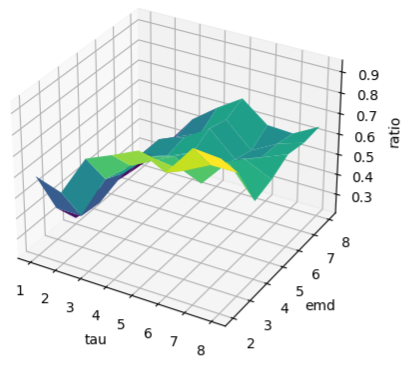} \end{minipage}    & \begin{minipage}{.3\linewidth} \centering \includegraphics[width=\linewidth]{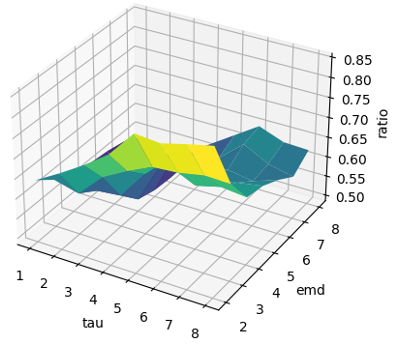} \end{minipage}  \\ \hline
Label    &\begin{minipage}{.3\linewidth} \centering \includegraphics[width=\linewidth]{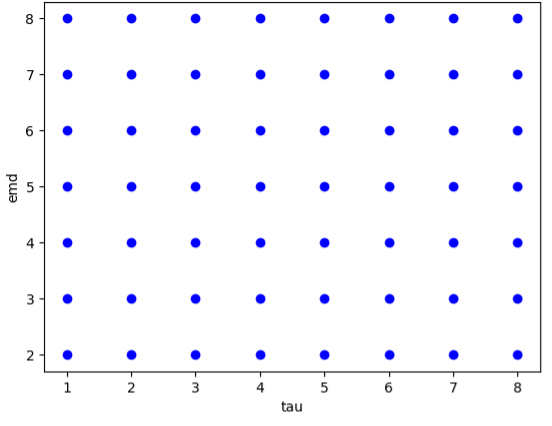} \end{minipage}& \begin{minipage}{.3\linewidth} \centering \includegraphics[width=\linewidth]{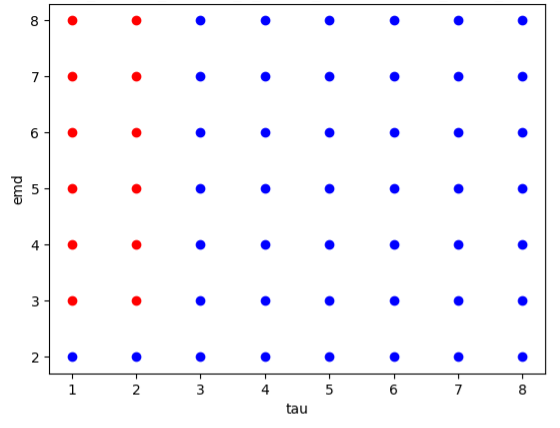} \end{minipage}    & \begin{minipage}{.3\linewidth} \centering \includegraphics[width=\linewidth]{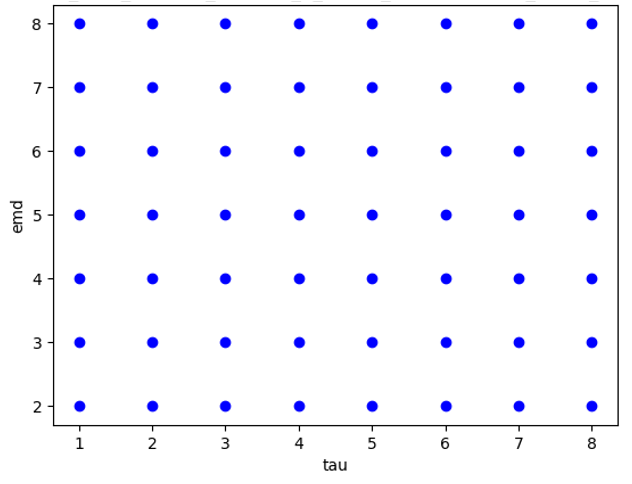} \end{minipage}
\end{tabular}

}

}

\caption{Performance of multiPCM (Full Table in Appendix~\ref{appsec:valid_multiPCM}): Profiles of correlation scores, correlation ratios, predicted label ($thres=0.45$) under grid search. Red dot indicates there isn't direct causality between the colored nodes, while blue indicates there is direct causality between the colored nodes.}

\label{tab:multiPCM}

\end{table}

The experimental results demonstrate the effectiveness of multiPCM in distinguishing direct and indirect causal relationships under different scenarios:
\begin{enumerate}
    \item \textbf{Direct Causality:} For cases involving direct causality ($Direct$ and $Both$), we observe that both correlation scores, $\rho_{All}$ and $\rho_{Direct}$, are significantly higher when the lag $\tau$ is small, and start dropping as lag increases. This trend is also consistent with observations from previous works on cross mapping in nonlinear systems.
    \item \textbf{Indirect Causality:} In the case of purely indirect causality, the correlation profiles tend to be inconsistent across different lags and embedding dimensions, showing fluctuating surfaces rather than a clear decreasing trend. This behavior is characteristic of indirect interactions, as these relationships become weaker and more unstable with increasing delay.
    \item \textbf{Optimal Hyperparameter Range:} 
    The grid search results for predicted labels further illustrate the behavior of multiPCM. There exists an ideal range of lag and embedding dimension values for accurately inferring causality: Here in Table~\ref{tab:multiPCM}, when the multiPCM lag ($\tau$) is less than 2 and the embedding dimension ($E$) is in the interval of $[3, 8]$, multiPCM produces consistent and accurate results across these three causal structures. In practice, if the approximate system dimension and timescale of the lag are known, the appropriate selections for $\tau$ and $E$ would likely align closely with the ground truth values: For the demonstrated 3-variable (3V) and 4-variable (4V) systems, the actual system dimension is 3 or 4, the generating dynamics use a lag of 1 (All these ground-truth $\tau$ and $E$ values fall in the detected ranges above). Notably, the selection of $E$ seems to be more tolerant for slight overestimation when the true state dimension is unknown. Thus, when the exact ground truth is unavailable, a slightly higher dimension for delay embedding may still yield reliable results, and can potentially help account for latent variables.
\end{enumerate}

These observations show that multiPCM is capable of correctly distinguishing direct and indirect causalities in multivariate scenarios. By performing cross mapping with multivariate embeddings, multiPCM achieves consistent causal inference that is robust across different lag and embedding dimension configurations.

\subsection{Prediction Consistency}

Mirage correlations are common in nonlinear dynamical systems. Coupled nonlinear systems often exhibit transient correlations between variables, which may change or disappear entirely when different subsequences are sampled. This presents a significant challenge for causal discovery, particularly for methods relying on consistent predictive relationships, such as Granger causality. The original CCM paper~\cite{sugihara2012detecting} illustrated this phenomenon using a bivariate system of competing species.

\begin{figure}[htb]
    \centering
    \begin{minipage}[b]{0.31\linewidth}
        \centering
        \includegraphics[width=\linewidth]{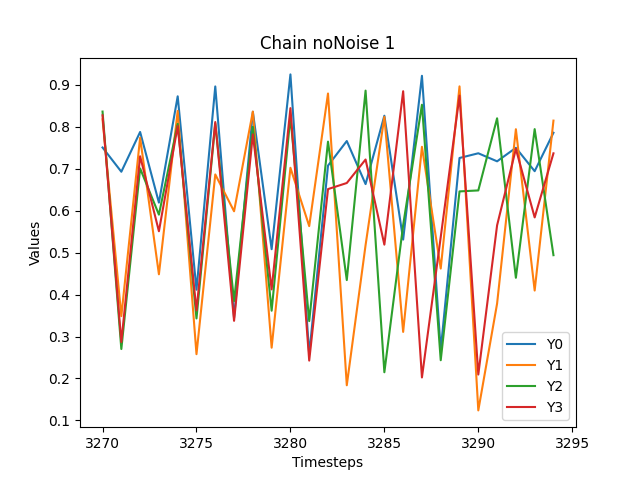}
        \captionsetup{justification=centering}
        \caption*{(a) Subsequence 1}
    \end{minipage}
    \begin{minipage}[b]{0.31\linewidth}
        \centering
        \includegraphics[width=\linewidth]{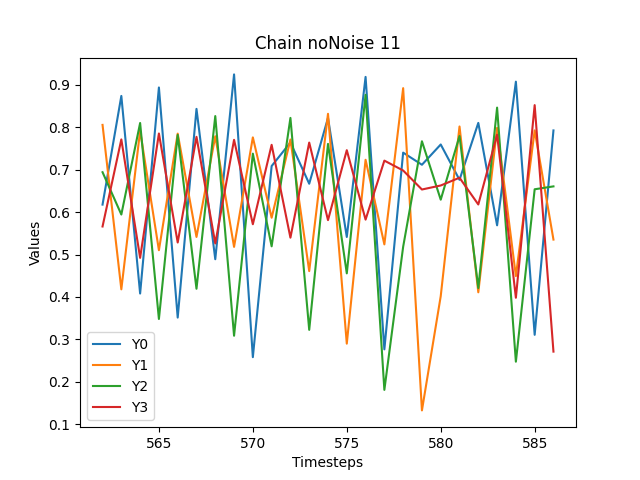}
        \captionsetup{justification=centering}
        \caption*{(b) Subsequence 2}
    \end{minipage}
    \begin{minipage}[b]{0.31\linewidth}
        \centering
        \includegraphics[width=\linewidth]{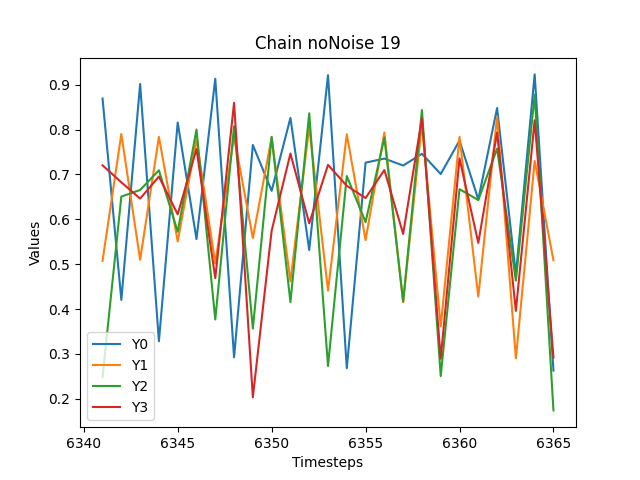}
        \captionsetup{justification=centering}
        \caption*{(c) Subsequence 3}
    \end{minipage}
    \caption{Visualizations of subsequences from a 4-species chain system: In the same sequence, correlations between variables can be positive, negative or zero when sampling from different start points.}
    \label{fig:4varChainNoNoise}
\end{figure}

To evaluate the robustness of our proposed approach in such scenarios, we first illustrate the chaotic behavior and mirage correlations in the noise-free four-species chain system ($w \Rightarrow x \Rightarrow y \Rightarrow z$), as defined in Section~\ref{sec:genData}. We randomly sample different starting points and visualize subsequences (of length 25) from these sampled points in Fig.~\ref{fig:4varChainNoNoise}. As shown, the correlations between variables vary widely across different subsequences, exhibiting positive, near-zero, and even negative correlations.

\begin{table}[htb]
\centering

\resizebox{1.0\textwidth}{!}{
\begin{tabular}{l|c|ccccc}
\textbf{Model}            & \textbf{  MXMap  } & \multicolumn{5}{c}{\textbf{RESIT-MLP}}                                                                             \\ \hline
\textbf{Prediction} &   
\begin{minipage}{.045\textwidth}
\centering
    \includegraphics[width=\linewidth]{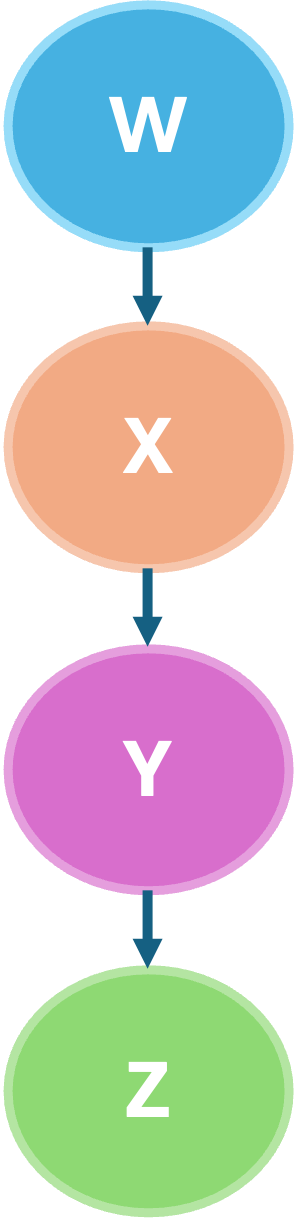}
\end{minipage} 
& \multicolumn{1}{c|}{
\begin{minipage}{.12\textwidth}
\centering
    \includegraphics[width=\linewidth]{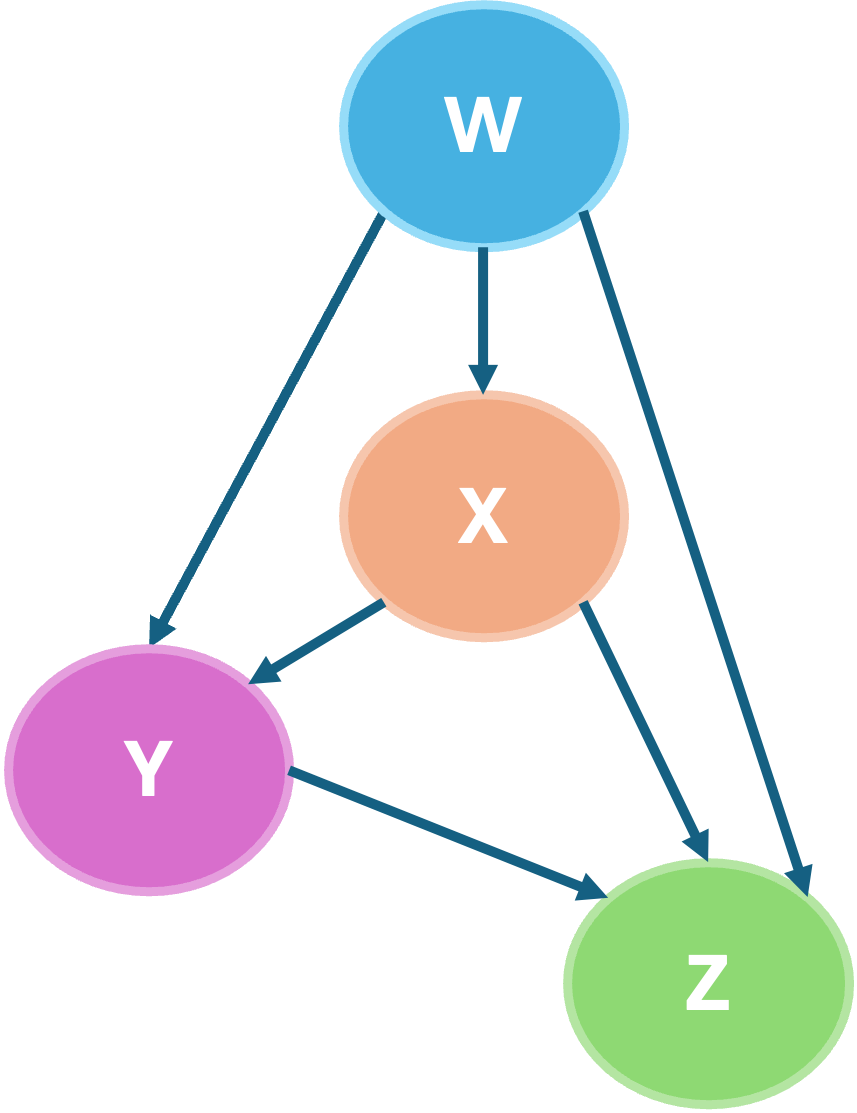}
\end{minipage} 
}  & \multicolumn{1}{c|}{
\begin{minipage}{.12\textwidth}
\centering
    \includegraphics[width=\linewidth]{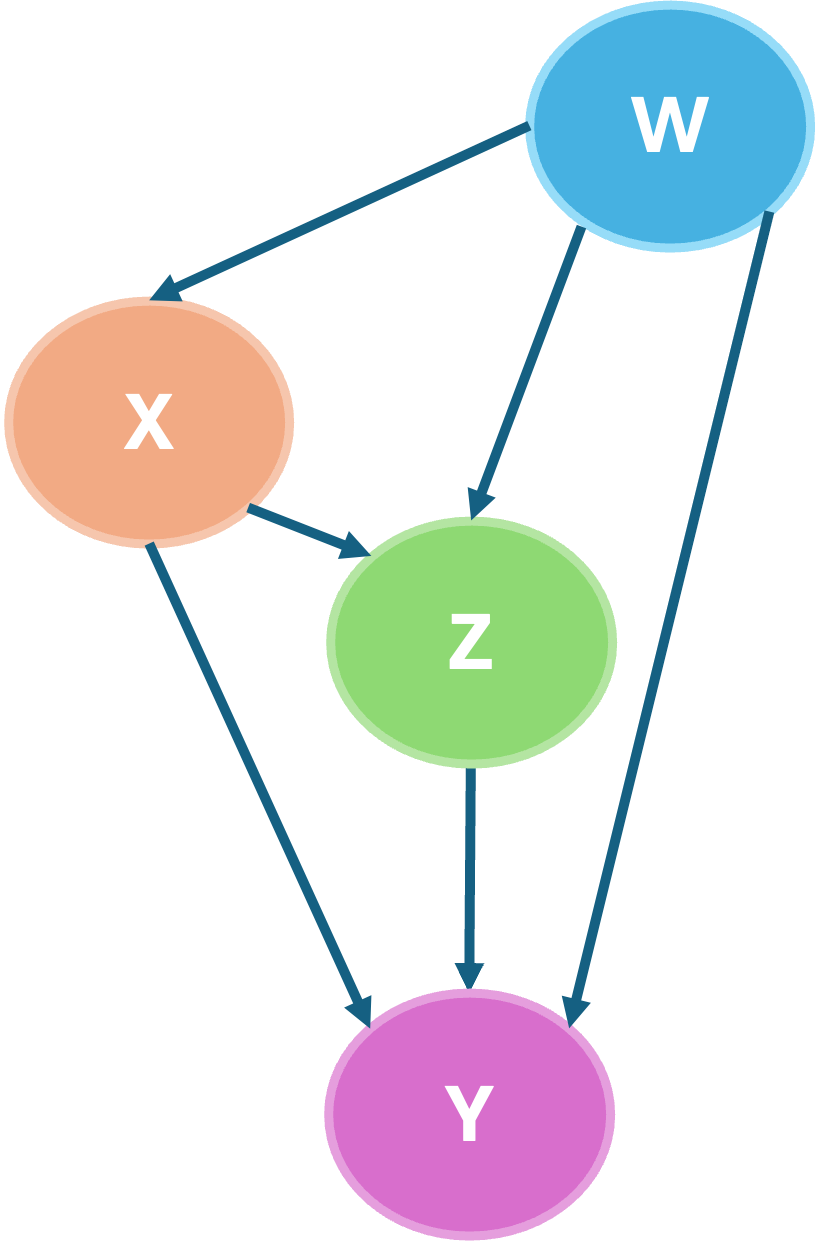}
\end{minipage} 
}  & \multicolumn{1}{c|}{
\begin{minipage}{.12\textwidth}
\centering
    \includegraphics[width=\linewidth]{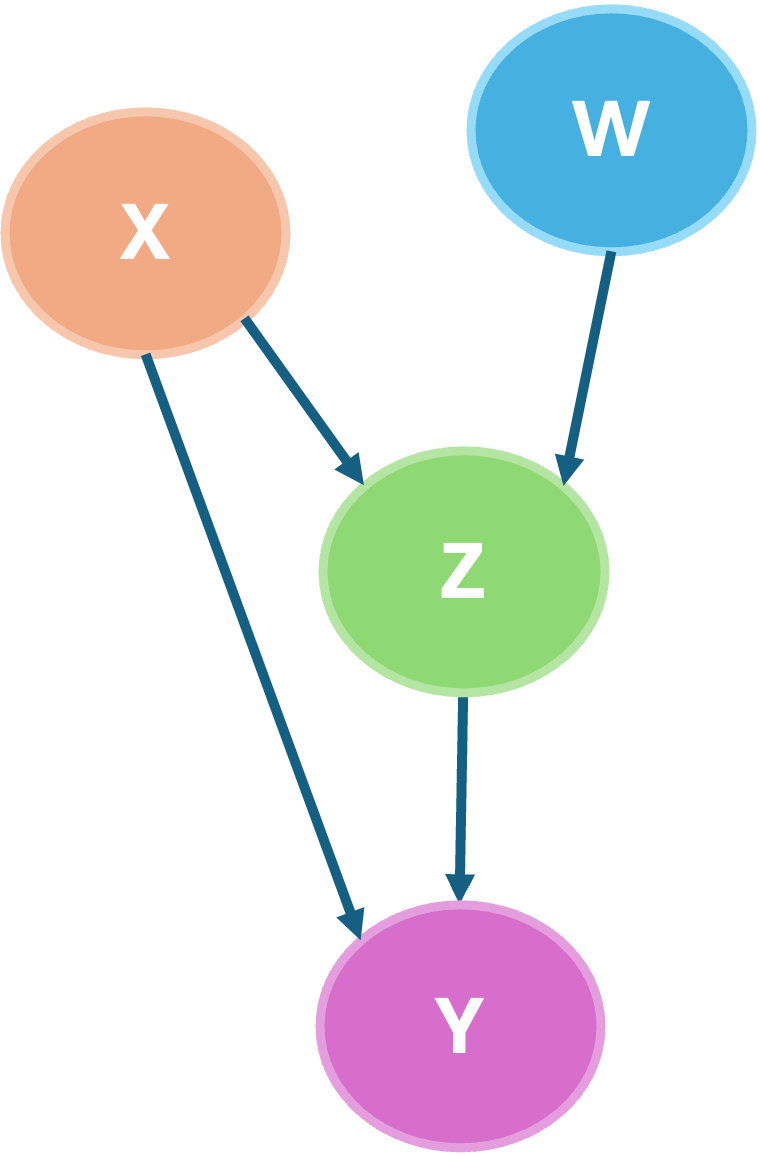}
\end{minipage} 
}  & \multicolumn{1}{c|}{
\begin{minipage}{.12\textwidth}
\centering
    \includegraphics[width=\linewidth]{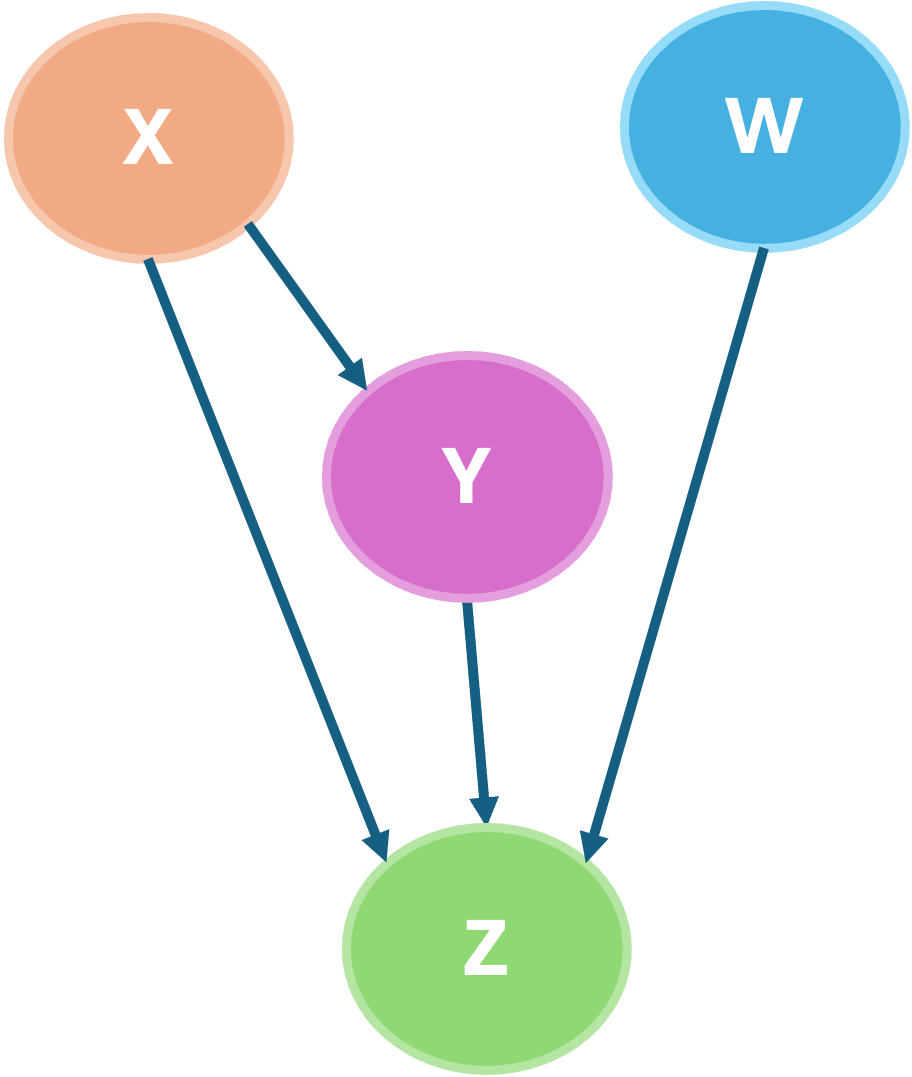}
\end{minipage} 
}  & 
\begin{minipage}{.12\textwidth}
\centering
    \includegraphics[width=\linewidth]{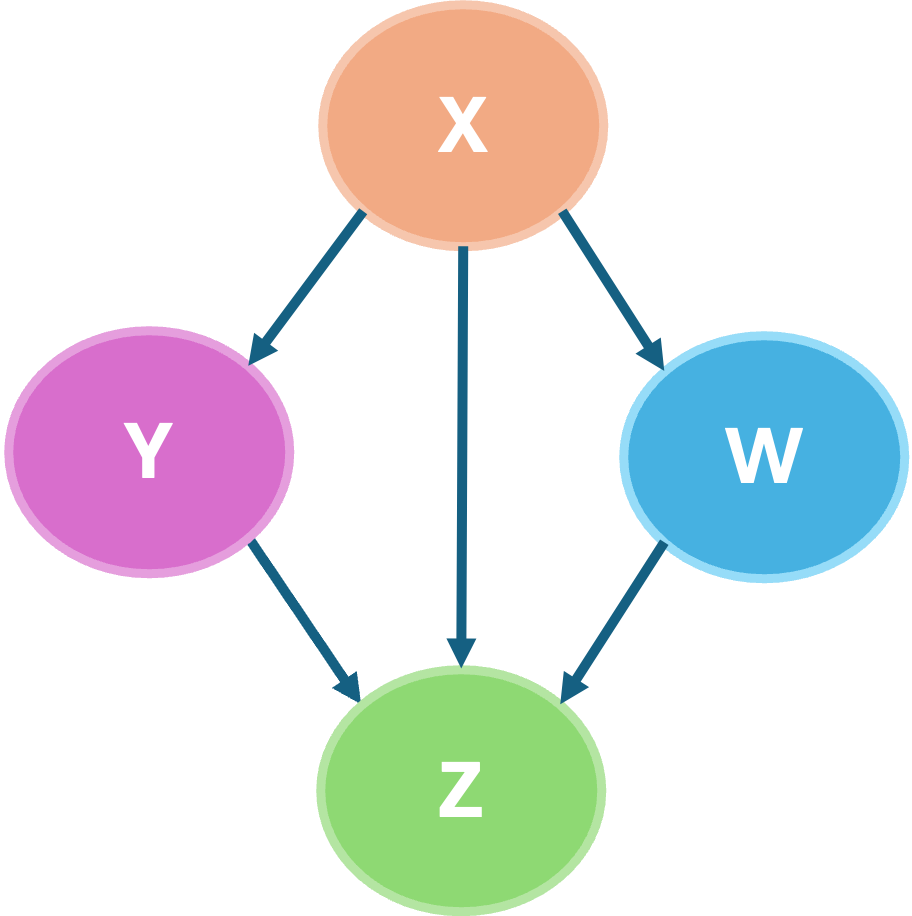}
\end{minipage} 
\\ \hline
\textbf{Precision}            & \textbf{1.0}    & \multicolumn{1}{c|}{0.5000} & \multicolumn{1}{c|}{0.3333} & \multicolumn{1}{c|}{0.2500} & \multicolumn{1}{c|}{0.5000} & 0.4000
\\ \hline
\textbf{Recall}            & \textbf{1.0}    & \multicolumn{1}{c|}{1.0} & \multicolumn{1}{c|}{0.6667} & \multicolumn{1}{c|}{0.3333} & \multicolumn{1}{c|}{0.6667} & 0.6667
\\ \hline
\textbf{F1}            & \textbf{1.0}    & \multicolumn{1}{c|}{0.6667} & \multicolumn{1}{c|}{0.4444} & \multicolumn{1}{c|}{0.2857} & \multicolumn{1}{c|}{0.5714} & 0.5000
\\ \hline
\textbf{SHD}            & \textbf{0}    & \multicolumn{1}{c|}{3} & \multicolumn{1}{c|}{5} & \multicolumn{1}{c|}{5} & \multicolumn{1}{c|}{3} & 4
\\ \hline
\textbf{Count}            & \textbf{10}    & \multicolumn{1}{c|}{4} & \multicolumn{1}{c|}{3} & \multicolumn{1}{c|}{1} & \multicolumn{1}{c|}{1} & 1
\end{tabular}
}

\caption{Comparison of MXMap and RESIT-MLP on 10 randomly sampled segments from a 4-species chain system $w \Rightarrow x \Rightarrow y \Rightarrow z$.}

\label{tab:PredCons}

\end{table}

To further assess prediction consistency, we apply the MXMap and RESIT (recapitulation in Appendix~\ref{appsec:resit}) frameworks to different segments of the sequences to determine the causal order of variables. Specifically, we generate 10 random positive integers as starting timestamps to sample 10 test sequences from the 4-species chain system (Eq.~\ref{eq:4var}), each with a length of 3500.

For MXMap, we select a $k$-nearest neighbor ($k$NN) size of 10, a PCM correlation ratio threshold of 0.6, and delay embedding parameters $\tau = 2$ and $dim = 6$. For RESIT, we used the scikit-learn MLP regressor with two layers of 32 units each, along with an HSIC threshold ($\alpha = 0.01$) for edge removal. 

Table~\ref{tab:PredCons} shows the results (evaluation metrics listed in Appendix~\ref{appsec:metrics}). MXMap consistently determined the correct causal order across all sampled segments, yielding stable predictions with perfect precision, recall, and F1 scores. In contrast, RESIT-MLP's predictions varied significantly depending on the starting point of each sequence. The causal graph predicted by RESIT changed across different segments, illustrating the sensitivity of its prediction to initial conditions due to the underlying predictive model assumption. Additionally, MXMap achieved better evaluation metrics across all four selected metrics.

\subsection{Comparison with Other Established Causal Inference Methods}

To demonstrate the effectiveness of MXMap in multivariate causal inference for nonlinear dynamical systems, we compare its performance on simulated multivariate dynamical systems (including cycles) with baseline methods (tsFCI, VAR-LiNGAM, PCMCI, Granger Causality, DYNOTEARS, SLARAC). When interpreting the predicted outputs, we consider non-oriented causal edges and bidirectional causal edges equivalent, which is in turn reflected in the metric calculation.

\subsubsection{Simulated Systems}
\label{sec:baseline_sim}

\begin{table}[htb]
\makebox[\linewidth]{%

\resizebox{0.95\textwidth}{!}{

\begin{tabular}{l|cc|cc|cc|cc}
\multicolumn{1}{c|}{\multirow{2}{*}{Structure}} & \multicolumn{2}{c|}{3V Chain}                             & \multicolumn{2}{c|}{3V Immorality}                        & \multicolumn{2}{c|}{3V No Cycle}                          & \multicolumn{2}{c}{3V Cycle}                             \\ \cline{2-9} 
\multicolumn{1}{c|}{}                           & \multicolumn{1}{l}{No Noise} & \multicolumn{1}{l|}{Noise} & \multicolumn{1}{l}{No Noise} & \multicolumn{1}{l|}{Noise} & \multicolumn{1}{l}{No Noise} & \multicolumn{1}{l|}{Noise} & \multicolumn{1}{l}{No Noise} & \multicolumn{1}{l}{Noise} \\ \hline
tsFCI                                           & 4                            & 3                          & 2                            & 2                          & 6                            & 4                          & 3                            & 5                         \\
VARLiNGAM                                       & 2                            & 4                          & 4                            & 4                          & 5                            & 6                          & 3                            & 4                         \\
Granger                                         & 4                            & 4                          & 4                            & 4                          & 5                            & 6                          & 2                            & 4                         \\
PCMCI                                           & \textbf{1}                   & \textbf{0}                 & \textbf{0}                   & \textbf{0}                 & 2                            & 1                          & 3                            & 3                         \\
DYNOTEARS                                       & 3                            & 4                          & 2                            & 2                          & 3                            & 1                          & 5                            & 3                         \\
SLARAC                                          & 6                            & 6                          & 6                            & 6                          & 5                            & 6                          & 5                            & 4                         \\ \hline
MXMap                                           & \textbf{1}                   & \textbf{0}                 & 1                            & \textbf{0}                 & \textbf{0}                   & \textbf{0}                 & \textbf{0}                   & \textbf{0}               
\end{tabular}
} }
\caption{SHD scores of MXMap and baselines for 3V settings on simulated no-noise and noisy dynamical systems.}
\label{tab:mxmap-sim-3V}
\end{table}

\begin{table}[htb]
\makebox[\linewidth]{%

\resizebox{0.72\textwidth}{!}{

\begin{tabular}{l|cc|cc|cc}
\multicolumn{1}{c|}{\multirow{2}{*}{Structure}} & \multicolumn{2}{c|}{4V Chain}                             & \multicolumn{2}{c|}{4V No Cycle}                          & \multicolumn{2}{c}{4V Cycle}                             \\ \cline{2-7} 
\multicolumn{1}{c|}{}                           & \multicolumn{1}{l}{No Noise} & \multicolumn{1}{l|}{Noise} & \multicolumn{1}{l}{No Noise} & \multicolumn{1}{l|}{Noise} & \multicolumn{1}{l}{No Noise} & \multicolumn{1}{l}{Noise} \\ \hline
tsFCI                                           & 3                            & 3                          & 5                            & 4                          & 4                            & 6                         \\
VARLiNGAM                                       & 4                            & 6                          & 7                            & 8                          & 4                            & 5                         \\
Granger                                         & 4                            & 6                          & 5                            & 8                          & 3                            & 6                         \\
PCMCI                                           & 1                            & \textbf{0}                 & \textbf{1}                   & \textbf{1}                 & 8                            & 4                         \\
DYNOTEARS                                       & 7                            & 10                         & 3                            & \textbf{1}                 & 7                            & 6                         \\
SLARAC                                          & 10                           & 10                         & 3                            & 2                          & 5                            & 6                         \\ \hline
MXMap                                           & \textbf{0}                   & \textbf{0}                 & \textbf{1}                   & \textbf{1}                 & \textbf{0}                   & \textbf{2}               
\end{tabular}
} }
\caption{SHD scores of MXMap and baselines for 4V settings on simulated no-noise and noisy dynamical systems.}
\label{tab:mxmap-sim-4V}
\end{table}

\begin{table}[htb]
\makebox[\linewidth]{%

\resizebox{0.95\textwidth}{!}{

\begin{tabular}{l|cc|cc|cc|cc}
\multicolumn{1}{c|}{\multirow{2}{*}{Structure}} & \multicolumn{2}{c|}{5V  No Cycle}                         & \multicolumn{2}{c|}{5V  Cycle}                            & \multicolumn{2}{c|}{6V  No Cycle}                         & \multicolumn{2}{c}{7V  Cycle}                            \\ \cline{2-9} 
\multicolumn{1}{c|}{}                           & \multicolumn{1}{l}{No Noise} & \multicolumn{1}{l|}{Noise} & \multicolumn{1}{l}{No Noise} & \multicolumn{1}{l|}{Noise} & \multicolumn{1}{l}{No Noise} & \multicolumn{1}{l|}{Noise} & \multicolumn{1}{l}{No Noise} & \multicolumn{1}{l}{Noise} \\ \hline
tsFCI                                           & 7                            & 4                          & 6                            & 8                          & 10                           & 11                         & 10                           & 10                        \\
VARLiNGAM                                       & 6                            & 6                          & 12                           & 12                         & 9                            & 11                         & 16                           & 15                        \\
Granger                                         & 6                            & 6                          & 12                           & 12                         & 11                           & 12                         & 15                           & 15                        \\
PCMCI                                           & 5                            & 5                          & 5                            & \textbf{2}                 & 11                           & \textbf{3}                 & 11                           & 10                        \\
DYNOTEARS                                       & 8                            & 15                         & 11                           & 12                         & 19                           & 18                         & 16                           & 22                        \\
SLARAC                                          & 16                           & 16                         & 18                           & 18                         & 25                           & 27                         & 21                           & 25                        \\ \hline
MXMap                                           & \textbf{1}                   & \textbf{1}                 & \textbf{0}                   & \textbf{2}                 & \textbf{2}                   & 4                          & \textbf{4}                   & \textbf{6}               
\end{tabular}
} }
\caption{SHD scores of MXMap and baselines for 5V-7V settings on simulated no-noise and noisy dynamical systems.}
\label{tab:mxmap-sim-5-7V}
\end{table}

Tables \ref{tab:mxmap-sim-3V}, \ref{tab:mxmap-sim-4V} and \ref{tab:mxmap-sim-5-7V} show the results of SHD scores (best in bold) on simulated systems with varying complexity from 3 to 7 variables. A more complete evaluation with all four metrics ($Prec$, $Rec$, $F1$, and $SHD$), along with visualizations of ground truth graphs, predicted causal graphs, is provided in Appendix~\ref{appsec:complet}. The time series data are generated under both noise-free and noisy settings (Gaussian additive noise, strength 0.01). 
Overall, MXMap consistently achieves good performance the baselines, yielding lower SHD scores which indicate fewer incorrect edges in the predicted causal graphs.

\subsubsection{ERA5 3-Variable Chain: $tcw \Rightarrow rad \Rightarrow  T_{2m}$}
\label{sec:weather_chain}

\begin{table}[hbt]
\centering
\resizebox{0.7\textwidth}{!}{
\begin{tabular}{c|cccc|c}
Method & \multicolumn{1}{c}{PC} & \multicolumn{1}{c}{FCI} & \multicolumn{1}{c}{LiNGAM} & \multicolumn{1}{c|}{PCMCI}& \multicolumn{1}{c}{MXMap} \\ \hline
Output & \begin{minipage}{.06\linewidth} \centering \includegraphics[width=\linewidth]{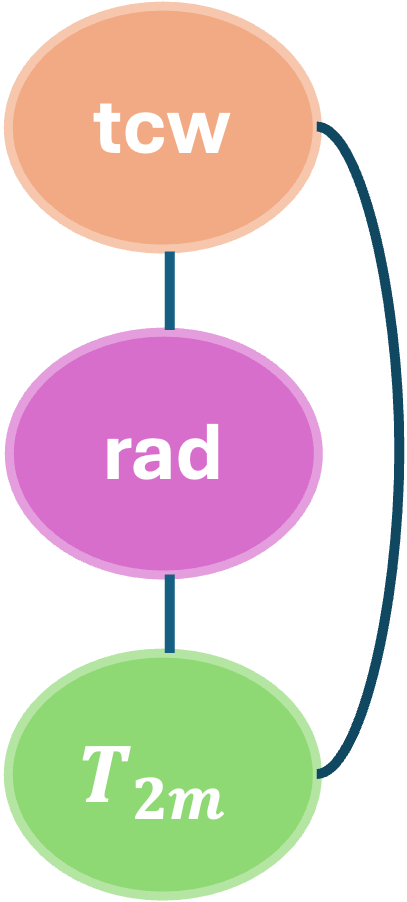} \end{minipage} & \begin{minipage}{.06\linewidth} \centering \includegraphics[width=\linewidth]{imgs/ERA5/pc-fci-era5.png} \end{minipage} & \begin{minipage}{.12\linewidth} \centering \includegraphics[width=\linewidth]{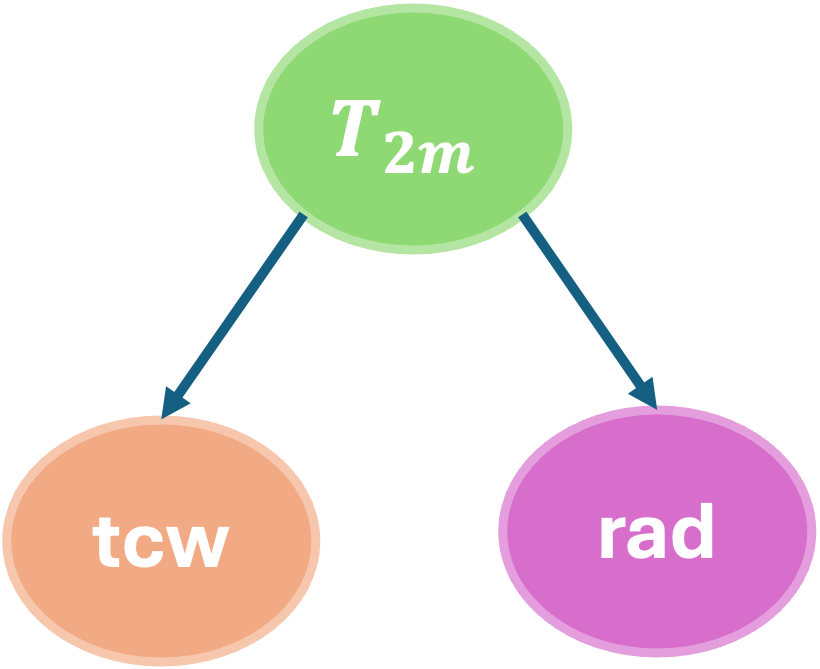} \end{minipage} & \begin{minipage}{.06\linewidth} \centering \includegraphics[width=\linewidth]{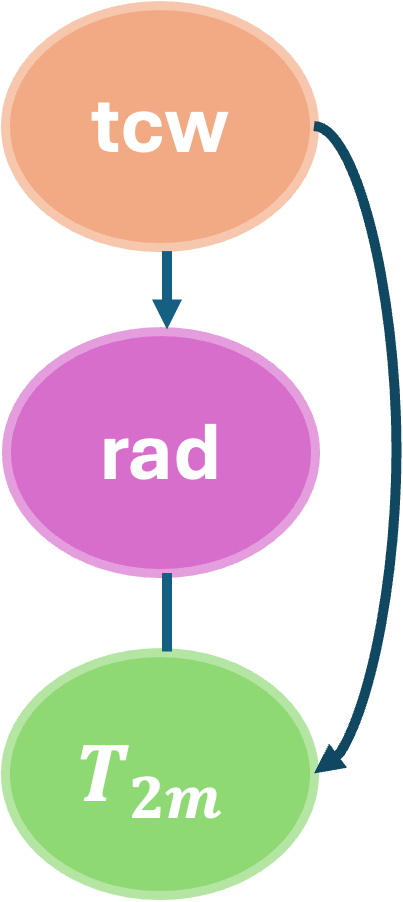} \end{minipage} & \begin{minipage}{.06\linewidth} \centering \includegraphics[width=\linewidth]{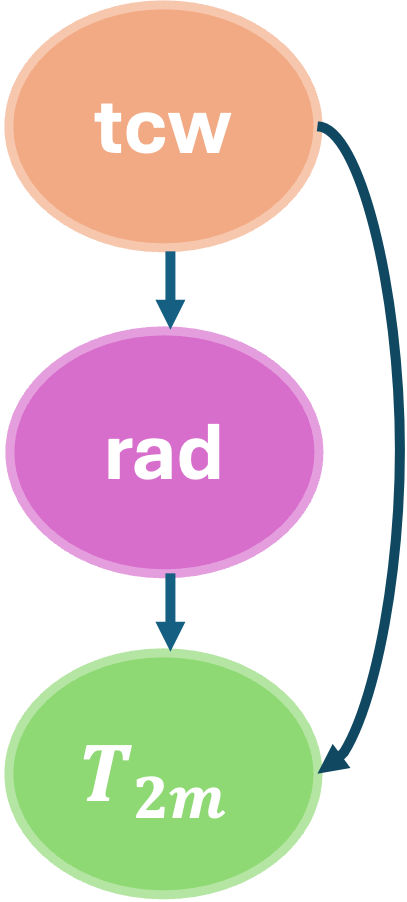} \end{minipage}  \\ \hline
$Prec$ & 0.33 & 0.33  & 0  & 0.50  & \textbf{0.67} \\ \hline
$Rec$ & \textbf{1.0} & \textbf{1.0}  & 0  & \textbf{1.0}  & \textbf{1.0} \\ \hline
$F1$ & 0.50  & 0.50  & 0  & 0.67  & \textbf{0.80} \\ \hline
$SHD$   & 4  & 4  & 4  & 2  & \textbf{1}                        
\end{tabular}
}
\caption{Causal inference methods on the ERA5 3V system.}
\label{tab:mxmap-era5}
\end{table}

For inferring the chain $tcw \Rightarrow rad \Rightarrow T_{2m}$, we take input sequence length of 6000, and consider the lag $\tau$ value to be 4 (for the lag value of delay embedding formulation, and max lag of the PCMCI method) While other methods either failed to identify causal directions correctly or predicted incorrect causal orders, MXMap consistently maintained the correct causal order and produced results closest to the expected ground truth.

\subsubsection{ERA5 5-Variable System: $tcw$, $rad_{solar}$, $rad_{terr}$, $T_{adv950}$ and $T_{2m}$}
\label{sec:era5_5V_eval}

\begin{table}[htb]
\centering
\resizebox{1\textwidth}{!}{
\begin{tabular}{l|llllll|l}
Methods                          & tsFCI                            & VARLiNGAM                        & Granger                          & PCMCI                            & DYNOTEARS                        & SLARAC                           & MXMap      \\ \hline
$rad_{solar} \Rightarrow T_{2m}$ & \cmark                       & \halfcheckmark & \halfcheckmark & \halfcheckmark & \halfcheckmark & \xmark                           & \cmark \\
$rad_{terr} \Rightarrow T_{2m}$  & \halfcheckmark & \halfcheckmark & \cmark                       & \halfcheckmark & \xmark                           & \cmark                       & \cmark \\
$T_{adv950} \Rightarrow T_{2m}$  & \xmark                           & \halfcheckmark & \halfcheckmark & \halfcheckmark & \cmark                       & \halfcheckmark & \cmark \\
$tcw\Rightarrow rad_{solar}$     & \xmark                           & \xmark                           & \halfcheckmark & \cmark                       & \cmark                       & \cmark                       & \cmark
\end{tabular}
}
\caption{Detection of Benchmark Causal Relationships in the ERA5 5V System (full visualizations in Table~\ref{apptab:mxmap-era5-5V}). A checkmark (green) indicates a correctly detected and oriented edge, a half-checkmark (gray) denotes a detected but ambiguously oriented edge, and a crossmark (red) represents an undetected or incorrectly oriented edge.}
\label{tab:mxmap-era5-5V}
\end{table}

The objective is to evaluate the performance via observing how many of these well-established causal relationships are detected by the methods $rad_{solar} \Rightarrow T_{2m}$, $rad_{terr} \Rightarrow T_{2m}$, $T_{adv950} \Rightarrow T_{2m}$, $tcw\Rightarrow rad_{solar}$ (explanations of these mechanisms in Appendix~\ref{appsec:era5}, full table with predicted graphs in Table~\ref{apptab:mxmap-era5-5V} in Appendix~\ref{appsec:era5_eval}). The results in Table~\ref{tab:mxmap-era5-5V} show that MXMap is overall able to correctly identify and orient edges that represent the 4 benchmark causal mechanisms in the 5-variable system, and outperforms the other baseline methods.
\section{Conclusion}

In this work, we proposed multiPCM, allowing us to more effectively distinguish between direct and indirect causal relationships. We integrated multiPCM with bivariate Convergent Cross Mapping (CCM) in a two-phase framework, MXMap, that first establishes an initial causal graph and then prunes indirect connections. Through experiments on simulated species interaction systems and real-world ERA5 meteorological data, we demonstrated that MXMap outperforms traditional methods and exhibits robust performance in complex, high-dimensional dynamical systems.

There are still limits to this current framework (discussed in Appendix~\ref{appsec:mxmap_limits}), demonstrated in runtime complexity, scalability, and possible failure cases in highly noisy environment or non-stationary systems. Future work could focus on enhancing the robustness of cross mapping under noisy conditions~\citep{monster2017causal}. Investigating and incorporating certain noise-handling mechanisms~\citep{zhang2024enhancing} in MXMap could further enhance its applicability in noisy real-world scenarios. Another direction is to explore adaptive parameter selection~\citep{shortreed2017outcome, machlanski2023hyperparameter} for MXMap, such as optimizing embedding dimensions and lags based on data properties and currently outputs. Current grid search methods are computationally expensive for larger datasets, and efficient heuristic or learning-based tuning could improve scalability. Finally, applying MXMap to other real-world domains, such as power systems, larger timescale climate modeling, and epidemiology, could further validate its versatility and reveal complex causal interactions.

\acks{This work is supported by Mitacs Accelerate Research Fellowship in collaboration with Hydro-Québec Research Institute (IREQ).}

\bibliography{ref}

\begin{thebibliography}{50}
\providecommand{\natexlab}[1]{#1}
\providecommand{\url}[1]{\texttt{#1}}
\expandafter\ifx\csname urlstyle\endcsname\relax
  \providecommand{\doi}[1]{doi: #1}\else
  \providecommand{\doi}{doi: \begingroup \urlstyle{rm}\Url}\fi

\bibitem[Barraquand et~al.(2021)Barraquand, Picoche, Detto, and Hartig]{barraquand2021inferring}
Fr{\'e}d{\'e}ric Barraquand, Coralie Picoche, Matteo Detto, and Florian Hartig.
\newblock Inferring species interactions using granger causality and convergent cross mapping.
\newblock \emph{Theoretical Ecology}, 14\penalty0 (1):\penalty0 87--105, 2021.

\bibitem[Beaumont et~al.(2021)Beaumont, Horsburgh, Pilgerstorfer, Droth, Oentaryo, Ler, Nguyen, Ferreira, Patel, and Leong]{Beaumont_CausalNex_2021}
Paul Beaumont, Ben Horsburgh, Philip Pilgerstorfer, Angel Droth, Richard Oentaryo, Steven Ler, Hiep Nguyen, Gabriel~Azevedo Ferreira, Zain Patel, and Wesley Leong.
\newblock {CausalNex}, October 2021.
\newblock URL \url{https://github.com/quantumblacklabs/causalnex}.

\bibitem[Buitinck et~al.(2013)Buitinck, Louppe, Blondel, Pedregosa, Mueller, Grisel, Niculae, Prettenhofer, Gramfort, Grobler, Layton, VanderPlas, Joly, Holt, and Varoquaux]{sklearn_api}
Lars Buitinck, Gilles Louppe, Mathieu Blondel, Fabian Pedregosa, Andreas Mueller, Olivier Grisel, Vlad Niculae, Peter Prettenhofer, Alexandre Gramfort, Jaques Grobler, Robert Layton, Jake VanderPlas, Arnaud Joly, Brian Holt, and Ga{\"{e}}l Varoquaux.
\newblock {API} design for machine learning software: experiences from the scikit-learn project.
\newblock In \emph{ECML PKDD Workshop: Languages for Data Mining and Machine Learning}, pages 108--122, 2013.

\bibitem[Butler et~al.(2023)Butler, Feng, and Djuri{\'c}]{butler2023causal}
Kurt Butler, Guanchao Feng, and Petar~M Djuri{\'c}.
\newblock On causal discovery with convergent cross mapping.
\newblock \emph{IEEE Transactions on Signal Processing}, 2023.

\bibitem[Camps-Valls et~al.(2023)Camps-Valls, Gerhardus, Ninad, Varando, Martius, Balaguer-Ballester, Vinuesa, Diaz, Zanna, and Runge]{camps2023discovering}
Gustau Camps-Valls, Andreas Gerhardus, Urmi Ninad, Gherardo Varando, Georg Martius, Emili Balaguer-Ballester, Ricardo Vinuesa, Emiliano Diaz, Laure Zanna, and Jakob Runge.
\newblock Discovering causal relations and equations from data.
\newblock \emph{Physics Reports}, 1044:\penalty0 1--68, 2023.

\bibitem[Chen et~al.(2022)Chen, Xu, Gao, Sugihara, Shen, Cai, Li, Wu, Yang, Yao, et~al.]{chen2022causation}
Ziyue Chen, Miaoqing Xu, Bingbo Gao, George Sugihara, Feixue Shen, Yanyan Cai, Anqi Li, Qi~Wu, Lin Yang, Qi~Yao, et~al.
\newblock Causation inference in complicated atmospheric environment.
\newblock \emph{Environmental Pollution}, 303:\penalty0 119057, 2022.

\bibitem[Cummins et~al.(2015)Cummins, Gedeon, and Spendlove]{cummins2015efficacy}
Bree Cummins, Tom{\'a}s Gedeon, and Kelly Spendlove.
\newblock On the efficacy of state space reconstruction methods in determining causality.
\newblock \emph{SIAM Journal on Applied Dynamical Systems}, 14\penalty0 (1):\penalty0 335--381, 2015.

\bibitem[Entner and Hoyer(2010)]{entner2010causal}
Doris Entner and Patrik~O Hoyer.
\newblock On causal discovery from time series data using fci.
\newblock \emph{Probabilistic graphical models}, 16, 2010.

\bibitem[Fraser and Swinney(1986)]{fraser1986independent}
Andrew~M Fraser and Harry~L Swinney.
\newblock Independent coordinates for strange attractors from mutual information.
\newblock \emph{Physical review A}, 33\penalty0 (2):\penalty0 1134, 1986.

\bibitem[Granger(1969)]{granger1969investigating}
Clive~WJ Granger.
\newblock Investigating causal relations by econometric models and cross-spectral methods.
\newblock \emph{Econometrica: journal of the Econometric Society}, pages 424--438, 1969.

\bibitem[Hersbach et~al.(2020)Hersbach, Bell, Berrisford, Hirahara, Hor{\'a}nyi, Mu{\~n}oz-Sabater, Nicolas, Peubey, Radu, Schepers, et~al.]{hersbach2020era5}
Hans Hersbach, Bill Bell, Paul Berrisford, Shoji Hirahara, Andr{\'a}s Hor{\'a}nyi, Joaqu{\'\i}n Mu{\~n}oz-Sabater, Julien Nicolas, Carole Peubey, Raluca Radu, Dinand Schepers, et~al.
\newblock The era5 global reanalysis.
\newblock \emph{Quarterly Journal of the Royal Meteorological Society}, 146\penalty0 (730):\penalty0 1999--2049, 2020.

\bibitem[Hoyer et~al.(2008)Hoyer, Janzing, Mooij, Peters, and Sch{\"o}lkopf]{hoyer2008nonlinear}
Patrik Hoyer, Dominik Janzing, Joris~M Mooij, Jonas Peters, and Bernhard Sch{\"o}lkopf.
\newblock Nonlinear causal discovery with additive noise models.
\newblock \emph{Advances in neural information processing systems}, 21, 2008.

\bibitem[Hyv{\"a}rinen et~al.(2010)Hyv{\"a}rinen, Zhang, Shimizu, and Hoyer]{hyvarinen2010estimation}
Aapo Hyv{\"a}rinen, Kun Zhang, Shohei Shimizu, and Patrik~O Hoyer.
\newblock Estimation of a structural vector autoregression model using non-gaussianity.
\newblock \emph{Journal of Machine Learning Research}, 11\penalty0 (5), 2010.

\bibitem[Jiang et~al.(2023)Jiang, Jiang, Wang, Pan, and Zhong]{jiang2023partial}
Qingchao Jiang, Jiashi Jiang, Wenjing Wang, Chunjian Pan, and Weimin Zhong.
\newblock Partial cross mapping based on sparse variable selection for direct fault root cause diagnosis for industrial processes.
\newblock \emph{IEEE Transactions on Neural Networks and Learning Systems}, 2023.

\bibitem[Kennel et~al.(1992)Kennel, Brown, and Abarbanel]{kennel1992determining}
Matthew~B Kennel, Reggie Brown, and Henry~DI Abarbanel.
\newblock Determining embedding dimension for phase-space reconstruction using a geometrical construction.
\newblock \emph{Physical review A}, 45\penalty0 (6):\penalty0 3403, 1992.

\bibitem[Keropyan et~al.(2023)Keropyan, Strieder, and Drton]{keropyan2023rank}
Grigor Keropyan, David Strieder, and Mathias Drton.
\newblock Rank-based causal discovery for post-nonlinear models.
\newblock In \emph{International Conference on Artificial Intelligence and Statistics}, pages 7849--7870. PMLR, 2023.

\bibitem[Klikov{\'a} and Raidl(2011)]{klikova2011reconstruction}
B~Klikov{\'a} and Ale{\v{s}} Raidl.
\newblock Reconstruction of phase space of dynamical systems using method of time delay.
\newblock In \emph{Proceedings of WDS}, volume~11, pages 83--87, 2011.

\bibitem[Kugiumtzis(1996)]{kugiumtzis1996state}
Dimitris Kugiumtzis.
\newblock State space reconstruction parameters in the analysis of chaotic time series—the role of the time window length.
\newblock \emph{Physica D: Nonlinear Phenomena}, 95\penalty0 (1):\penalty0 13--28, 1996.

\bibitem[Leng et~al.(2020)Leng, Ma, Kurths, Lai, Lin, Aihara, and Chen]{leng2020partial}
Siyang Leng, Huanfei Ma, J{\"u}rgen Kurths, Ying-Cheng Lai, Wei Lin, Kazuyuki Aihara, and Luonan Chen.
\newblock Partial cross mapping eliminates indirect causal influences.
\newblock \emph{Nature communications}, 11\penalty0 (1):\penalty0 2632, 2020.

\bibitem[Liu et~al.(2024)Liu, Huang, Gao, Ke, Bondell, and Gong]{liu2024causal}
Wenqin Liu, Biwei Huang, Erdun Gao, Qiuhong Ke, Howard Bondell, and Mingming Gong.
\newblock Causal discovery with mixed linear and nonlinear additive noise models: A scalable approach.
\newblock In \emph{Causal Learning and Reasoning}, pages 1237--1263. PMLR, 2024.

\bibitem[Lotka(1925)]{lotka1925elements}
AJ~Lotka.
\newblock Elements of physical biology.
\newblock \emph{Williams and Wilkins}, 1925.

\bibitem[Machlanski et~al.(2023)Machlanski, Samothrakis, and Clarke]{machlanski2023hyperparameter}
Damian Machlanski, Spyridon Samothrakis, and Paul Clarke.
\newblock Hyperparameter tuning and model evaluation in causal effect estimation.
\newblock \emph{arXiv preprint arXiv:2303.01412}, 2023.

\bibitem[Milnor(1985)]{milnor1985concept}
John Milnor.
\newblock On the concept of attractor.
\newblock \emph{Communications in Mathematical Physics}, 99:\penalty0 177--195, 1985.

\bibitem[M{\o}nster et~al.(2017)M{\o}nster, Fusaroli, Tyl{\'e}n, Roepstorff, and Sherson]{monster2017causal}
Dan M{\o}nster, Riccardo Fusaroli, Kristian Tyl{\'e}n, Andreas Roepstorff, and Jacob~F Sherson.
\newblock Causal inference from noisy time-series data—testing the convergent cross-mapping algorithm in the presence of noise and external influence.
\newblock \emph{Future Generation Computer Systems}, 73:\penalty0 52--62, 2017.

\bibitem[Pamfil et~al.(2020)Pamfil, Sriwattanaworachai, Desai, Pilgerstorfer, Georgatzis, Beaumont, and Aragam]{pamfil2020dynotears}
Roxana Pamfil, Nisara Sriwattanaworachai, Shaan Desai, Philip Pilgerstorfer, Konstantinos Georgatzis, Paul Beaumont, and Bryon Aragam.
\newblock Dynotears: Structure learning from time-series data.
\newblock In \emph{International Conference on Artificial Intelligence and Statistics}, pages 1595--1605. Pmlr, 2020.

\bibitem[Pedregosa et~al.(2011)Pedregosa, Varoquaux, Gramfort, Michel, Thirion, Grisel, Blondel, Prettenhofer, Weiss, Dubourg, Vanderplas, Passos, Cournapeau, Brucher, Perrot, and Duchesnay]{scikit-learn}
F.~Pedregosa, G.~Varoquaux, A.~Gramfort, V.~Michel, B.~Thirion, O.~Grisel, M.~Blondel, P.~Prettenhofer, R.~Weiss, V.~Dubourg, J.~Vanderplas, A.~Passos, D.~Cournapeau, M.~Brucher, M.~Perrot, and E.~Duchesnay.
\newblock Scikit-learn: Machine learning in {P}ython.
\newblock \emph{Journal of Machine Learning Research}, 12:\penalty0 2825--2830, 2011.

\bibitem[Peters et~al.(2014)Peters, Mooij, Janzing, and Sch{\"o}lkopf]{peters2014causal}
Jonas Peters, Joris~M Mooij, Dominik Janzing, and Bernhard Sch{\"o}lkopf.
\newblock Causal discovery with continuous additive noise models.
\newblock \emph{The Journal of Machine Learning Research}, 15\penalty0 (1):\penalty0 2009--2053, 2014.

\bibitem[Pielke~Sr and Matsui(2005)]{pielke2005should}
Roger~A Pielke~Sr and Toshihisa Matsui.
\newblock Should light wind and windy nights have the same temperature trends at individual levels even if the boundary layer averaged heat content change is the same?
\newblock \emph{Geophysical Research Letters}, 32\penalty0 (21), 2005.

\bibitem[Ramanathan et~al.(1989)Ramanathan, Cess, Harrison, Minnis, Barkstrom, Ahmad, and Hartmann]{ramanathan1989cloud}
VLRD Ramanathan, RD~Cess, EF~Harrison, P~Minnis, BR~Barkstrom, E~Ahmad, and D~Hartmann.
\newblock Cloud-radiative forcing and climate: Results from the earth radiation budget experiment.
\newblock \emph{Science}, 243\penalty0 (4887):\penalty0 57--63, 1989.

\bibitem[Roques and Chekroun(2011)]{roques2011probing}
Lionel Roques and Micka{\"e}l~D Chekroun.
\newblock Probing chaos and biodiversity in a simple competition model.
\newblock \emph{Ecological Complexity}, 8\penalty0 (1):\penalty0 98--104, 2011.

\bibitem[Runge et~al.(2019)Runge, Nowack, Kretschmer, Flaxman, and Sejdinovic]{runge2019detecting}
Jakob Runge, Peer Nowack, Marlene Kretschmer, Seth Flaxman, and Dino Sejdinovic.
\newblock Detecting and quantifying causal associations in large nonlinear time series datasets.
\newblock \emph{Science advances}, 5\penalty0 (11):\penalty0 eaau4996, 2019.

\bibitem[Runge et~al.(2020)Runge, Tibau, Bruhns, Mu{\~n}oz-Mar{\'\i}, and Camps-Valls]{runge2020causality}
Jakob Runge, Xavier-Andoni Tibau, Matthias Bruhns, Jordi Mu{\~n}oz-Mar{\'\i}, and Gustau Camps-Valls.
\newblock The causality for climate competition.
\newblock In \emph{NeurIPS 2019 Competition and Demonstration Track}, pages 110--120. Pmlr, 2020.

\bibitem[Runge et~al.(2023)Runge, Gerhardus, Varando, Eyring, and Camps-Valls]{runge2023causal}
Jakob Runge, Andreas Gerhardus, Gherardo Varando, Veronika Eyring, and Gustau Camps-Valls.
\newblock Causal inference for time series.
\newblock \emph{Nature Reviews Earth \& Environment}, 4\penalty0 (7):\penalty0 487--505, 2023.

\bibitem[Sauer et~al.(1991)Sauer, Yorke, and Casdagli]{sauer1991j}
Tim Sauer, James~A Yorke, and Martin Casdagli.
\newblock J stat phys.
\newblock \emph{Embedology}, 65\penalty0 (3-4):\penalty0 579--616, 1991.

\bibitem[Shimizu(2014)]{shimizu2014lingam}
Shohei Shimizu.
\newblock Lingam: Non-gaussian methods for estimating causal structures.
\newblock \emph{Behaviormetrika}, 41\penalty0 (1):\penalty0 65--98, 2014.

\bibitem[Shimizu et~al.(2006)Shimizu, Hoyer, Hyv{\"a}rinen, Kerminen, and Jordan]{shimizu2006linear}
Shohei Shimizu, Patrik~O Hoyer, Aapo Hyv{\"a}rinen, Antti Kerminen, and Michael Jordan.
\newblock A linear non-gaussian acyclic model for causal discovery.
\newblock \emph{Journal of Machine Learning Research}, 7\penalty0 (10), 2006.

\bibitem[Shortreed and Ertefaie(2017)]{shortreed2017outcome}
Susan~M Shortreed and Ashkan Ertefaie.
\newblock Outcome-adaptive lasso: variable selection for causal inference.
\newblock \emph{Biometrics}, 73\penalty0 (4):\penalty0 1111--1122, 2017.

\bibitem[Spirtes et~al.(2013)Spirtes, Meek, and Richardson]{spirtes2013causal}
Peter~L Spirtes, Christopher Meek, and Thomas~S Richardson.
\newblock Causal inference in the presence of latent variables and selection bias.
\newblock \emph{arXiv preprint arXiv:1302.4983}, 2013.

\bibitem[Stephens(2005)]{stephens2005cloud}
Graeme~L Stephens.
\newblock Cloud feedbacks in the climate system: A critical review.
\newblock \emph{Journal of climate}, 18\penalty0 (2):\penalty0 237--273, 2005.

\bibitem[Sugihara et~al.(2012)Sugihara, May, Ye, Hsieh, Deyle, Fogarty, and Munch]{sugihara2012detecting}
George Sugihara, Robert May, Hao Ye, Chih-hao Hsieh, Ethan Deyle, Michael Fogarty, and Stephan Munch.
\newblock Detecting causality in complex ecosystems.
\newblock \emph{science}, 338\penalty0 (6106):\penalty0 496--500, 2012.

\bibitem[Takens(2006)]{takens2006detecting}
Floris Takens.
\newblock Detecting strange attractors in turbulence.
\newblock In \emph{Dynamical Systems and Turbulence, Warwick 1980: proceedings of a symposium held at the University of Warwick 1979/80}, pages 366--381. Springer, 2006.

\bibitem[Vlachos and Kugiumtzis(2008)]{vlachos2008state}
I~Vlachos and D~Kugiumtzis.
\newblock State space reconstruction for multivariate time series prediction.
\newblock \emph{arXiv preprint arXiv:0809.2220}, 2008.

\bibitem[Vlachos and Kugiumtzis(2010)]{vlachos2010nonuniform}
Ioannis Vlachos and Dimitris Kugiumtzis.
\newblock Nonuniform state-space reconstruction and coupling detection.
\newblock \emph{Physical Review E}, 82\penalty0 (1):\penalty0 016207, 2010.

\bibitem[Volterra(1931)]{volterra1931theorie}
Vito Volterra.
\newblock \emph{Th{\'e}orie math{\'e}matique de la lutte pour la vie}.
\newblock Gauthiers-Villars, 1931.

\bibitem[Weichwald et~al.(2020)Weichwald, Jakobsen, Mogensen, Petersen, Thams, and Varando]{weichwald2020causal}
Sebastian Weichwald, Martin~E Jakobsen, Phillip~B Mogensen, Lasse Petersen, Nikolaj Thams, and Gherardo Varando.
\newblock Causal structure learning from time series: Large regression coefficients may predict causal links better in practice than small p-values.
\newblock In \emph{NeurIPS 2019 Competition and Demonstration Track}, pages 27--36. PMLR, 2020.

\bibitem[Whitney(1936)]{whitney1936differentiable}
Hassler Whitney.
\newblock Differentiable manifolds.
\newblock \emph{Annals of Mathematics}, pages 645--680, 1936.

\bibitem[Zhang et~al.(2024)Zhang, Mirall{\`e}s, Rousseau-Rizzi, Zinflou, Boulet, and Wu]{zhang2024enhancing}
Elise Zhang, Fran{\c{c}}ois Mirall{\`e}s, Rapha{\"e}l Rousseau-Rizzi, Arnaud Zinflou, Benoit Boulet, and Di~Wu.
\newblock Enhancing convergent cross mapping: Simple preprocessing for noise-resilient causal discovery.
\newblock \emph{Authorea Preprints}, 2024.

\bibitem[Zhang et~al.(2015)Zhang, Wang, Zhang, and Sch{\"o}lkopf]{zhang2015estimation}
Kun Zhang, Zhikun Wang, Jiji Zhang, and Bernhard Sch{\"o}lkopf.
\newblock On estimation of functional causal models: general results and application to the post-nonlinear causal model.
\newblock \emph{ACM Transactions on Intelligent Systems and Technology (TIST)}, 7\penalty0 (2):\penalty0 1--22, 2015.

\bibitem[Zheng et~al.(2018)Zheng, Aragam, Ravikumar, and Xing]{zheng2018dags}
Xun Zheng, Bryon Aragam, Pradeep~K Ravikumar, and Eric~P Xing.
\newblock Dags with no tears: Continuous optimization for structure learning.
\newblock \emph{Advances in neural information processing systems}, 31, 2018.

\bibitem[Zheng et~al.(2024)Zheng, Huang, Chen, Ramsey, Gong, Cai, Shimizu, Spirtes, and Zhang]{zheng2024causal}
Yujia Zheng, Biwei Huang, Wei Chen, Joseph Ramsey, Mingming Gong, Ruichu Cai, Shohei Shimizu, Peter Spirtes, and Kun Zhang.
\newblock Causal-learn: Causal discovery in python.
\newblock \emph{Journal of Machine Learning Research}, 25\penalty0 (60):\penalty0 1--8, 2024.

\end{thebibliography}

\newpage
\pagestyle{plain}
\newpage

\appendix

\section{Validation of multiPCM with More Generated Systems}
\label{appsec:valid_multiPCM}

A more complete grid-search results is presented in Table~\ref{tab:app_multiPCM}, with more cases of simulated systems and additional profiles of correlation scores $\rho_{All}$ and $\rho_{Direct}$, on top of the ratio and label. 

Input length is $L = 3500$, the ranges of lags and embedding dimensions are ($\tau, E \in \{1, 2, 3, \ldots, 8\}$). The predicted labels indicating whether direct causality is rejected based on a PCM threshold of 0.45 (this threshold selection is discussed in Appendix~\ref{appsec:thres_multiPCM}). \textcolor{red}{Red} label indicates rejection, suggesting there is only indirect causality, while \textcolor{blue}{blue} label indicates acceptance, suggesting direct causality exists hence we should not remove the link between the colored nodes.

\begin{table}[htb]
\centering

\makebox[\linewidth]{%

\resizebox{1.2\textwidth}{!}{

\begin{tabular}{c|c|cc|cc}
Type         & $Direct$ & \multicolumn{2}{c|}{$Indirect$} & \multicolumn{2}{c}{$Both$} \\ \hline
Causality         &\begin{minipage}{.15\linewidth} \centering \includegraphics[width=0.3\linewidth]{imgs/ValidPCM/4VDirect.png} \end{minipage}& \multicolumn{1}{c|}{\begin{minipage}{.15\linewidth} \centering \includegraphics[width=0.3\linewidth]{imgs/ValidPCM/4VIndirect1.png} \end{minipage}}    & \begin{minipage}{.12\linewidth} \centering \includegraphics[width=\linewidth]{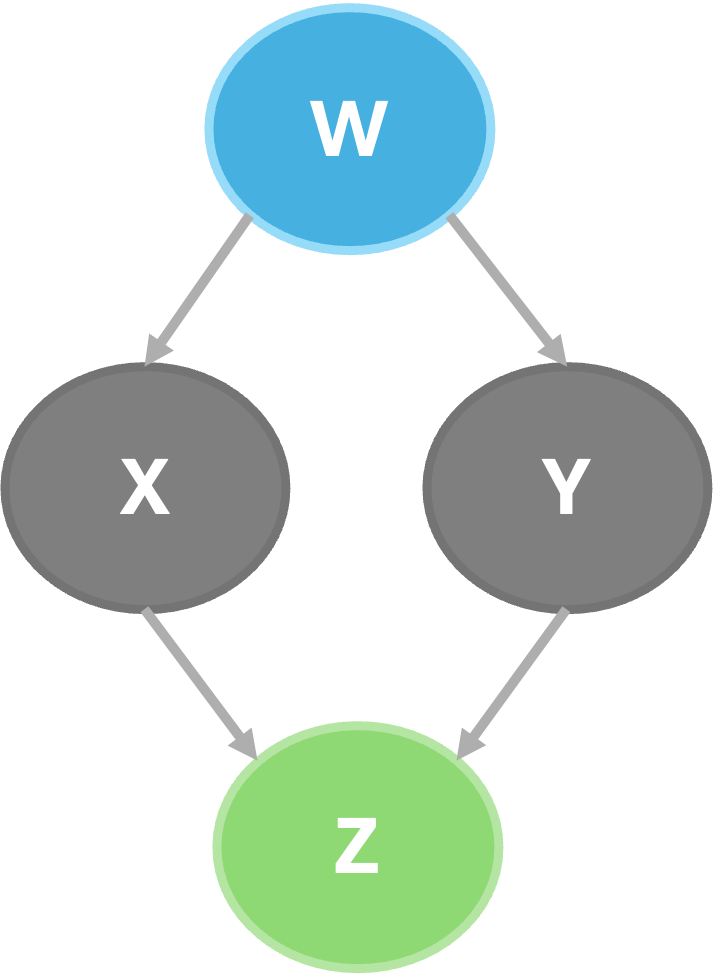} \end{minipage} & \multicolumn{1}{c|}{\begin{minipage}{.15\linewidth} \centering \includegraphics[width=0.5\linewidth]{imgs/ValidPCM/4VBoth1.png} \end{minipage}} & \begin{minipage}{.12\linewidth} \centering \includegraphics[width=\linewidth]{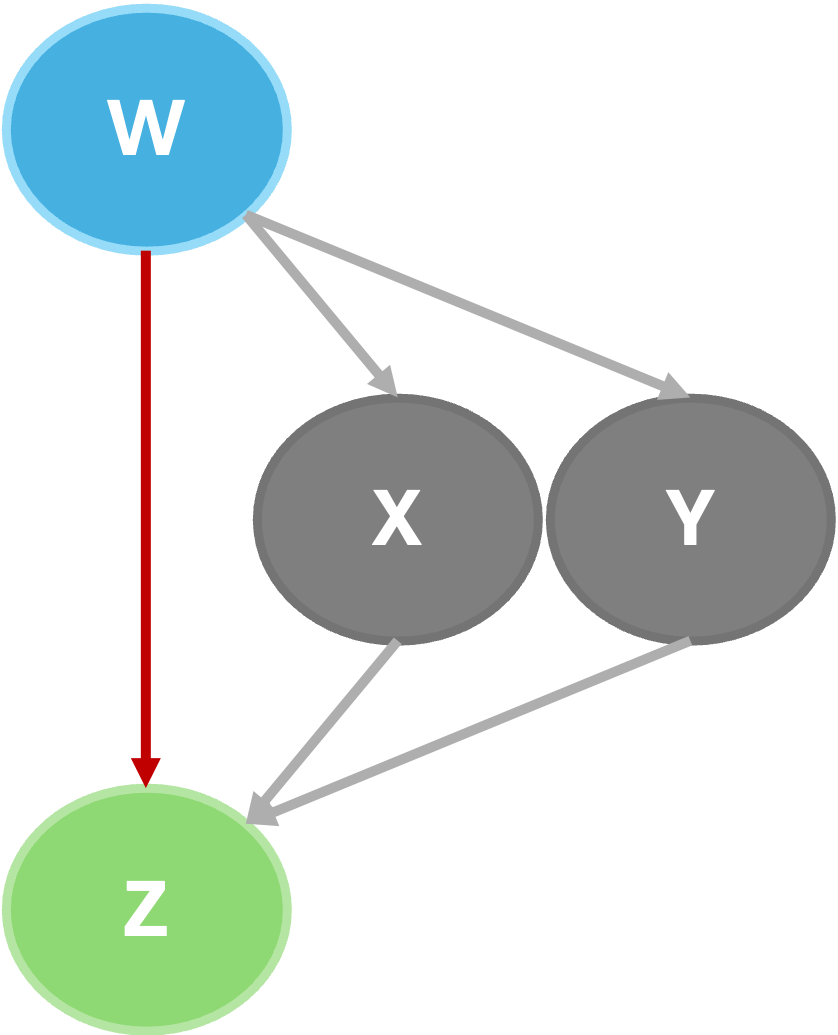} \end{minipage} \\ \hline
$\rho_{All}$    &\begin{minipage}{.153\linewidth} \centering \includegraphics[width=\linewidth]{imgs/ValidPCM/4VDirect_sc1.png} \end{minipage}& \multicolumn{1}{c|}{\begin{minipage}{.153\linewidth} \centering \includegraphics[width=\linewidth]{imgs/ValidPCM/4VIndirect1_sc1.png} \end{minipage}}    & \begin{minipage}{.153\linewidth} \centering \includegraphics[width=\linewidth]{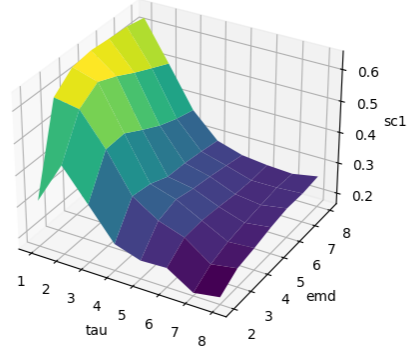} \end{minipage}    & \multicolumn{1}{c|}{\begin{minipage}{.153\linewidth} \centering \includegraphics[width=\linewidth]{imgs/ValidPCM/4VBoth1_sc1.png} \end{minipage}} & \begin{minipage}{.153\linewidth} \centering \includegraphics[width=\linewidth]{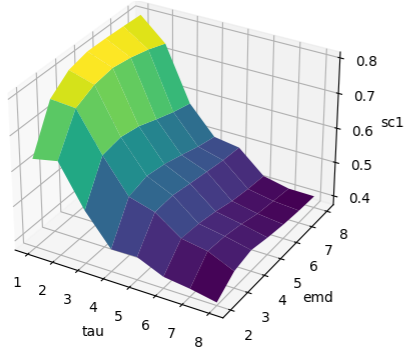} \end{minipage}  \\ \hline
$\rho_{Direct}$ &\begin{minipage}{.153\linewidth} \centering \includegraphics[width=\linewidth]{imgs/ValidPCM/4VDirect_sc2.png} \end{minipage}& \multicolumn{1}{c|}{\begin{minipage}{.153\linewidth} \centering \includegraphics[width=\linewidth]{imgs/ValidPCM/4VIndirect1_sc2.png} \end{minipage}}    &  \begin{minipage}{.153\linewidth} \centering \includegraphics[width=\linewidth]{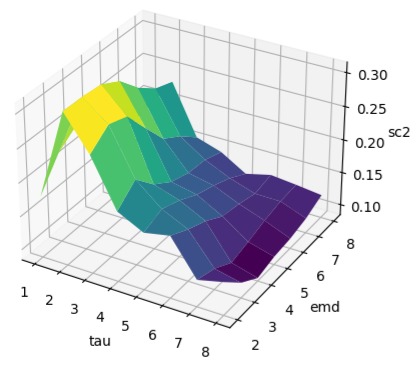} \end{minipage}  & \multicolumn{1}{c|}{\begin{minipage}{.153\linewidth} \centering \includegraphics[width=\linewidth]{imgs/ValidPCM/4VBoth1_sc2.png} \end{minipage}} & \begin{minipage}{.153\linewidth} \centering \includegraphics[width=\linewidth]{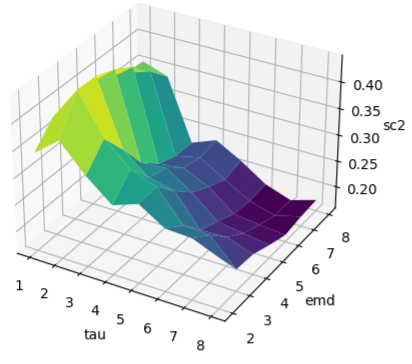} \end{minipage} \\ \hline
Ratio             &\begin{minipage}{.153\linewidth} \centering \includegraphics[width=\linewidth]{imgs/ValidPCM/4VDirect_ratio.png} \end{minipage}& \multicolumn{1}{c|}{\begin{minipage}{.153\linewidth} \centering \includegraphics[width=\linewidth]{imgs/ValidPCM/4VIndirect1_ratio.png} \end{minipage}}    & \begin{minipage}{.153\linewidth} \centering \includegraphics[width=\linewidth]{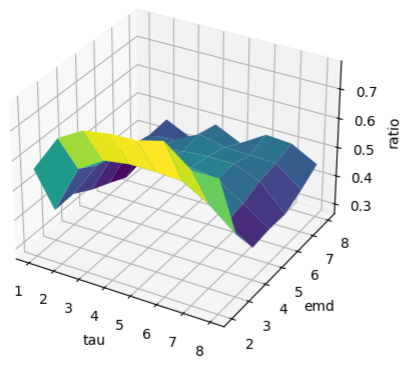} \end{minipage}    & \multicolumn{1}{c|}{\begin{minipage}{.153\linewidth} \centering \includegraphics[width=\linewidth]{imgs/ValidPCM/4VBoth1_ratio.png} \end{minipage}} & \begin{minipage}{.153\linewidth} \centering \includegraphics[width=\linewidth]{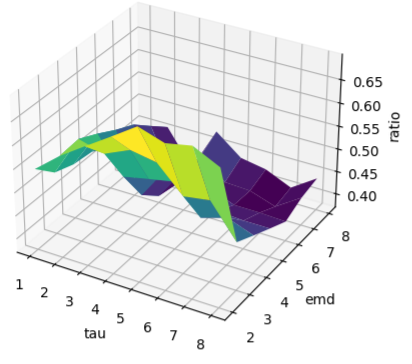} \end{minipage} \\ \hline
Label    &\begin{minipage}{.153\linewidth} \centering \includegraphics[width=\linewidth]{imgs/ValidPCM/4VDirect_label.png} \end{minipage}& \multicolumn{1}{c|}{\begin{minipage}{.153\linewidth} \centering \includegraphics[width=\linewidth]{imgs/ValidPCM/4VIndirect1_label.png} \end{minipage}}    & \begin{minipage}{.153\linewidth} \centering \includegraphics[width=\linewidth]{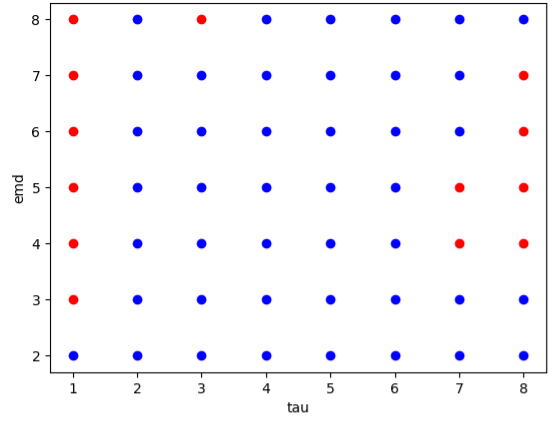} \end{minipage} & \multicolumn{1}{c|}{\begin{minipage}{.153\linewidth} \centering \includegraphics[width=\linewidth]{imgs/ValidPCM/4VBoth1_label.png} \end{minipage}} & \begin{minipage}{.153\linewidth} \centering \includegraphics[width=\linewidth]{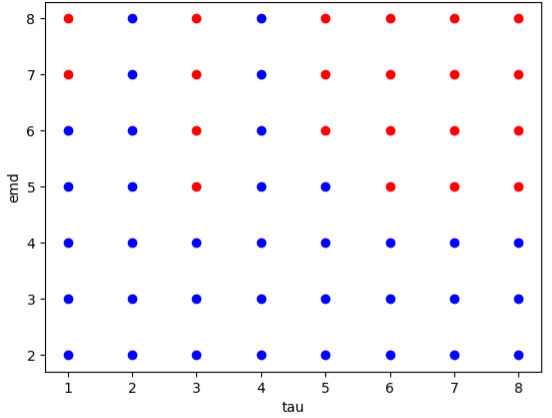} \end{minipage}
\end{tabular}
}
}

\caption{Performance of multiPCM: Correlation scores, correlation ratio, predicted label ($thres=0.45$) under grid search. Red dot indicates there isn't direct causality between the colored nodes, while blue indicates there is direct causality.}

\label{tab:app_multiPCM}

\end{table}

\section{Empirical Threshold Selection for multiPCM in Simulated Systems}
\label{appsec:thres_multiPCM}

The threshold for correlation ratio used in multiPCM was selected empirically by testing on our simulated systems under three distinct causality scenarios: direct, indirect, and both. The data of these scenarios are generated based on the description of Section~\ref{sec:genData}, varying from 3-variable to 7-variable with increasing complexity. The data generation is noise-free for simplicity.

Here we use lag $\tau=1$ and embedding dimension $E=7$ for multiPCM (since the discrete data generation uses 1 as the lag, and the highest dimension of all the systems is 7, we round the dimension up to the highest value to ensure performance across all simulated systems). We illustrate the output labels with varying ratio thresholds ranging from 0.05 to 0.95, with a step size of 0.05. Lower ratio threshold implies that the edge removal is more lenient and tends to retain more edges that lie between direct and indirect; while higher threshold indicates the edge removal is stricter and will only keep the edges whose direct correlation is strong enough.

For direct and both causality scenarios, the link should be retained (labeled "\textcolor{blue}{direct}"); For indirect causality scenarios, the link should be removed (labeled "\textcolor{red}{indirect}"). 

The selected threshold should achieve a relatively good accuracy across all systems, here we set the tolerance for each threshold value to 2 mistaken labels at each causality senario.

\begin{figure}[htb]
    \centering

    \subfigure[Direct (expected "\textcolor{blue}{direct}")]{
        \includegraphics[width=0.31\linewidth]{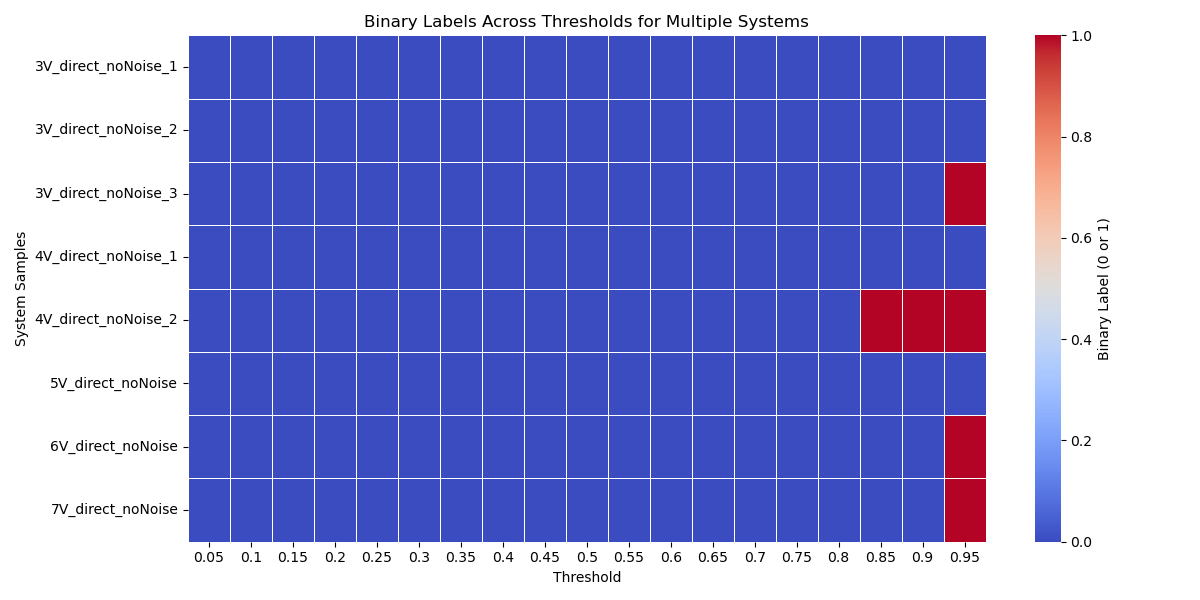}
        \label{appfig:direct_noNoise}
    }
    \hfill
    \subfigure[Indirect (expected "\textcolor{red}{indirect}")]{
        \includegraphics[width=0.31\linewidth]{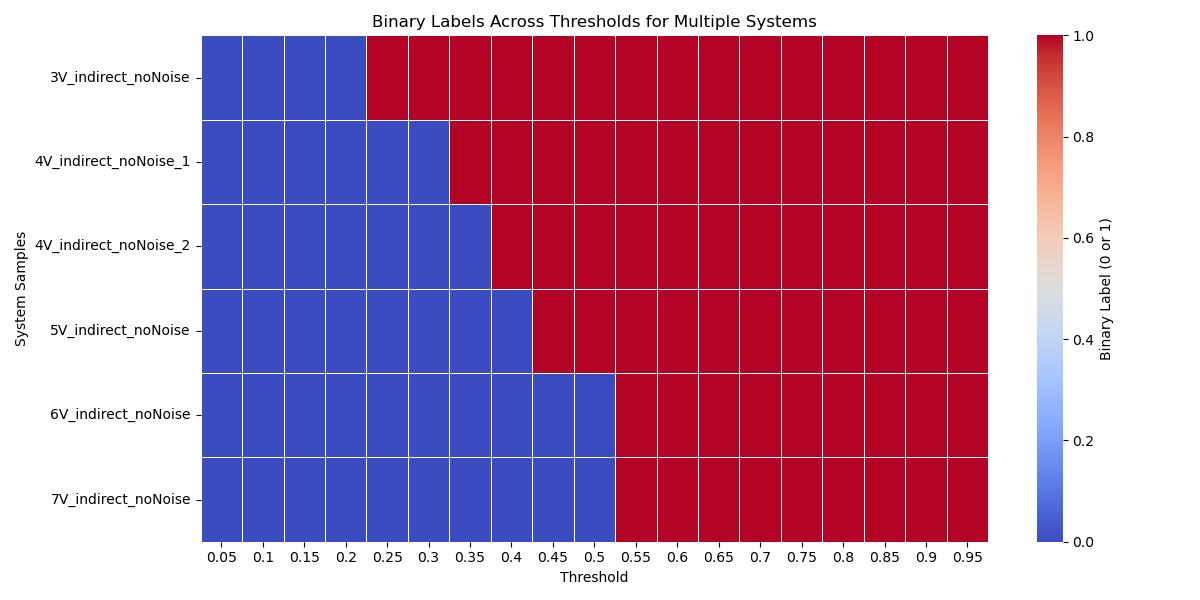}
        \label{appfig:indirect_noNoise}
    }
    \hfill
    \subfigure[Both (expected "\textcolor{blue}{direct}")]{
        \includegraphics[width=0.31\linewidth]{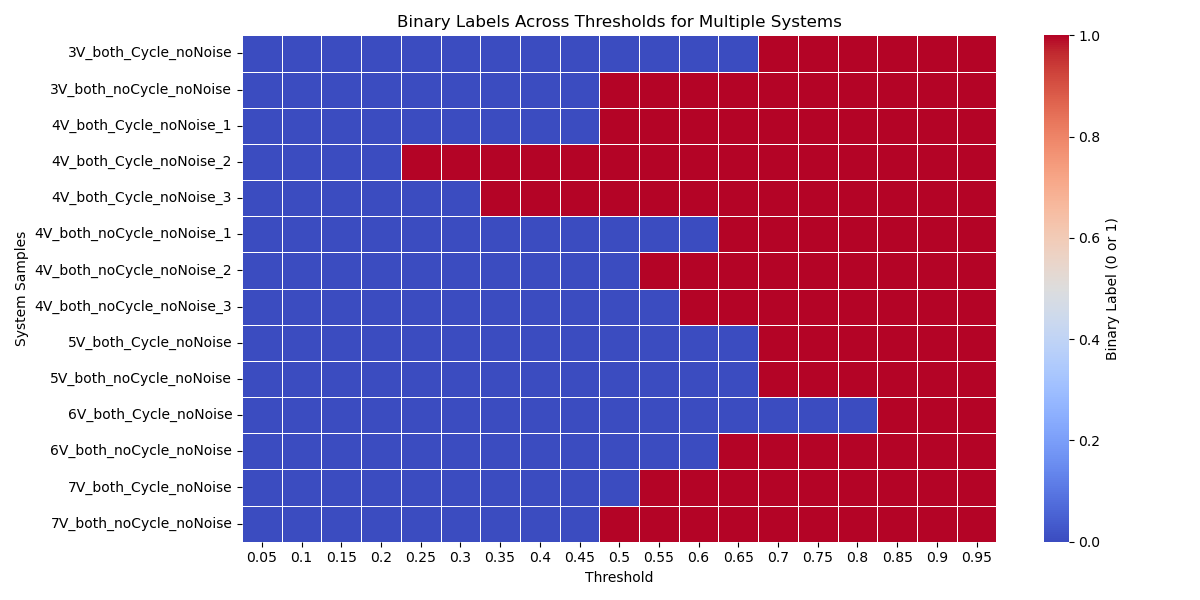}
        \label{appfig:both_noNoise}
    }
    \caption{Different PCM ratio thresholds in three causality scenarios and how they impact the predicted labels.}
    \label{appfig:noNoise-pcmThres}
\end{figure}

Fig.~\ref{appfig:noNoise-pcmThres} illustrates the results in heatmaps. In the "Direct" scenario, the thresholds that satisfy the tolerance are from 0.05 all the way to 0.9 (inclusive). In the "Indirect" scenario, the thresholds that satisfy the tolerance are from 0.45 to 0.95 (inclusive). In the "Both" scenario, the thresholds that satisfy the tolerance are from 0.05 to 0.45 (inclusive). Overall, 0.45 is relatively the best threshold value across all the simulation scenarios. 

Note that the appropriate threshold may vary depending on the system and the specific purpose of the study. For the ERA5 chain $tcw \Rightarrow rad \Rightarrow  T_{2m}$ discussed in Section~\ref{sec:weather_chain}, with MXMap lag $\tau=4$ and dimension $E=6$, the PCM ratio threshold required to remove the likely-indirect link $tcw\Rightarrow T_{2m}$ is 0.7. This higher threshold value may suggest the presence of a weaker but direct causal relationship between total cloud water $tcw$ and near-ground temperature $T_{2m}$, or it could reflect the influence of unaccounted latent variables. If the goal is to establish a causal graph with the correct order while tolerating some additional links, a denser graph may be acceptable for certain applications. In such cases, tolerating the retention of certain edges by multiPCM can still yield useful insights.

\section{ERA5 Meteorological Data: Candidate Testing Systems}
\label{appsec:era5}

\subsection{Cloud ($tcw$) $\Rightarrow$ Radiation ($rad$) $\Rightarrow$ Ground-Level Temperature ($T_{2m}$)}
\label{appsec:weather-chain}

This causal chain used in our work, where clouds impact radiation levels and in turn affect ground-level temperatures, is well-supported in meteorological research, especially concerning winter seasons.

Clouds play a significant role in modulating surface radiation, as they both reflect incoming solar radiation (shortwave) and trap outgoing terrestrial radiation (longwave)~\citep{stephens2005cloud}, hence $tcw\Rightarrow rad$.

During the day, clouds primarily act as a barrier to incoming solar radiation. Thick and extensive cloud cover reflects a significant portion of the solar radiation back into space, thereby lowering the temperature. At night, the longwave terrestrial radiation play a predominant role, which, combined with the presence of cloud cover (like an insulating blanket), leads to a warming effect. 

The warming effect is especially pronounced in winter~\citep{ramanathan1989cloud}, because the reduced daylight and lower solar angles reduce the overall shortwave radiation, rendering cloud cover's longwave effect more prominent. Research also indicates that during winter, the impact of clouds on radiation and temperature is more pronounced in polar regions, or in temperate zones with longer nights and lower solar angle, emphasizing the effect of cloud cover on maintaining warmer near-ground temperatures.

The timescale of this mechanism is usually within an hour or within a couple of hours~\citep{stephens2005cloud,pielke2005should}: the impact of cloud cover on radiation flux is almost instantaneous, and the response of ground-level temperature depends on several factors (e.g. time of the day, humidity, wind). This information is crucial for choosing the lag value for the state-space reconstruction used in cross mapping based methods and the maximal lag $\tau$ used in PCMCI.

\subsection{Introducing More Variables to the System}
\label{appsec:era5_moreV}

To enhance the complexity and realism of the ERA5 testing system, we introduce additional variables that reflect key meteorological processes. These include the following: 

\subsubsection{Near-Ground Temperature Advection $T_{adv950}$ }

$T_{adv950}$ stands for the near-ground temperature advection at the $950 hPa$ pressure level. It captures the transport of temperature influenced by atmospheric motion, and should be a direct cause of the near-ground temperature $T_{2m}$, hence $T_{adv950} \Rightarrow T_{2m}$.

\subsubsection{Radiation Components: Solar Radiation ($rad_{solar}$) and Terrestrial Radiation ($rad_{terr}$)}

Separating the radiation components into solar and terrestrial radiation enables finer analysis of their individual roles in energy balance and temperature dynamics:
\begin{itemize}
    \item \textbf{Solar Radiation ($rad_{solar}$)} is mainly and directly affected by cloud cover due to its role in shielding or transmitting shortwave radiation, hence $tcw\Rightarrow rad_{solar}$.
    \item \textbf{Terrestrial Radiation ($rad_{terr}$)} and cloud cover have a more complex interaction mechanism, and their causal relationship would be weaker and less prominent compared to $tcw\Rightarrow rad_{solar}$.
\end{itemize}
Both radiation components should be directly impacting the temperature, hence $rad_{solar} \Rightarrow T_{2m}$ and $rad_{terr} \Rightarrow T_{2m}$.

\subsection{Evaluating Causal Relationships}
\label{appsec:era5_eval}
We evaluate the model performance on the following systems:
\begin{enumerate}
    \item \textbf{The 3-variable chain $tcw \Rightarrow rad \Rightarrow  T_{2m}$:} Since the ground truth is known and relatively well justified, the causal discovery of this system will be evaluated in the same way as the simulated systems, using the metrics to compute the differences between truth causal graph and predicted graph;
    \item \textbf{The expanded 5-variable system with $tcw$, $rad_{solar}$, $rad_{terr}$, $T_{adv950}$ and $T_{2m}$:} While establishing the exact ground truth causal graph for this higher-dimensional weather system is challenging due to its inherent complexity, the following bivariate causal relationships are well-understood and serve as a benchmark for evaluation: $rad_{solar} \Rightarrow T_{2m}$, $rad_{terr} \Rightarrow T_{2m}$, $T_{adv950} \Rightarrow T_{2m}$, $tcw\Rightarrow rad_{solar}$; We evaluate how well the causal inference methods detect these known relationships consistently. 
\end{enumerate}

We present the full results along with the predicted graphs in Appendix Table~\ref{apptab:mxmap-era5-5V} to supplement the results presented in main Section~\ref{sec:era5_5V_eval}.

\begin{table}[htb]
\centering
\caption{Full results for detection of Benchmark Causal Relationships in the ERA5 5V System. A checkmark (green) indicates a correctly detected and oriented edge, a half-checkmark (gray) denotes a detected but ambiguously oriented edge, and a crossmark (red) represents an undetected or incorrectly oriented edge.}
\label{apptab:mxmap-era5-5V}

\resizebox{\textwidth}{!}{
\begin{tabular}{c|cccc}
Methods                          & tsFCI                            & VARLiNGAM                        & Granger                          & PCMCI                            \\ \hline
Output                           &  
\begin{minipage}{.24\linewidth} \centering \includegraphics[width=\linewidth]{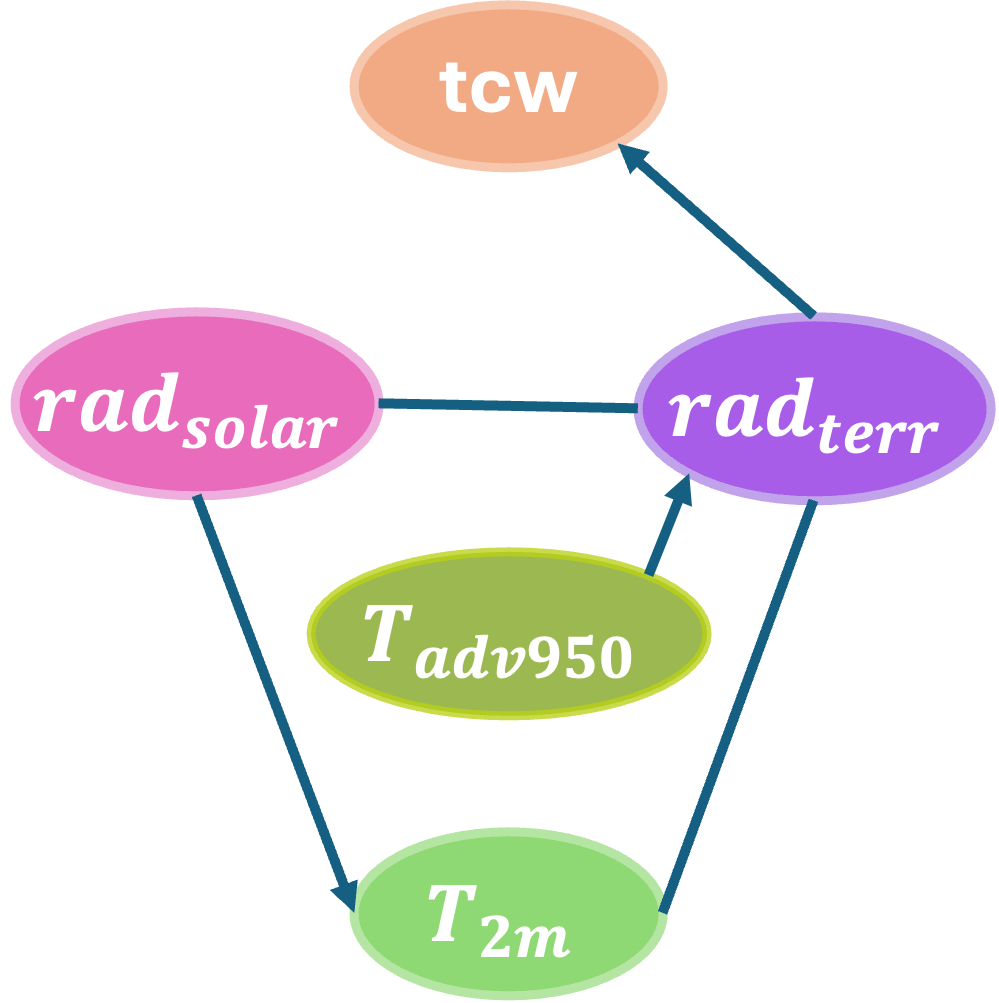} \end{minipage} &  
\begin{minipage}{.24\linewidth} \centering \includegraphics[width=\linewidth]{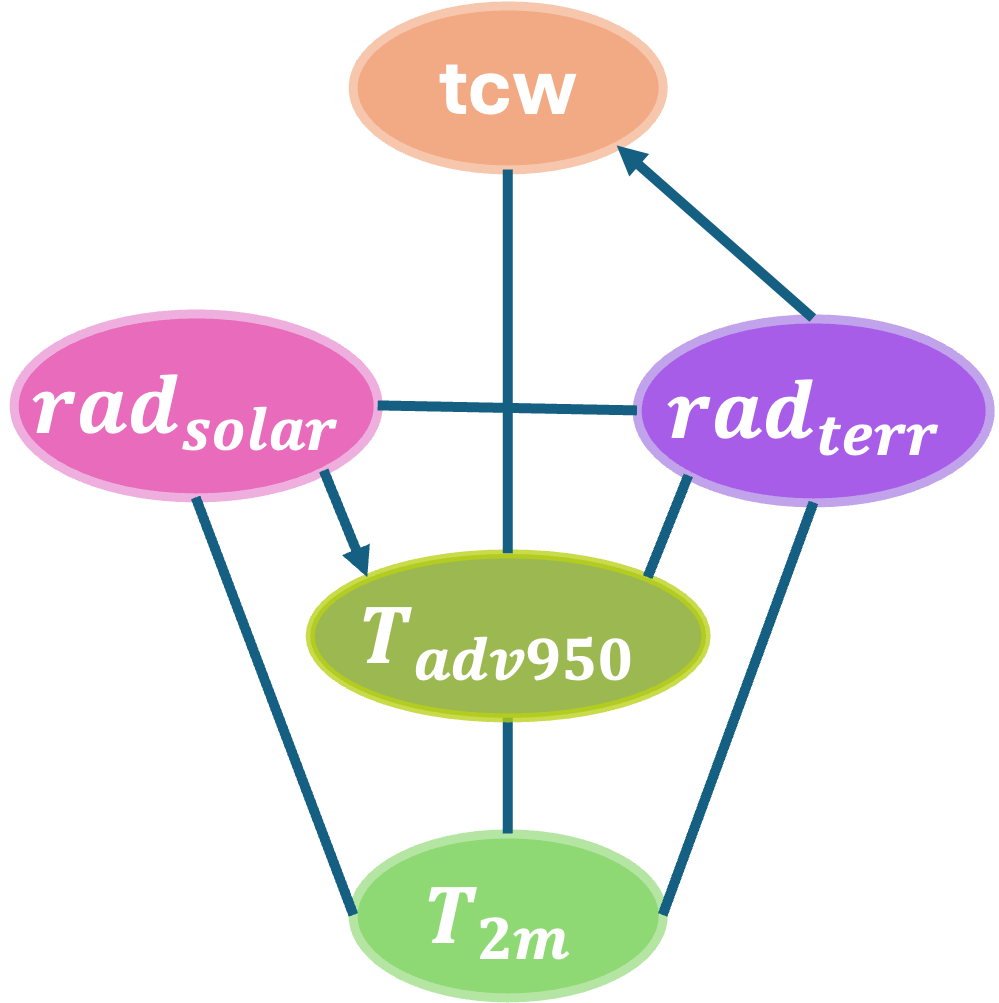} \end{minipage} &  
\begin{minipage}{.24\linewidth} \centering \includegraphics[width=\linewidth]{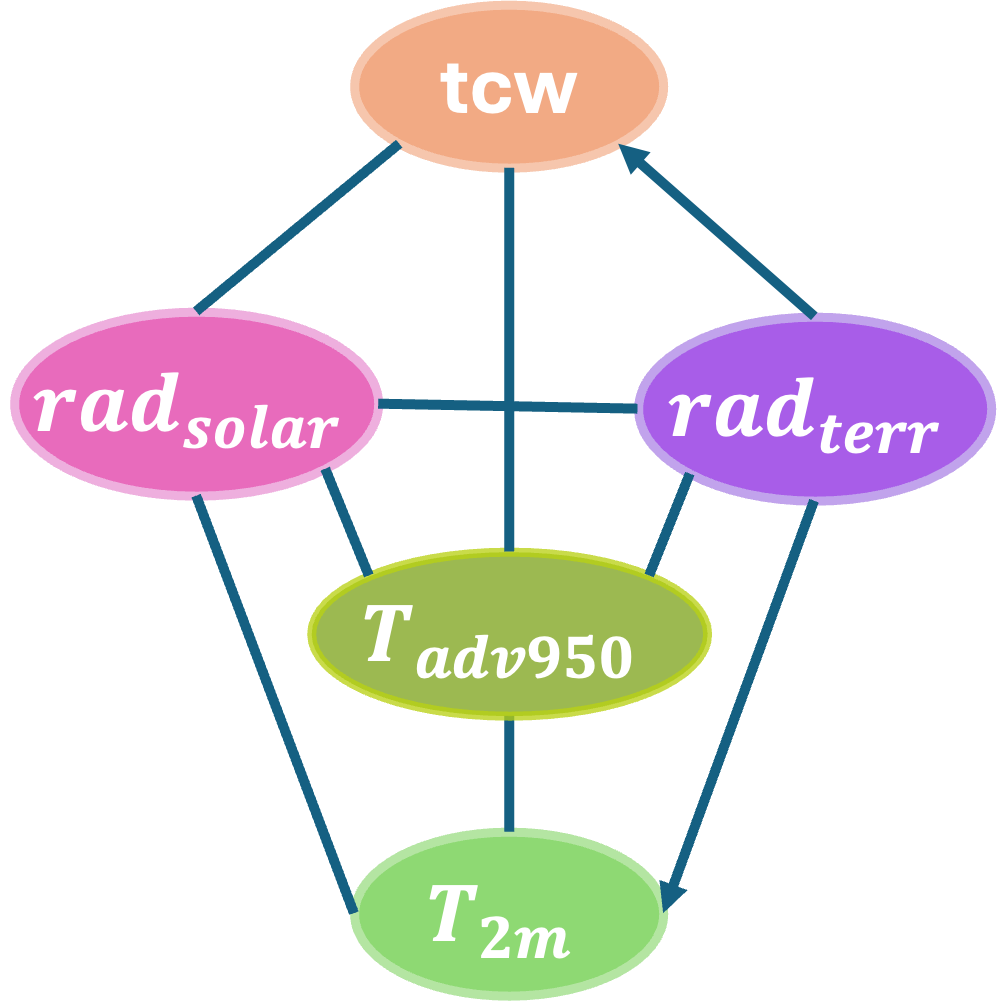} \end{minipage} &  
\begin{minipage}{.24\linewidth} \centering \includegraphics[width=\linewidth]{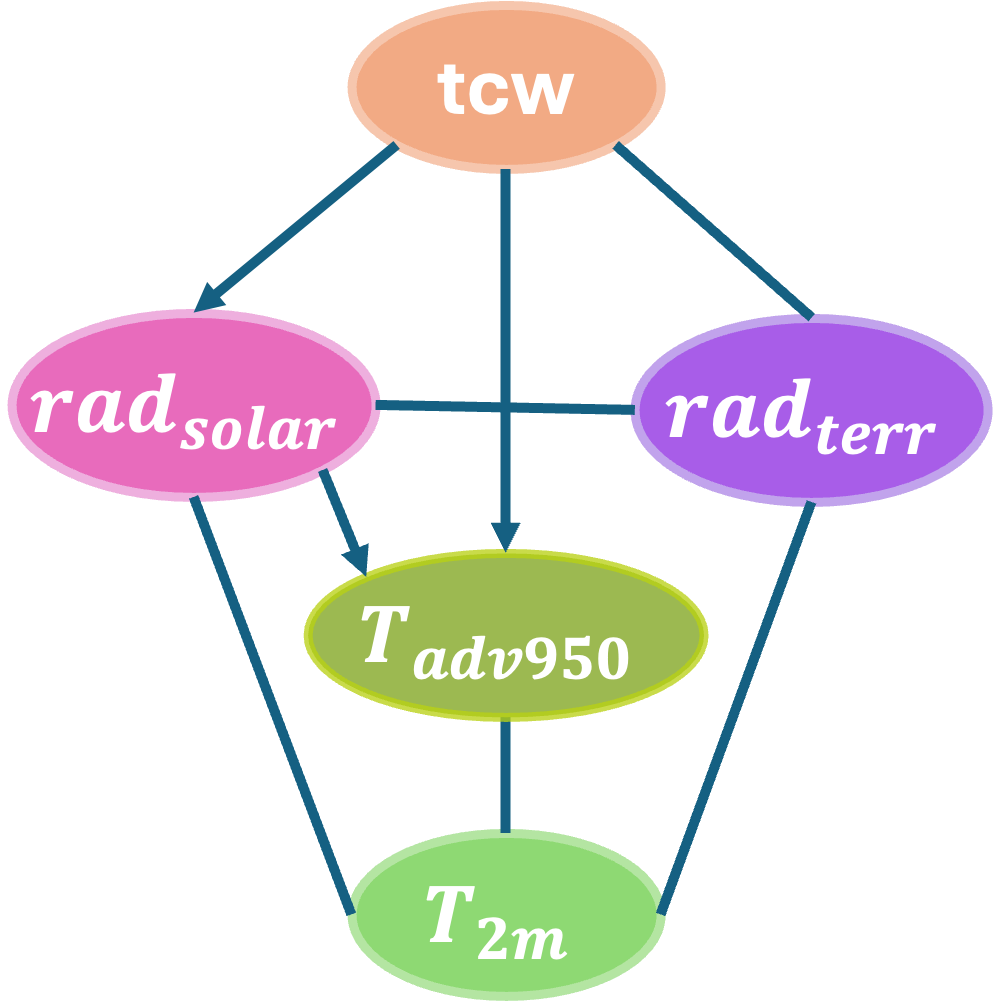} \end{minipage} \\ \hline
$rad_{solar} \Rightarrow T_{2m}$ & \cmark                       & \halfcheckmark & \halfcheckmark & \halfcheckmark \\
$rad_{terr} \Rightarrow T_{2m}$  & \halfcheckmark & \halfcheckmark & \cmark                       & \halfcheckmark \\
$T_{adv950} \Rightarrow T_{2m}$  & \xmark                           & \halfcheckmark & \halfcheckmark & \halfcheckmark \\
$tcw\Rightarrow rad_{solar}$     & \xmark                           & \xmark                           & \halfcheckmark & \cmark                       
\end{tabular}
}

\vspace{1em}

\resizebox{0.9\textwidth}{!}{
\begin{tabular}{c|ccc}
Methods                          & DYNOTEARS                        & SLARAC                           & MXMap                            \\ \hline
Output                           &  
\begin{minipage}{.24\linewidth} \centering \includegraphics[width=\linewidth]{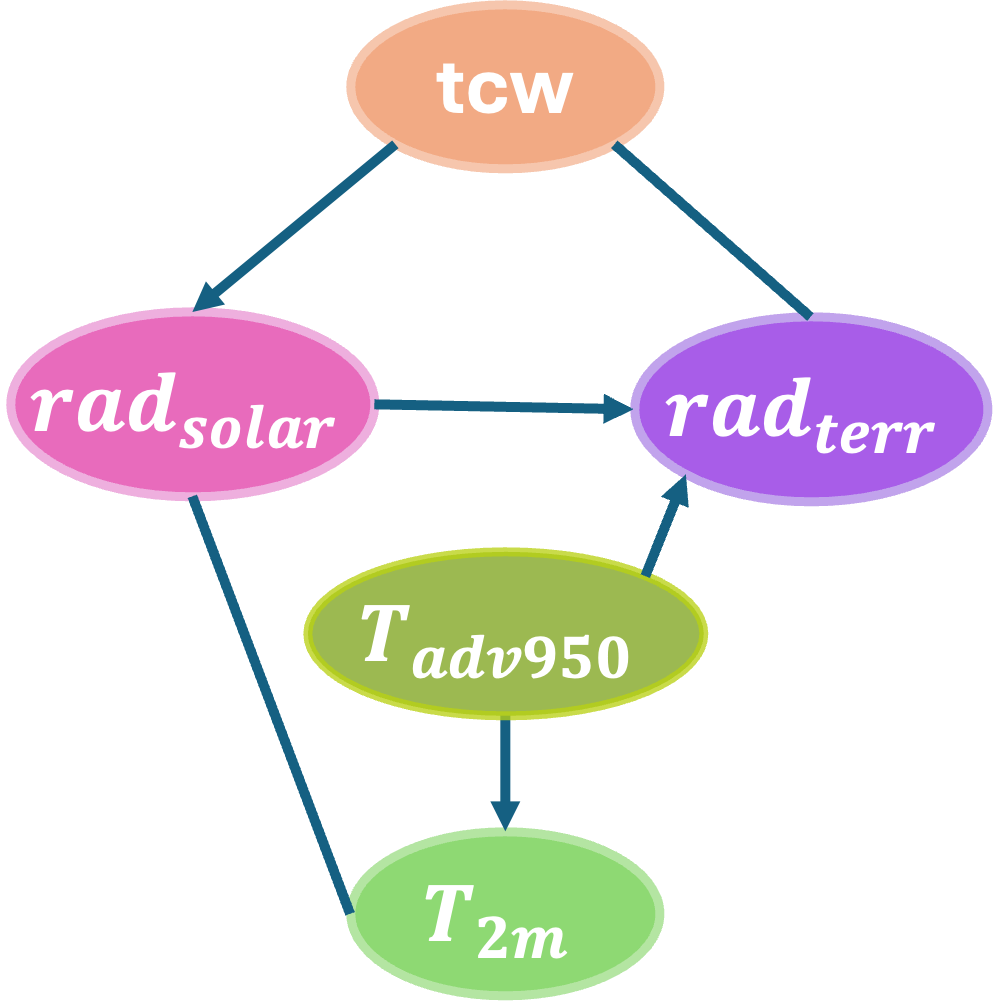} \end{minipage} &  
\begin{minipage}{.3\linewidth} \centering \includegraphics[width=\linewidth]{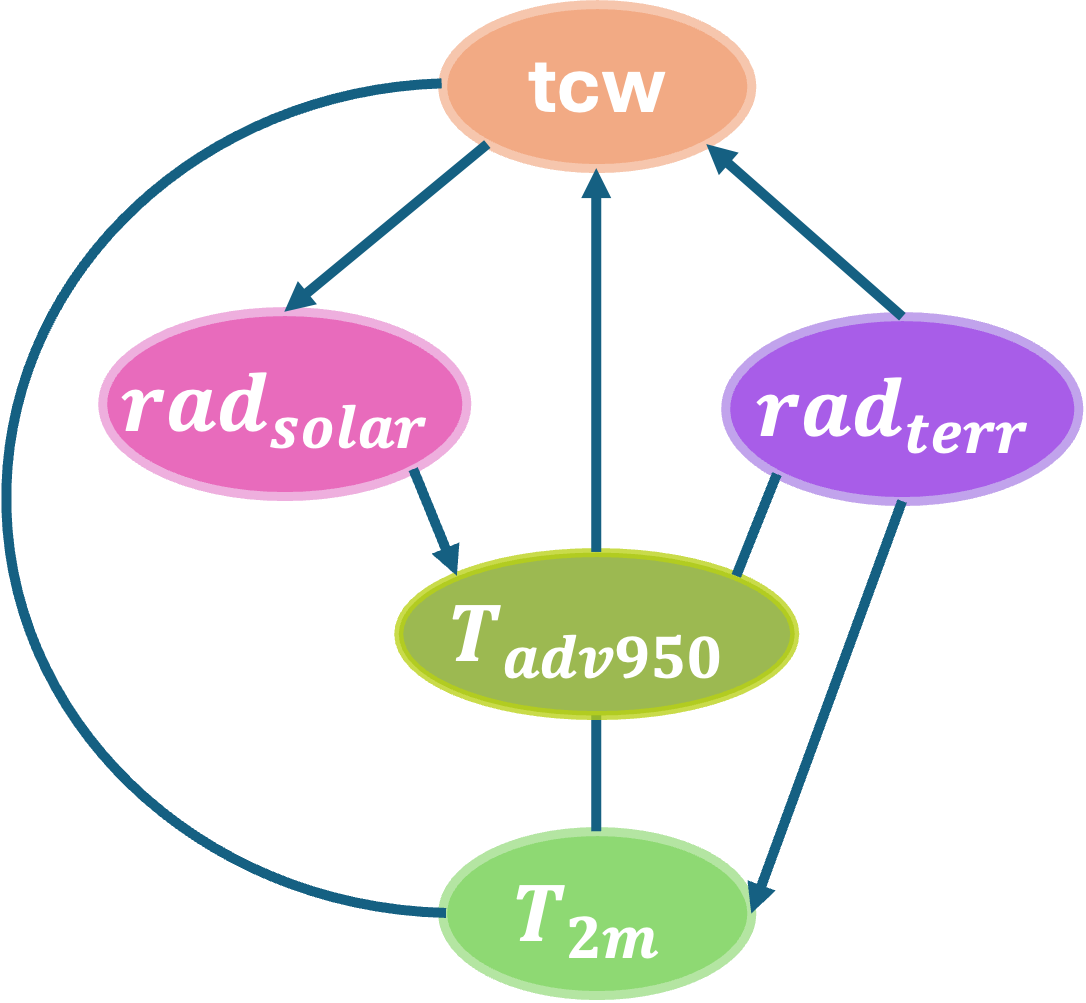} \end{minipage} &  
\begin{minipage}{.24\linewidth} \centering \includegraphics[width=\linewidth]{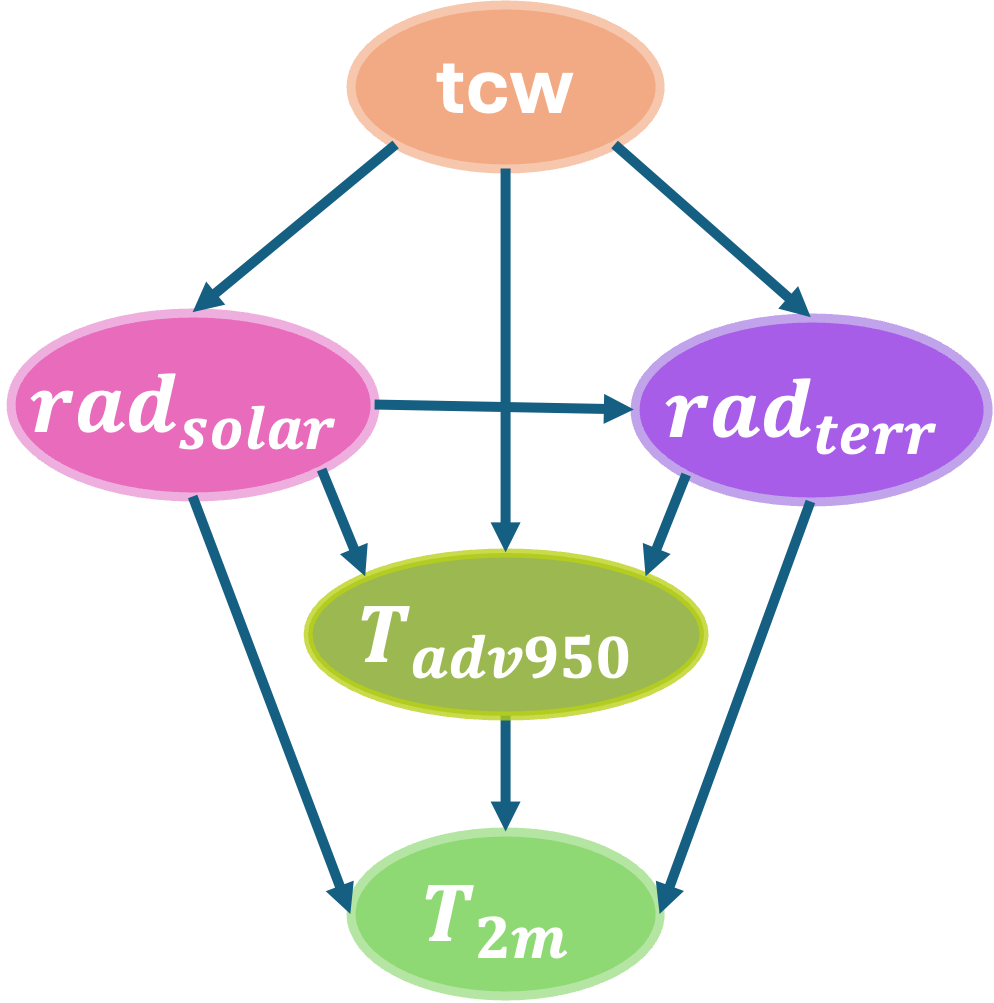} \end{minipage} \\ \hline
$rad_{solar} \Rightarrow T_{2m}$ & \halfcheckmark & \xmark                           & \cmark \\
$rad_{terr} \Rightarrow T_{2m}$  & \xmark                           & \cmark                       & \cmark \\
$T_{adv950} \Rightarrow T_{2m}$  & \cmark                       & \halfcheckmark & \cmark \\
$tcw\Rightarrow rad_{solar}$     & \cmark                       & \cmark                       & \cmark                       
\end{tabular}
}

\end{table}

\section{RESIT (Regression with Subsequent Independence Test) Framework}
\label{appsec:resit}

\begin{algorithm}[htb]
    \caption{Regression with Subsequent Independence Test (RESIT)}
    \label{alg:RESIT}
      \SetAlgoLined
    \KwIn{i.i.d. samples of a $p$-dimensional distribution on $(X_1, \ldots, X_p)$}
    \KwOut{Set of parents for each variable: $\{\text{pa}(1), \ldots, \text{pa}(p)\}$}

    $S \gets \{1, \ldots, p\}$

    $\pi \gets [\ ]$
    
    \textbf{Phase 1: Determine topological order}
    
    \While{$S \neq \emptyset$}{
        \For{$k \in S$}{
            Regress $X_k$ on $\{X_i\}_{i \in S \setminus \{k\}}$ ;
            
            Measure dependence between residual $r = X_k - \hat{X_k}$ and predictors $\{X_i\}_{i \in S \setminus \{k\}}$ ;
        }
        
        Let $k^*$ be the variable with the weakest dependence ;
        
        $S \gets S \setminus \{k^*\}$ ;
        
        $\text{pa}(k^*) \gets S$ ;
        
        $\pi \gets [k^*, \pi]$ ;
        
    }

    \textbf{Phase 2: Remove superfluous edges}
    
    \For{$k = 2$ \KwTo $p$}{
        \For{$\ell \in \text{pa}(\pi(k))$}{
            Regress $X_{\pi(k)}$ on $\{X_i\}_{i \in \text{pa}(\pi(k)) \setminus \{\ell\}}$ ;
            
            \If{residuals are independent of $\{X_i\}_{i \in \{\pi(1), \ldots, \pi(k-1)\}}$}{
                $\text{pa}(\pi(k)) \gets \text{pa}(\pi(k)) \setminus \{\ell\}$ ;
            }
        }
    }
\end{algorithm}

RESIT is an approach that extends the principles of ANM by iteratively conducting regression followed by an independence test on the residuals \citep{peters2014causal}. The original RESIT framework contains two phases, the causal order determination phase, and the edge elimination phase and is outlined in Algorithm~\ref{alg:RESIT}.

In Phase 1, one variable is selected as a prediction target (alleged effect) in each iteration, and the remaining variables are used as predictors (alleged causes) to fit a regression model. The regression error is computed, and the dependence between the residue and the predictors is measured. The variable with the weakest dependence (most likely effect) is identified and removed from the set. This process is repeated until all variables are ordered, establishing a topological order.

In Phase 2, for each variable with at least one parent, the set of parent variables is retrieved. For each parent variable, a regression model is fitted to predict the target variable using all other parents except the current one. The independence of the regression residue is compared with a threshold, and if it is sufficiently independent, the parent variable is removed from the parent set. This phase refines the causal structure by eliminating unnecessary dependencies, resulting in a clearer causal graph.

The RESIT algorithm iteratively establishes a causal order of variables by evaluating the strength of their causal relationships through regression and independence testing, ultimately refining the identified causal structure.

\section{Overview of the Baseline Methods}
\label{appsec:baseline_methods}


Time-Series Fast Causal Inference (tsFCI)~\citep{entner2010causal} is an extension of the Fast Causal Inference (FCI)~\citep{spirtes2013causal} algorithm for time-series data. It is designed to infer causal relationships from temporal data, taking into account potential latent confounders. The output is a Partial Ancestral Graph (PAG) with repeating structures over time steps to reflect the time-lagged causal impact. The predicted result allows unoriented edges, here we interpret such edge as bidirectional.

VAR-LiNGAM (Vector Autoregressive Linear Non-Gaussian Acyclic Model)~\citep{hyvarinen2010estimation} enhances the LiNGAM~\citep{shimizu2006linear} framework with vector autoregressive (VAR) models, enabling causal discovery in multivariate time series assuming non-Gaussian distribution and acyclic causal graph. 

PCMCI (Peter and Clark Momentary Conditional Independence)~\citep{runge2019detecting}, is a method designed to detect causal links in time series data. PCMCI extends the PC algorithm for temporal data, combining it with momentary conditional independence tests to account for lagged dependencies and reduce the risk of false discoveries in highly autocorrelated data. PCMCI identifies causal links at different levels of time lags, hence its time-unrolled causal graphs will be fully oriented, while the conventional causal graph may have contemporaneous, bidirectional or feedback processes, In our work, such link is interpreted as bidirectional.

Granger Causality~\citep{granger1969investigating} is a statistical hypothesis test used to determine whether one time-series can predict another, implying a causal relationship. It is based on the idea that a time-series $\mathbf{X}$ is said to Granger-cause another series 
$\mathbf{Y}$ if including past values of $\mathbf{X}$ improves the prediction of 
$\mathbf{Y}$ over a model that only uses past values of $\mathbf{Y}$.

DYNOTEARS~\citep{pamfil2020dynotears} is a Bayesian structure learning algorithm for time series data. It extends the DAG-NOTEARS~\citep{zheng2018dags} framework, allowing for detection of contemporaneous and time-lagged relationships between variables in a time-series. The method enforces acyclicity via a differentiable constraint, enabling scalable and accurate causal discovery. 

SLARAC (Subsampled Linear Auto-Regression Absolute Coefficients)~\citep{weichwald2020causal} is among the \texttt{tidybench} time-series causal discovery benchmarks. It is among the best performing models in the NeurIPS 2019 Causality for Climate competition~\citep{runge2020causality}, emphasizes the use of large regression coefficients over traditional small $p$-value thresholds.

Causal discovery with observational data is a rich and fast-developing field, offering diverse methods and baselines. For a comprehensive review of recent advances, please refer to~\citep{camps2023discovering}, where methodologies, challenges and applications are discussed in depth.

\section{Experimental Setup}
\label{appsec:exp_setup}

RESIT is implemented with the \texttt{LiNGAM} library and MLP regressor from \texttt{scikit-learn}~\citep{scikit-learn, sklearn_api}. To ensure a good enough fit quality, we choose the following parameters for they MLP regressor: 2 hidden layers of size 100, max number of iteration 1000.

tsFCI is implemented with FCI from \texttt{causal-learn} using the \texttt{fisherz} as independence test and level of significance $\alpha=0.05$; VAR-LiNGAM and Granger implemented with \texttt{causal-learn} with max lag of 1; DYNOTEARS is implemented using the \texttt{Causalnex} library, and SLARAC from \texttt{tidybench} with max lag of 1. PCMCI from the \texttt{tigramite} package is implemented using robust correlation (\texttt{RobustCorr}), with a significance level of $\alpha_{pc}=0.2$ for PC and a maximum lag of 1. As recommended by the original authors~\citep{runge2019detecting}, $\alpha_{pc}$ should not be interpreted as a strict measure of statistical significance. Instead, it is used in Phase 1 (PC) to reduce the search space for independence tests, and a slightly larger value for $\alpha_{pc}$ often provides faster inference while maintaining relatively good accuracy. 


\section{Evaluation Metrics for Graph Outputs}
\label{appsec:metrics}
The adjacency matrix of each output graph is defined with entries of 0 and 1, where the row index represents the cause variable and the column index represents the effect variable. An entry of 1 at $(row, column)$ indicates the existence of a causal edge between the two variables. To evaluate the model performance, the following metric scores are calculated on the adjacency matrices of the ground truth and the predicted graphs
:

\begin{itemize}
    \item \textbf{Precision:} Precision measures the proportion of correctly identified causal edges out of all edges predicted by the model. Mathematically, it is defined as the number of true positive edges divided by the total number of predicted edges. In the context of adjacency matrices, it is calculated as:
        \begin{equation}
            \text { Precision }=\frac{\sum_{i, j} \mathbb{I}\left(\hat{A}_{i j}=1 \text { and } A_{i j}=1\right)}{\sum_{i, j} \mathbb{I}\left(\hat{A}_{i j}=1\right)}
        \end{equation}
        where $\hat{A}$ is the predicted adjacency matrix, $A$ is the ground truth adjacency matrix, and $\mathbb{I}$ is the indicator function.
    \item \textbf{Recall:} Recall measures the proportion of correctly identified causal edges out of all actual edges in the ground truth. It is defined as the number of true positive edges divided by the total number of true edges. For adjacency matrices, recall is calculated as:
    \begin{equation}
            \text { Recall }=\frac{\sum_{i, j} \mathbb{I}\left(\hat{A}_{i j}=1 \text { and } A_{i j}=1\right)}{\sum_{i, j} \mathbb{I}\left(A_{i j}=1\right)}
        \end{equation}
        
    \item \textbf{F1:} The F1 score is the harmonic mean of precision and recall, providing a single metric that balances both concerns. It is particularly useful when the class distribution is imbalanced. The F1 score for adjacency matrices is calculated as:
    \begin{equation}
        \text { F1 Score }=2 \times \frac{\text { Precision } \times \text { Recall }}{\text { Precision }+ \text { Recall }}
    \end{equation}
    \item \textbf{Structural Hamming Distance (SHD):} It counts the number of edge additions, deletions, and reversals needed to transform the predicted adjacency matrix into the ground truth adjacency matrix. A lower SHD indicates a closer match to the true causal structure. Formally, SHD is calculated as:
    \begin{equation}
        \mathrm{SHD}=\sum_{i, j} \mathbb{I}\left(\hat{A}_{i j} \neq A_{i j}\right)
    \end{equation}
\end{itemize}

These four metrics provide a comprehensive evaluation of a model's performance in inferring causal structures from multivariate dynamical systems.

\section{MXMap: Current Limits}
\label{appsec:mxmap_limits}

\subsection{Runtime Complexity}

\begin{figure}[htb]
    \centering
    \includegraphics[width=0.7\linewidth]{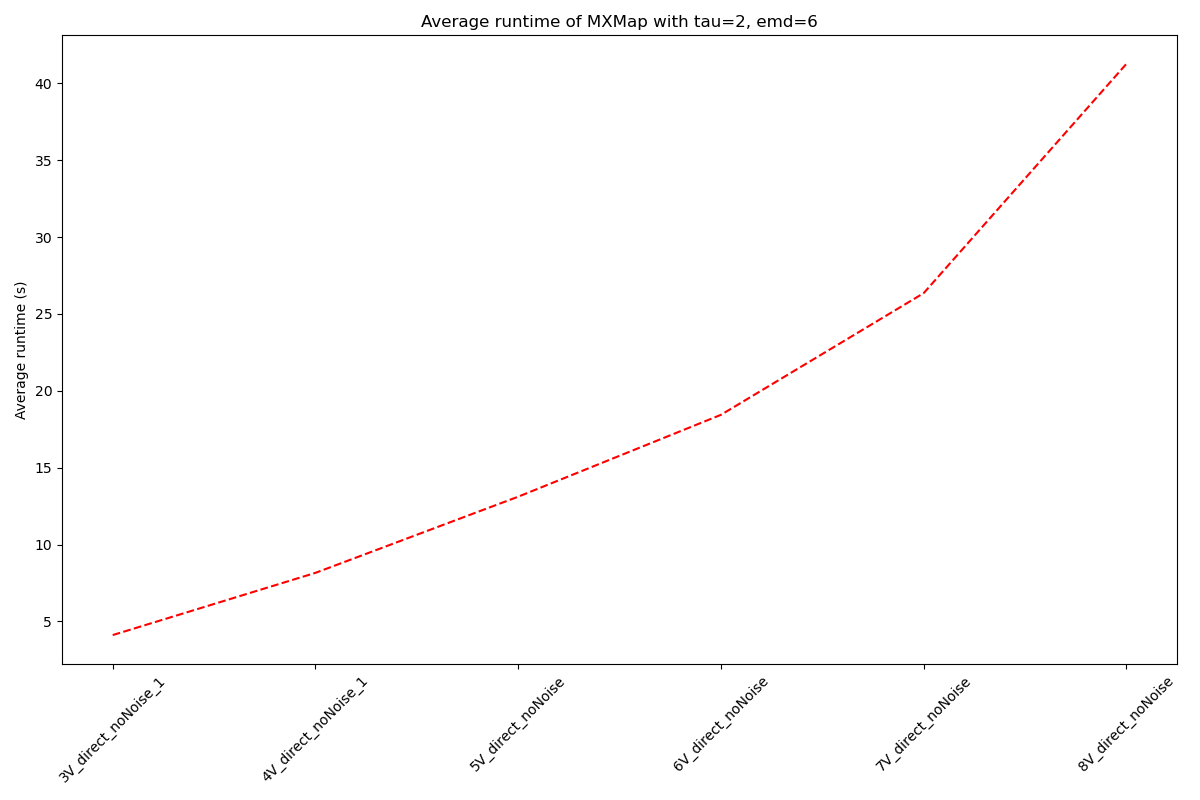}
    \caption{MXMap runtime as number of variables increase (chain structure).}
    \label{appfig:mxmap_runtime}
\end{figure}

MXMap has a runtime complexity of $\mathcal{O}(n^2)$ due to its pairwise processing design. To validate the runtime behavior, we conduct additional experiments using the chain causal structure with an increasing number of variables (from 3V chain to 8V chain) and plot the runtime in Fig.~\ref{appfig:mxmap_runtime}, where we can confirm the $\mathcal{O}(n^2)$ complexity.

To compare MXMap with the selected baseline methods, we track their runtime on the 8-variable chain (the highest dimensional system used in this work) with input length 4000. The approximate average runtimes were as follows:
\begin{itemize}
    \item \textbf{MXMap:} 40 seconds
    \item \textbf{tsFCI:} 15 seconds
    \item \textbf{Granger:} 3 seconds
    \item \textbf{VAR-LiNGAM:} 4 seconds
    \item \textbf{DYNOTEARS:} 5 seconds
    \item \textbf{SLARAC:} 15 seconds
\end{itemize}

Although MXMap is not the fastest method, its causal detection performance remains the most satisfactory, since its assumptions are tailored to nonlinear coupled dynamical systems such as the simulated systems used in the experiments. The $\mathcal{O}(n^2)$ complexity may have limited the method's scalability, it is precisely this design that enables MXMap to reliably detect causal cycles, a capability lacking in many alternative methods. This can be interpreted as a trade-off between computational cost and performance highlights MXMap’s ability to handle cyclic causal structures and achieve strong causal discovery in systems that match its assumptions.

\subsection{Potential Failure Cases}

MXMap, like other cross-mapping-based methods, relies on specific assumptions that may limit its applicability under certain conditions. The following outlines the potential failure cases for MXMap:

\begin{enumerate}
    \item \textbf{System doesn't have an attractor:} The presence of an attractor is fundamental to all cross-mapping-based methods, ensuring that delay embedding reconstruction accurately captures the system’s state-space structure. If such assumption is violated (e.g.,  transient systems, non-stationary time series, or systems dominated by stochasticity), there won't be reliable stable state-space structure to correctly establish time-index correspondence for nearest neighborhoods.
    \item \textbf{Strong synchrony or forcing:} As discussed in the original CCM paper~\citep{sugihara2012detecting}, strong forcing or synchrony can obscure the intrinsic dynamics of the forced (effect) variable, driving it to adopt the almost same manifold topology as the forcing (causal) variable. This compromises the reconstruction of independent state spaces, leading to unreliable causal inference.
    \item \textbf{Improper lag $\tau$ and dimension $E$ when constructing delay embeddings:} Incorrect choices of the delay-embedding parameters can hinder state-space reconstruction and lead to failure in CCM, as discussed by \href{https://www.youtube.com/watch?v=lKj4hr_2-Vg}{the tutorial} with examples.
    \item \textbf{High noise levels:} Observations with high noise may lead to failure in cross maps~\citep{monster2017causal}.
\end{enumerate}

\section{Complete Causal Discovery Evaluation of MXMap and Baseline Methods on Simulated Systems}
\label{appsec:complet}

In this section, we present the complete analysis of performance for each model. The ground truth causal graphs and predicted graphs are visualized for each coupled system, and the four metrics are calculated for evaluation (best score in bold). In most of the cases, MXMap demonstrates optimal or second-best performance in discovering the underlying causal structure.

\begin{table}[htb]
\begin{tabular}{l|c|c|c|c|c|c}
Method    & Ground Truth      & Predicted & Precision     & Recall       & F1            & SHD        \\ \hline
tsFCI     & \multirow{7}{*}[-3em]{\begin{minipage}{.17\linewidth} \centering \includegraphics[width=\linewidth]{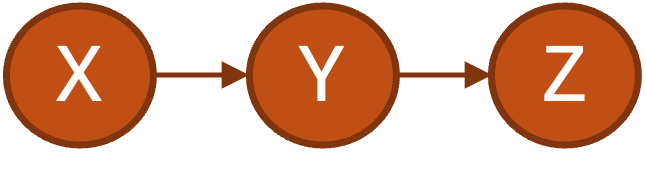} \end{minipage}} & \begin{minipage}{.17\linewidth} \centering \includegraphics[width=\linewidth]{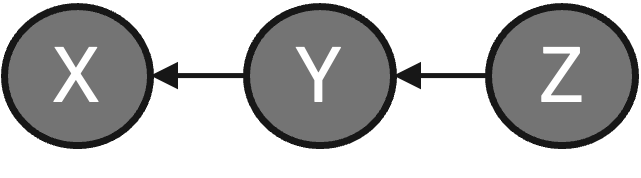} \end{minipage}& 0             & 0            & 0             & 4          \\
VARLiNGAM &                   &\begin{minipage}{.17\linewidth} \centering \includegraphics[width=\linewidth]{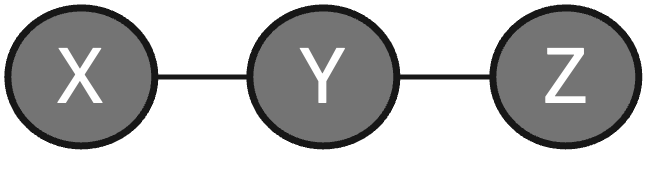} \end{minipage}& 0.50          & \textbf{1.0} & 0.67          & 2          \\
Granger   &                   &\begin{minipage}{.17\linewidth} \centering \includegraphics[width=\linewidth]{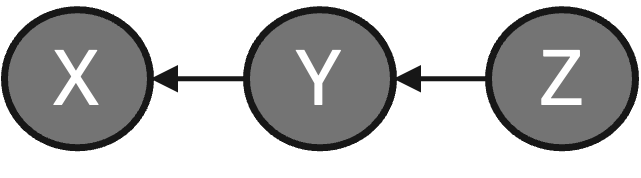} \end{minipage}& 0             & 0            & 0             & 4          \\
PCMCI     &                   &\begin{minipage}{.17\linewidth} \centering \includegraphics[width=\linewidth]{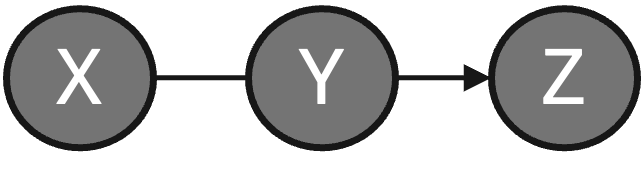} \end{minipage}& \textbf{0.67} & \textbf{1.0} & \textbf{0.80} & \textbf{1} \\
DYNOTEARS &                   &\begin{minipage}{.17\linewidth} \centering \includegraphics[width=\linewidth]{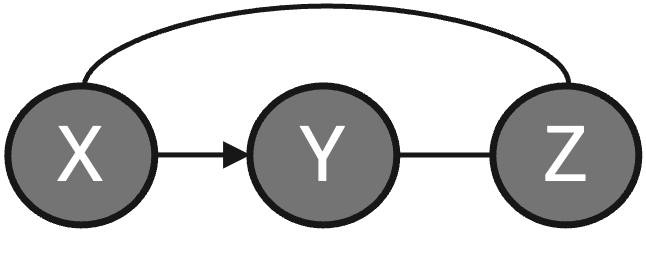} \end{minipage}& 0.40          & \textbf{1.0} & 0.57          & 3          \\
SLARAC    &                   &\begin{minipage}{.17\linewidth} \centering \includegraphics[width=\linewidth]{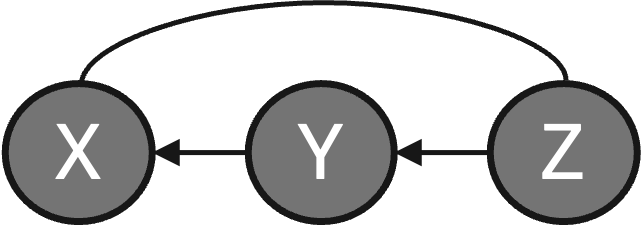} \end{minipage}& 0             & 0            & 0             & 6          \\ \cline{1-1} \cline{3-7} 
MXMap     &                   & \begin{minipage}{.17\linewidth} \centering \includegraphics[width=\linewidth]{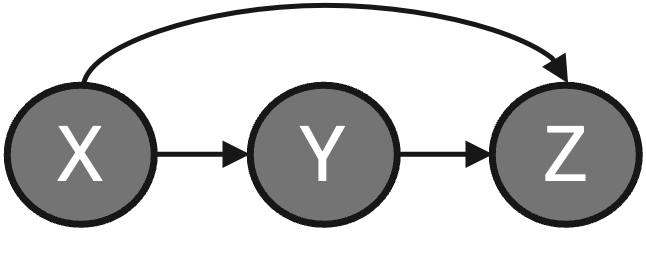} \end{minipage}& \textbf{0.67} & \textbf{1.0} & \textbf{0.80} & \textbf{1}
\end{tabular}
\caption{3V Chain (No Noise)}
\label{tab:3V_chain_noN}
\end{table}

\begin{table}[htb]
\begin{tabular}{l|c|c|c|c|c|c}
Method    & Ground Truth      & Predicted & Precision    & Recall        & F1           & SHD        \\ \hline
tsFCI     & \multirow{7}{*}[-3em]{\begin{minipage}{.17\linewidth} \centering \includegraphics[width=\linewidth]{imgs/sim_new/gt/3V_chain_gt.png} \end{minipage}} & \begin{minipage}{.17\linewidth} \centering \includegraphics[width=\linewidth]{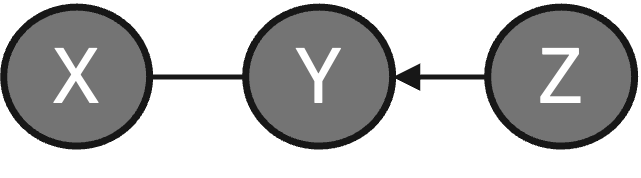} \end{minipage}& 0.33         & 0.50          & 0.40         & 3          \\
VARLiNGAM &                   & \begin{minipage}{.17\linewidth} \centering \includegraphics[width=\linewidth]{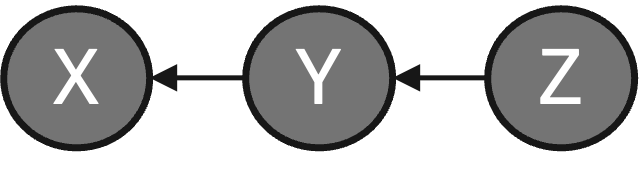} \end{minipage}& 0            & 0    & 0            & 4          \\
Granger   &                   & \begin{minipage}{.17\linewidth} \centering \includegraphics[width=\linewidth]{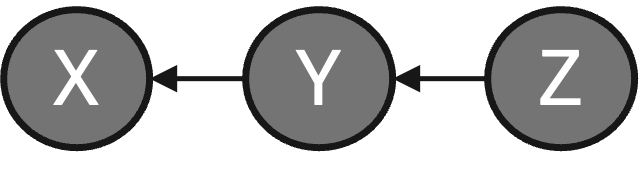} \end{minipage} & 0            & 0             & 0            & 4          \\
PCMCI     &                   &\begin{minipage}{.17\linewidth} \centering \includegraphics[width=\linewidth]{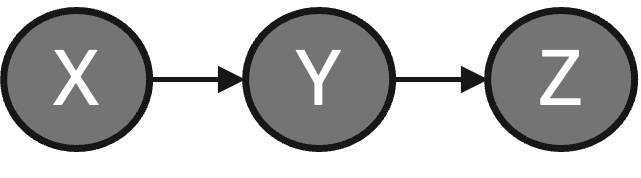} \end{minipage}& \textbf{1.0} & \textbf{1.0}  & \textbf{1.0} & \textbf{0} \\
DYNOTEARS &                   &\begin{minipage}{.17\linewidth} \centering \includegraphics[width=\linewidth]{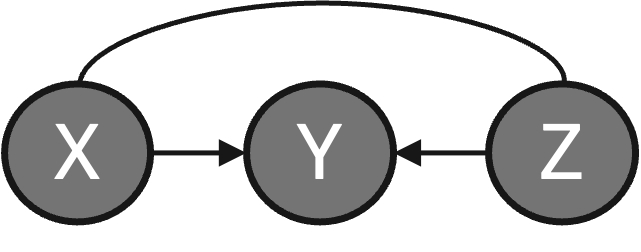} \end{minipage} & 0.25         & 0.50 & 0.33         & 4          \\
SLARAC    &                   & \begin{minipage}{.17\linewidth} \centering \includegraphics[width=\linewidth]{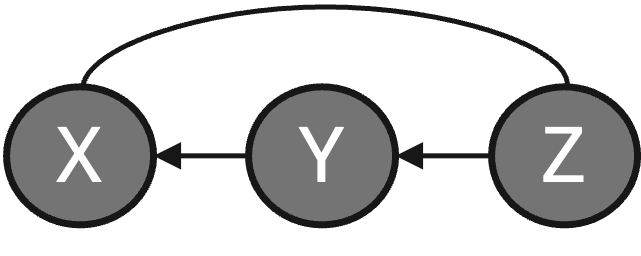} \end{minipage}& 0            & 0             & 0            & 6          \\ \cline{1-1} \cline{3-7} 
MXMap     &                   &\begin{minipage}{.17\linewidth} \centering \includegraphics[width=\linewidth]{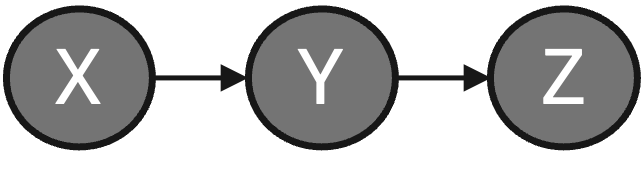} \end{minipage} & \textbf{1.0} & \textbf{1.0}  & \textbf{1.0} & \textbf{0}
\end{tabular}
\caption{3V Chain (Gaussian Additive Noise, Level 0.01)}
\label{tab:3V_chain_gN}
\end{table}

\begin{table}[htb]
\begin{tabular}{l|c|c|c|c|c|c}
Method    & Ground Truth      & Predicted & Precision    & Recall       & F1           & SHD        \\ \hline
tsFCI     & \multirow{7}{*}[-9.5em]{\begin{minipage}{.17\linewidth} \centering \includegraphics[width=\linewidth]{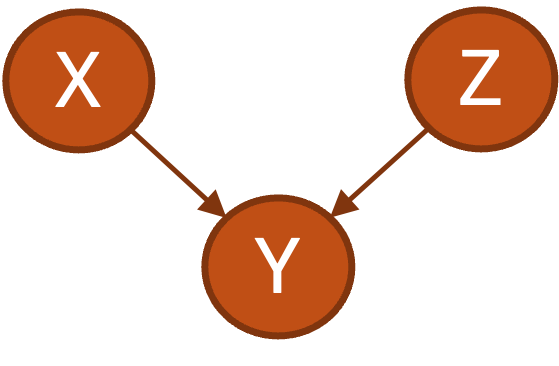} \end{minipage}} & \begin{minipage}{.17\linewidth} \centering \includegraphics[width=\linewidth]{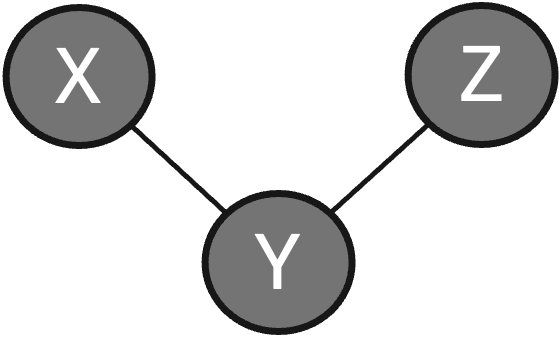} \end{minipage}          & 0.50         & 1.0          & 0.67         & 2          \\
VARLiNGAM &                   & \begin{minipage}{.17\linewidth} \centering \includegraphics[width=\linewidth]{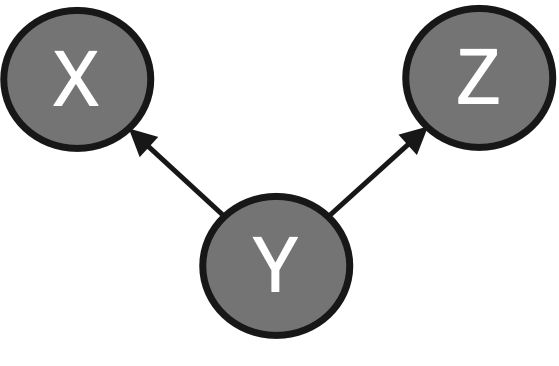} \end{minipage}  & 0            & 0            & 0            & 4          \\
Granger   &                   & \begin{minipage}{.17\linewidth} \centering \includegraphics[width=\linewidth]{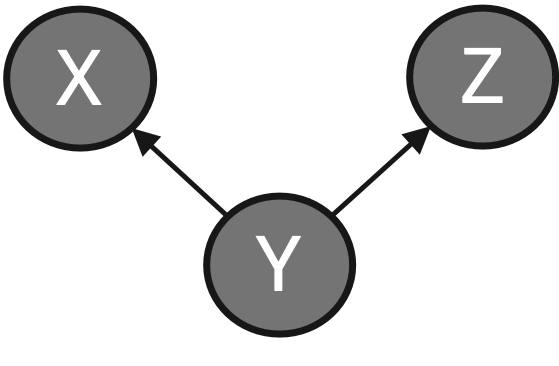} \end{minipage}  & 0            & 0            & 0            & 4          \\
PCMCI     &                   & \begin{minipage}{.17\linewidth} \centering \includegraphics[width=\linewidth]{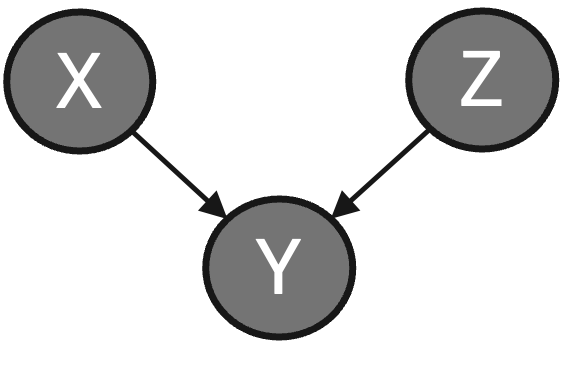} \end{minipage}  & \textbf{1.0} & \textbf{1.0} & \textbf{1.0} & \textbf{0} \\
DYNOTEARS &                   & \begin{minipage}{.17\linewidth} \centering \includegraphics[width=\linewidth]{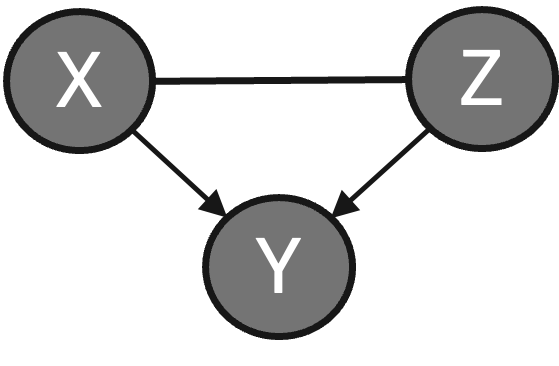} \end{minipage}  & 0.50         & \textbf{1.0} & 0.67         & 2          \\
SLARAC    &                   & \begin{minipage}{.17\linewidth} \centering \includegraphics[width=\linewidth]{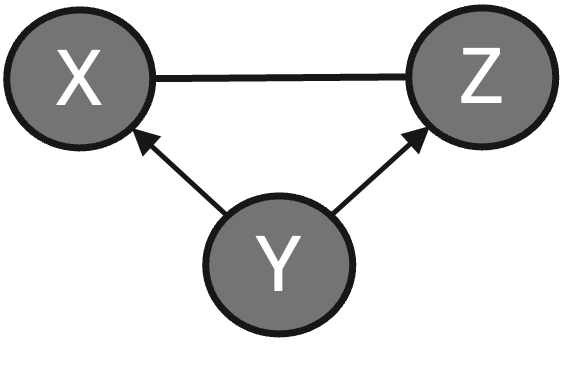} \end{minipage}  & 0            & 0            & 0            & 6          \\ \cline{1-1} \cline{3-7} 
MXMap     &                   & \begin{minipage}{.17\linewidth} \centering \includegraphics[width=\linewidth]{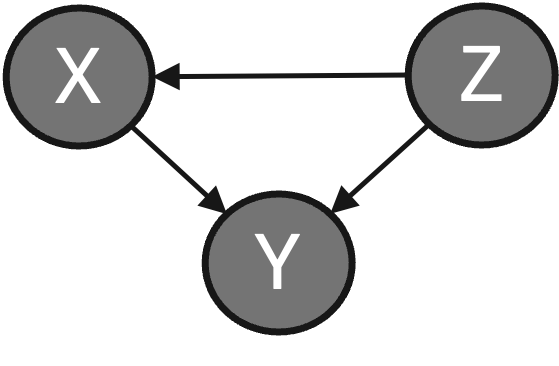} \end{minipage}  & 0.67         & \textbf{1.0} & 0.80         & \textbf{1}
\end{tabular}
\caption{3V Immorality (No Noise)}
\label{tab:3V_immo_noN}
\end{table}

\begin{table}[htb]
\begin{tabular}{l|c|c|c|c|c|c}
Method    & Ground Truth      & Predicted & Precision    & Recall       & F1           & SHD        \\ \hline
tsFCI     & \multirow{7}{*}[-9.5em]{\begin{minipage}{.17\linewidth} \centering \includegraphics[width=\linewidth]{imgs/sim_new/gt/3V_immorality_gt.png} \end{minipage}} & \begin{minipage}{.17\linewidth} \centering \includegraphics[width=\linewidth]{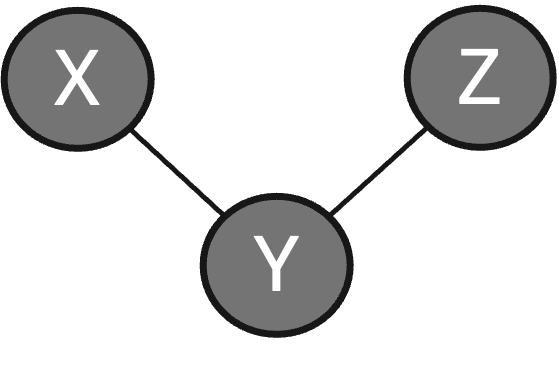} \end{minipage}           & 0.50         & \textbf{1.0} & 0.67         & 2          \\
VARLiNGAM &                   & \begin{minipage}{.17\linewidth} \centering \includegraphics[width=\linewidth]{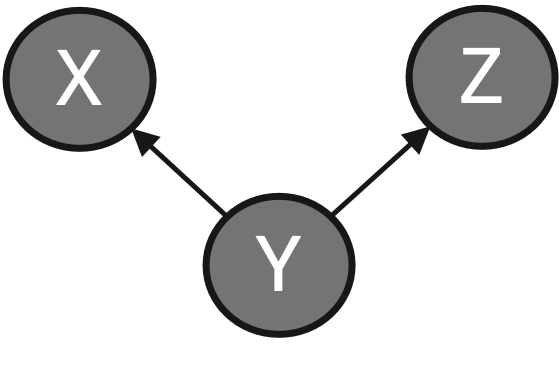} \end{minipage}  & 0            & 0            & 0            & 4          \\
Granger   &                   & \begin{minipage}{.17\linewidth} \centering \includegraphics[width=\linewidth]{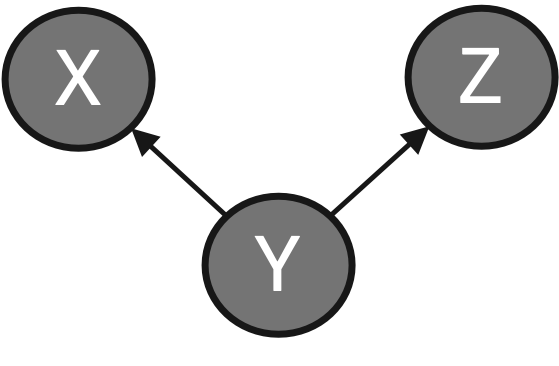} \end{minipage} & 0            & 0            & 0            & 4          \\
PCMCI     &                   & \begin{minipage}{.17\linewidth} \centering \includegraphics[width=\linewidth]{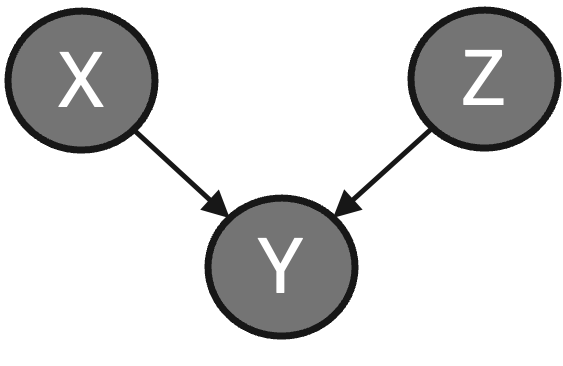} \end{minipage}  & \textbf{1.0} & \textbf{1.0} & \textbf{1.0} & \textbf{0} \\
DYNOTEARS &                   &  \begin{minipage}{.17\linewidth} \centering \includegraphics[width=\linewidth]{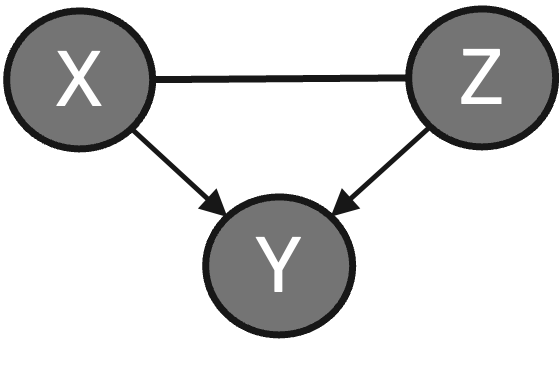} \end{minipage}  & 0.50         & \textbf{1.0} & 0.67         & 2          \\
SLARAC    &                   & \begin{minipage}{.17\linewidth} \centering \includegraphics[width=\linewidth]{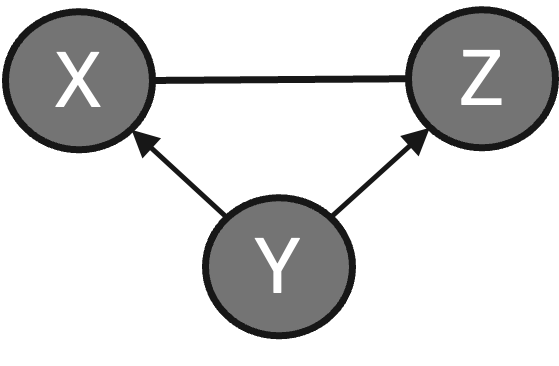} \end{minipage}  & 0            & 0            & 0            & 6          \\ \cline{1-1} \cline{3-7} 
MXMap     &                   &  \begin{minipage}{.17\linewidth} \centering \includegraphics[width=\linewidth]{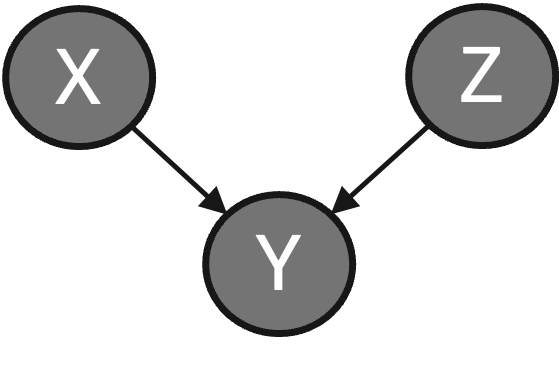} \end{minipage}  & \textbf{1.0} & \textbf{1.0} & \textbf{1.0} & \textbf{0}
\end{tabular}
\caption{3V Immorality (Gaussian Additive Noise, Level 0.01)}
\label{tab:3V_immo_gN}
\end{table}

\begin{table}[htb]
\begin{tabular}{l|c|c|c|c|c|c}
Method    & Ground Truth      & Predicted & Precision    & Recall       & F1           & SHD        \\ \hline
tsFCI     & \multirow{7}{*}[-6.2em]{\begin{minipage}{.17\linewidth} \centering \includegraphics[width=\linewidth]{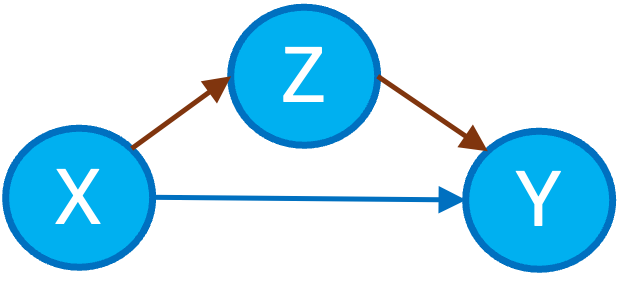} \end{minipage}} &\begin{minipage}{.17\linewidth} \centering \includegraphics[width=\linewidth]{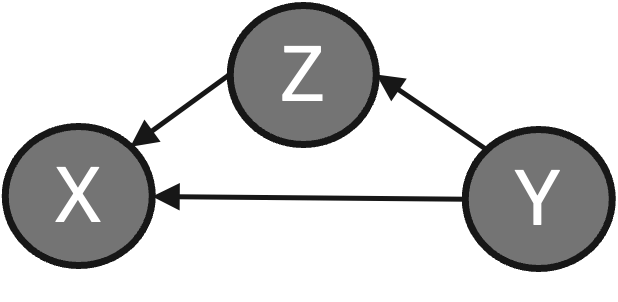} \end{minipage}           & 0            & 0            & 0            & 6          \\
VARLiNGAM &                   &  \begin{minipage}{.17\linewidth} \centering \includegraphics[width=\linewidth]{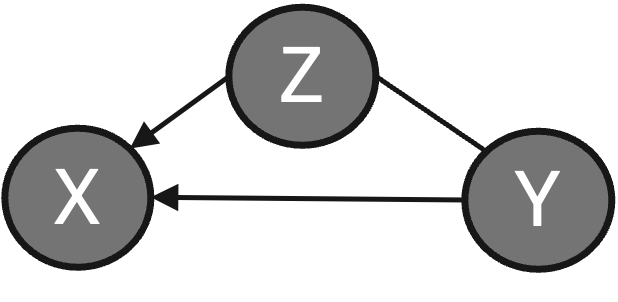} \end{minipage} & 0.25         & 0.33         & 0.29         & 5          \\
Granger   &                   &\begin{minipage}{.17\linewidth} \centering \includegraphics[width=\linewidth]{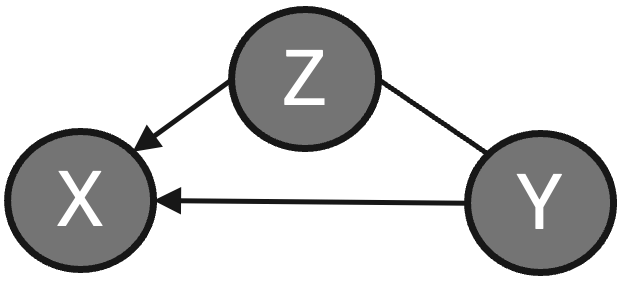} \end{minipage}  & 0.25         & 0.33         & 0.29         & 5          \\
PCMCI     &                   & \begin{minipage}{.17\linewidth} \centering \includegraphics[width=\linewidth]{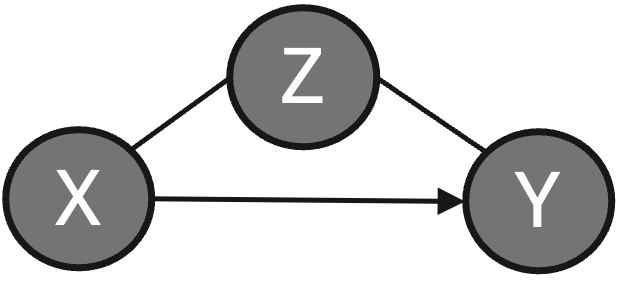} \end{minipage} & 0.60         & \textbf{1.0} & 0.75         & 2          \\
DYNOTEARS &                   & \begin{minipage}{.17\linewidth} \centering \includegraphics[width=\linewidth]{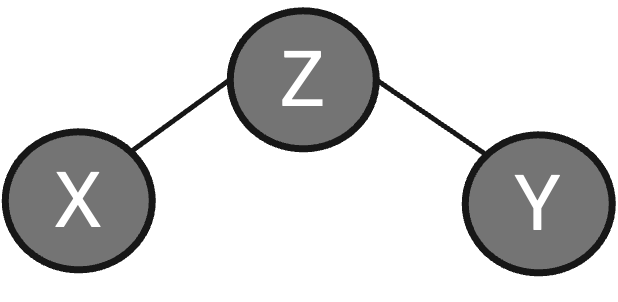} \end{minipage} & 0.50         & 0.67         & 0.57         & 3          \\
SLARAC    &                   &  \begin{minipage}{.17\linewidth} \centering \includegraphics[width=\linewidth]{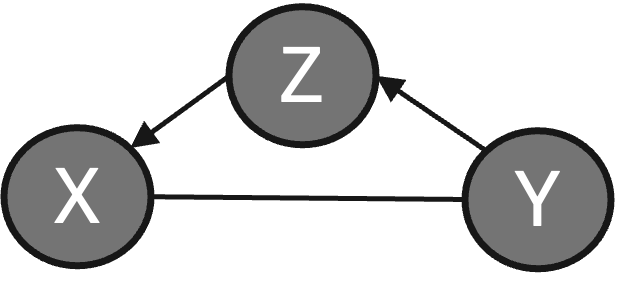} \end{minipage} & 0.25         & 0.33         & 0.29         & 5          \\ \cline{1-1} \cline{3-7} 
MXMap     &                   & \begin{minipage}{.17\linewidth} \centering \includegraphics[width=\linewidth]{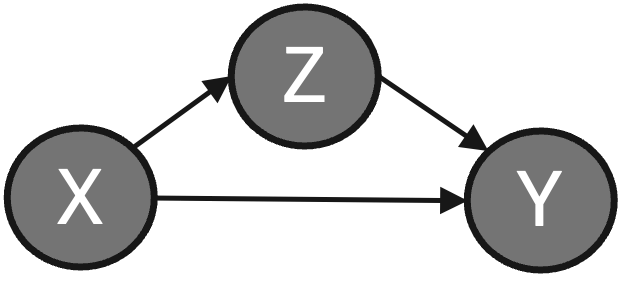} \end{minipage} & \textbf{1.0} & \textbf{1.0} & \textbf{1.0} & \textbf{0}
\end{tabular}
\caption{3V No Cycle (No Noise)}
\label{tab:3V_noCycle_noN}
\end{table}

\begin{table}[htb]
\begin{tabular}{l|c|c|c|c|c|c}
Method    & Ground Truth      & Predicted & Precision    & Recall       & F1           & SHD        \\ \hline
tsFCI     & \multirow{7}{*}[-6.2em]{\begin{minipage}{.17\linewidth} \centering \includegraphics[width=\linewidth]{imgs/sim_new/gt/3V_both_noCycle_gt.png} \end{minipage}} & \begin{minipage}{.17\linewidth} \centering \includegraphics[width=\linewidth]{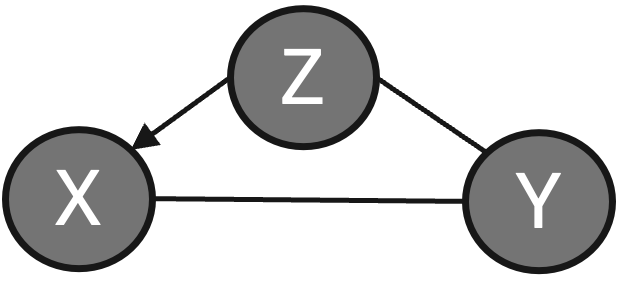} \end{minipage}          & 0.40         & 0.67         & 0.5          & 4          \\
VARLiNGAM &                   & \begin{minipage}{.17\linewidth} \centering \includegraphics[width=\linewidth]{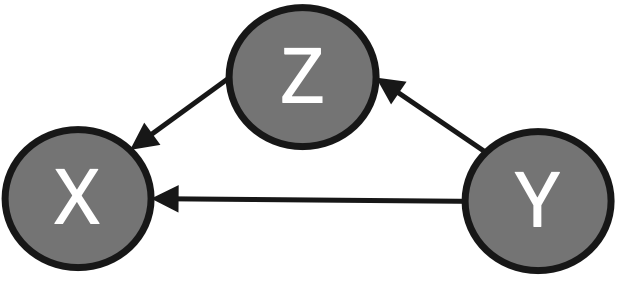} \end{minipage} & 0            & 0            & 0            & 6          \\
Granger   &                   & \begin{minipage}{.17\linewidth} \centering \includegraphics[width=\linewidth]{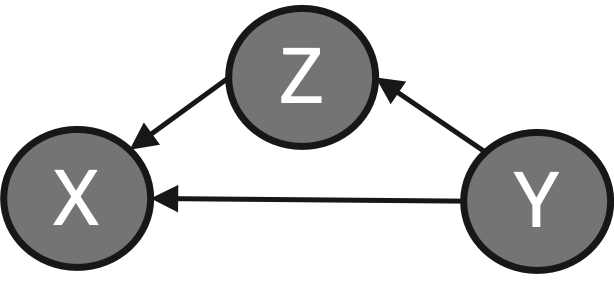} \end{minipage} & 0            & 0            & 0            & 6          \\
PCMCI     &                   & \begin{minipage}{.17\linewidth} \centering \includegraphics[width=\linewidth]{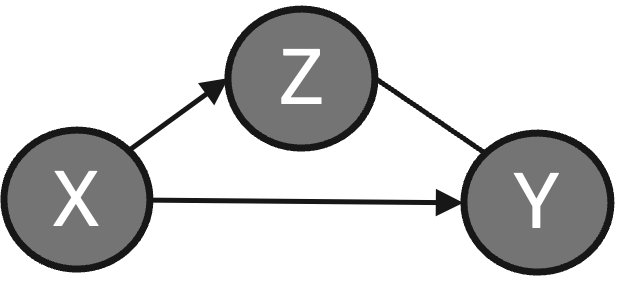} \end{minipage} & 0.75         & \textbf{1.0} & 0.86         & 1          \\
DYNOTEARS &                   & \begin{minipage}{.17\linewidth} \centering \includegraphics[width=\linewidth]{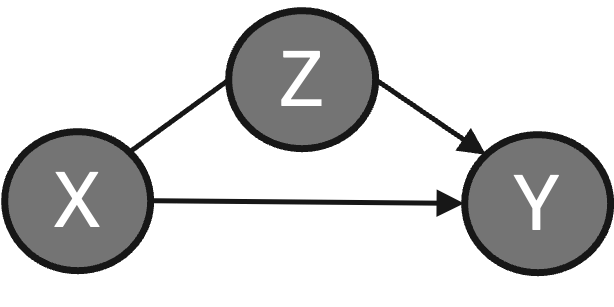} \end{minipage}& 0.75         & \textbf{1.0} & 0.86         & 1          \\
SLARAC    &                   & \begin{minipage}{.17\linewidth} \centering \includegraphics[width=\linewidth]{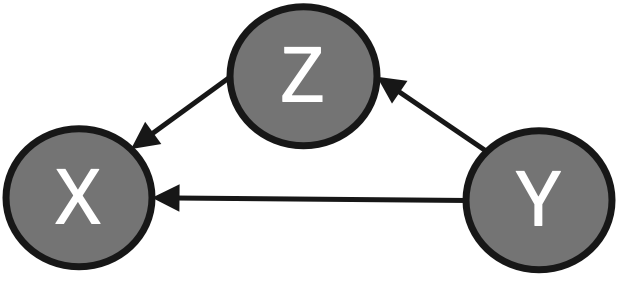} \end{minipage} & 0            & 0            & 0            & 6          \\ \cline{1-1} \cline{3-7} 
MXMap     &                   & \begin{minipage}{.17\linewidth} \centering \includegraphics[width=\linewidth]{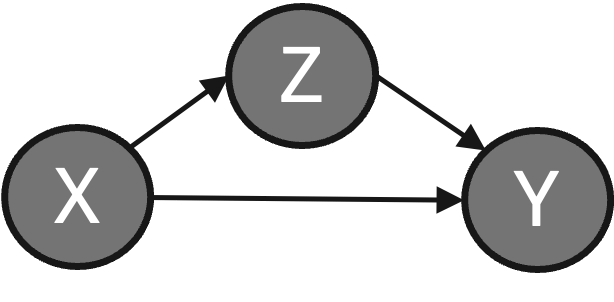} \end{minipage} & \textbf{1.0} & \textbf{1.0} & \textbf{1.0} & \textbf{0}
\end{tabular}
\caption{3V No Cycle (Gaussian Additive Noise, Level 0.01)}
\label{tab:3V_noCycle_gN}
\end{table}

\begin{table}[htb]
\begin{tabular}{l|c|c|c|c|c|c}
Method    & Ground Truth      & Predicted & Precision    & Recall       & F1           & SHD        \\ \hline
tsFCI     & \multirow{7}{*}[-6.2em]{\begin{minipage}{.17\linewidth} \centering \includegraphics[width=\linewidth]{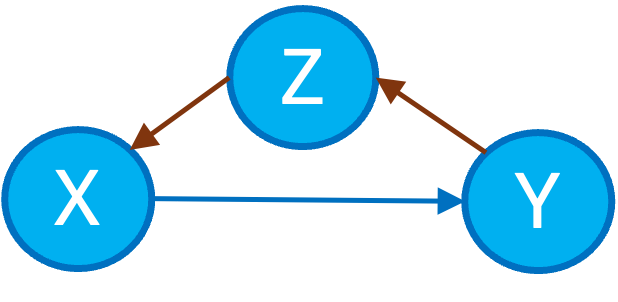} \end{minipage}} &  \begin{minipage}{.17\linewidth} \centering \includegraphics[width=\linewidth]{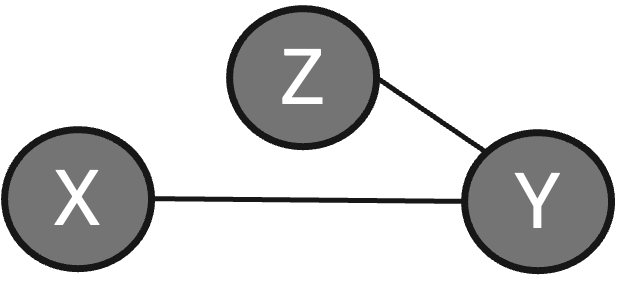} \end{minipage}         & 0.50         & 0.67         & 0.57         & 3          \\
VARLiNGAM &                   & \begin{minipage}{.17\linewidth} \centering \includegraphics[width=\linewidth]{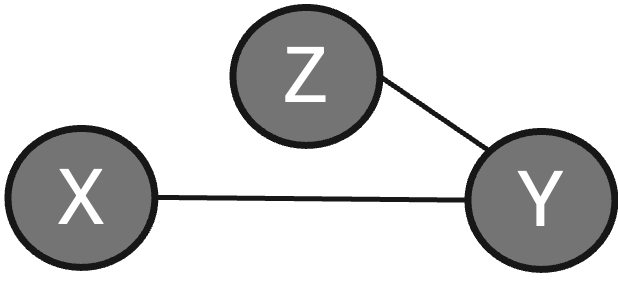} \end{minipage} & 0.50         & 0.67         & 0.57         & 3          \\
Granger   &                   & \begin{minipage}{.17\linewidth} \centering \includegraphics[width=\linewidth]{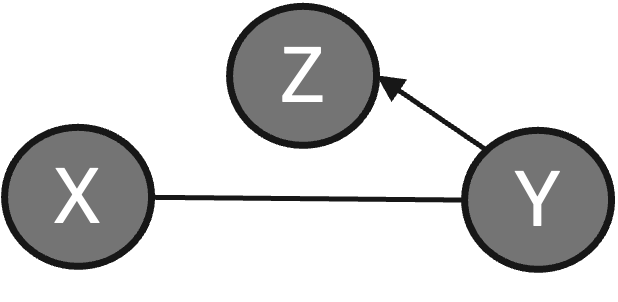} \end{minipage} & 0.67         & 0.67         & 0.67         & 2          \\
PCMCI     &                   & \begin{minipage}{.17\linewidth} \centering \includegraphics[width=\linewidth]{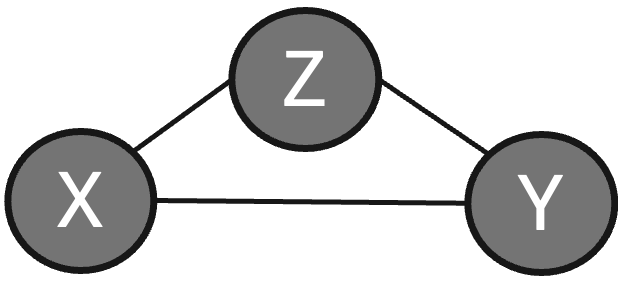} \end{minipage} & 0.50         & \textbf{1.0} & 0.67         & 3          \\
DYNOTEARS &                   & \begin{minipage}{.17\linewidth} \centering \includegraphics[width=\linewidth]{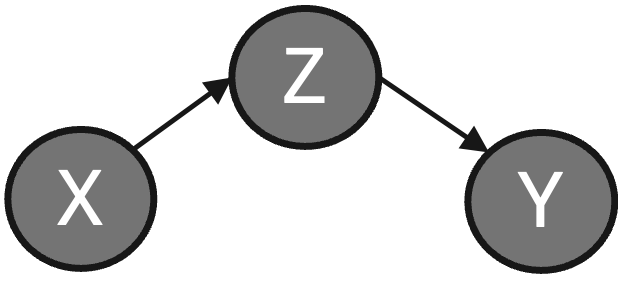} \end{minipage} & 0            & 0            & 0            & 5          \\
SLARAC    &                   & \begin{minipage}{.17\linewidth} \centering \includegraphics[width=\linewidth]{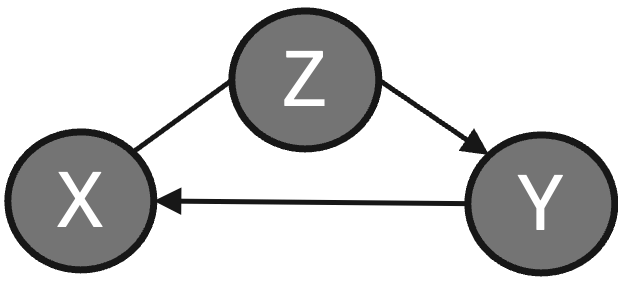} \end{minipage} & 0.25         & 0.33         & 0.29         & 5          \\ \cline{1-1} \cline{3-7} 
MXMap     &                   & \begin{minipage}{.17\linewidth} \centering \includegraphics[width=\linewidth]{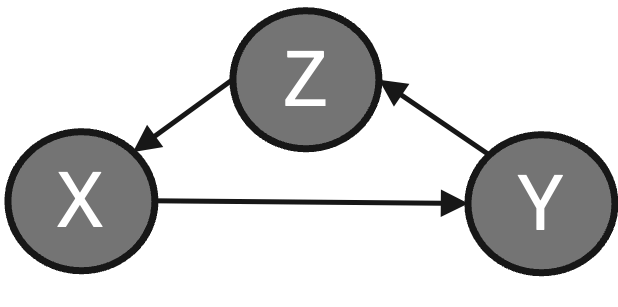} \end{minipage} & \textbf{1.0} & \textbf{1.0} & \textbf{1.0} & \textbf{0}
\end{tabular}
\caption{3V Cycle (No Noise)}
\label{tab:3V_Cycle_noN}
\end{table}

\begin{table}[htb]
\begin{tabular}{l|c|c|c|c|c|c}
Method    & Ground Truth      & Predicted & Precision    & Recall       & F1           & SHD        \\ \hline
tsFCI     & \multirow{7}{*}[-6.2em]{\begin{minipage}{.17\linewidth} \centering \includegraphics[width=\linewidth]{imgs/sim_new/gt/3V_Cycle_gt.png} \end{minipage}} & \begin{minipage}{.17\linewidth} \centering \includegraphics[width=\linewidth]{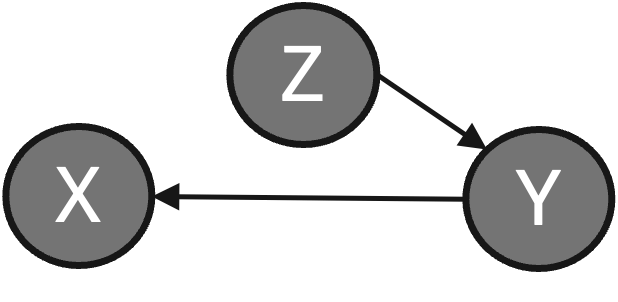} \end{minipage}    & 0            & 0            & 0            & 5          \\
VARLiNGAM &                   &  \begin{minipage}{.17\linewidth} \centering \includegraphics[width=\linewidth]{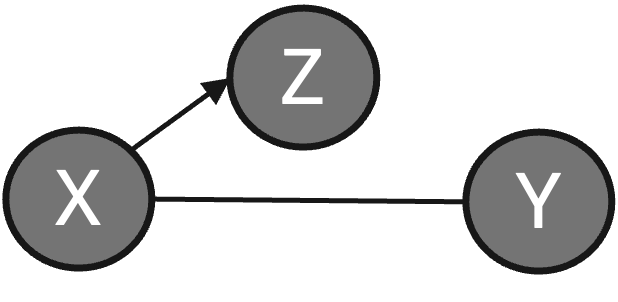} \end{minipage}  & 0.33         & 0.33         & 0.33         & 4          \\
Granger   &                   &  \begin{minipage}{.17\linewidth} \centering \includegraphics[width=\linewidth]{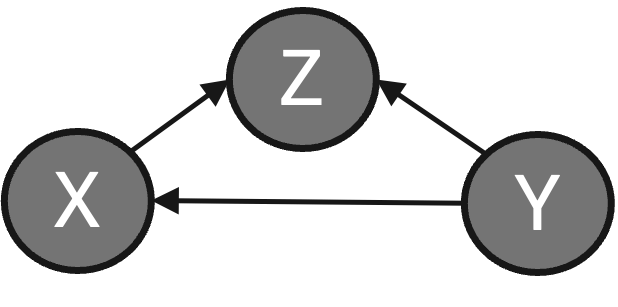} \end{minipage} & 0.33         & 0.33         & 0.33         & 4          \\
PCMCI     &                   &  \begin{minipage}{.17\linewidth} \centering \includegraphics[width=\linewidth]{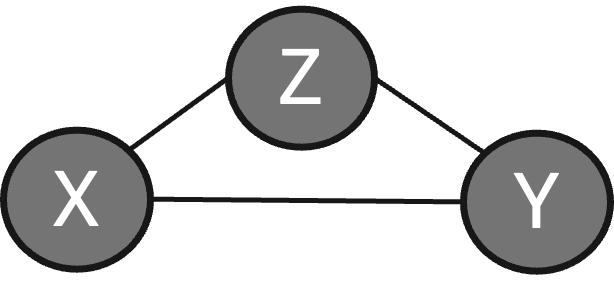} \end{minipage} & 0.50         & \textbf{1.0} & 0.67         & 3          \\
DYNOTEARS &                   & \begin{minipage}{.17\linewidth} \centering \includegraphics[width=\linewidth]{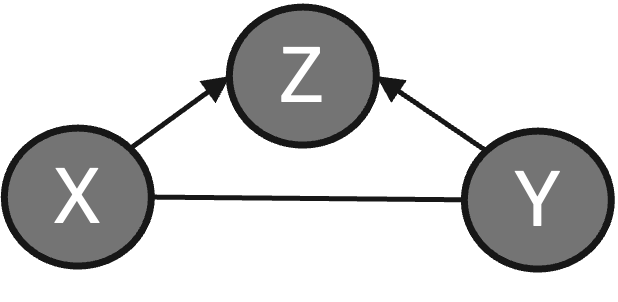} \end{minipage}& 0.50         & 0.65         & 0.57         & 3          \\
SLARAC    &                   & \begin{minipage}{.17\linewidth} \centering \includegraphics[width=\linewidth]{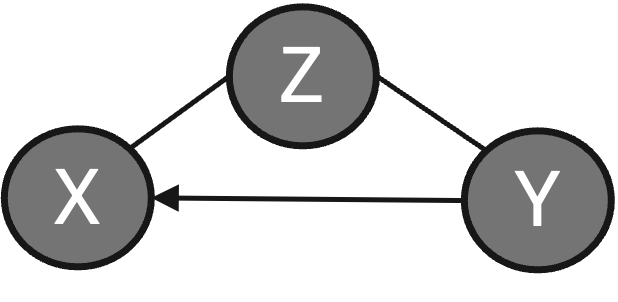} \end{minipage} & 0.40         & 0.67         & 0.50         & 4          \\ \cline{1-1} \cline{3-7} 
MXMap     &                   & \begin{minipage}{.17\linewidth} \centering \includegraphics[width=\linewidth]{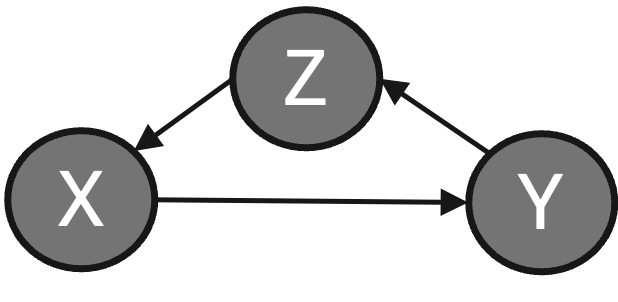} \end{minipage} & \textbf{1.0} & \textbf{1.0} & \textbf{1.0} & \textbf{0}
\end{tabular}
\caption{3V Cycle (Gaussian Additive Noise, Level 0.01)}
\label{tab:3V_Cycle_gN}
\end{table}

\begin{table}[htb]
\begin{tabular}{l|c|c|c|c|c|c}
Method    & Ground Truth      & Predicted & Precision    & Recall       & F1           & SHD        \\ \hline
tsFCI     & \multirow{7}{*}[-0.7em]{\begin{minipage}{.17\linewidth} \centering \includegraphics[width=\linewidth]{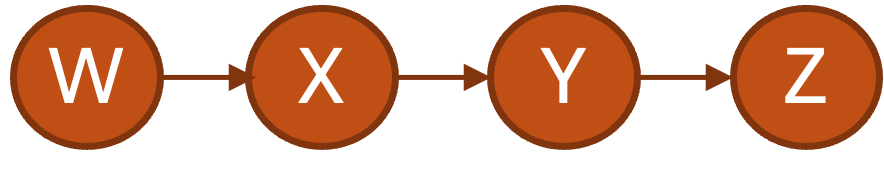} \end{minipage}} &  \begin{minipage}{.17\linewidth} \centering \includegraphics[width=\linewidth]{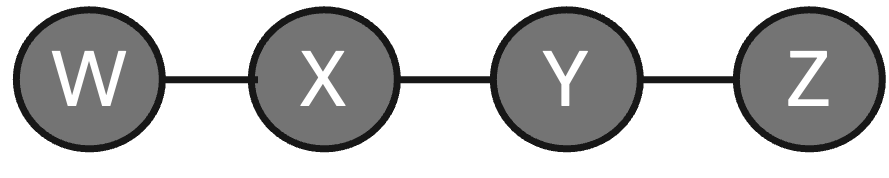} \end{minipage}         & 0.50         & \textbf{1.0} & 0.67         & 3          \\
VARLiNGAM &                   &  \begin{minipage}{.17\linewidth} \centering \includegraphics[width=\linewidth]{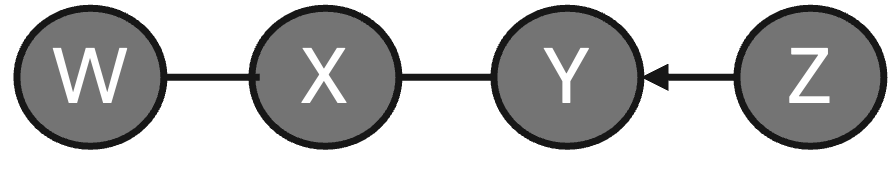} \end{minipage}     & 0.40         & 0.67         & 0.50         & 4          \\
Granger   &                   & \begin{minipage}{.17\linewidth} \centering \includegraphics[width=\linewidth]{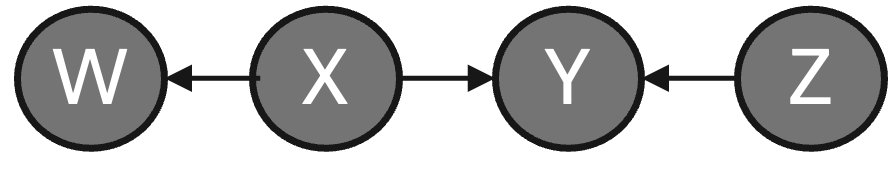} \end{minipage}    & 0.33         & 0.33         & 0.33         & 4          \\
PCMCI     &                   & \begin{minipage}{.17\linewidth} \centering \includegraphics[width=\linewidth]{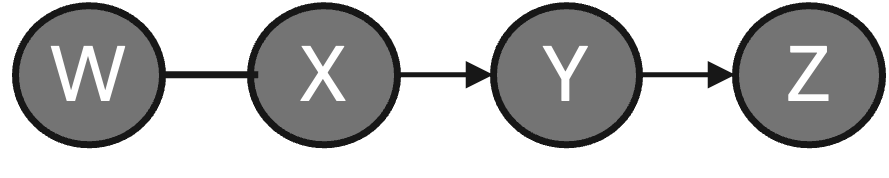} \end{minipage}    & 0.75         & \textbf{1.0} & 0.86         & 1          \\
DYNOTEARS &                   & \begin{minipage}{.17\linewidth} \centering \includegraphics[width=\linewidth]{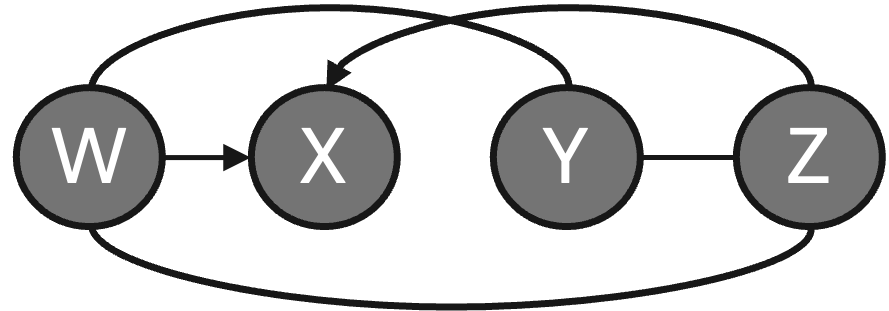} \end{minipage}     & 0.25         & 0.67         & 0.36         & 7          \\
SLARAC    &                   &           & 0.11         & 0.33         & 0.17         & 10         \\ \cline{1-1} \cline{3-7} 
MXMap     &                   & \begin{minipage}{.17\linewidth} \centering \includegraphics[width=\linewidth]{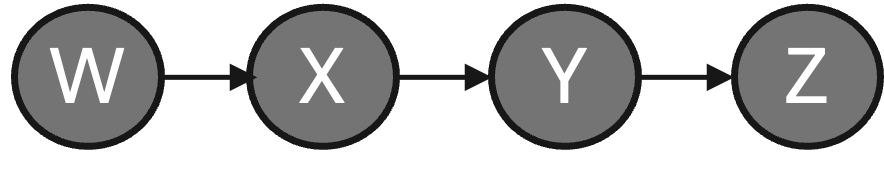} \end{minipage}     & \textbf{1.0} & \textbf{1.0} & \textbf{1.0} & \textbf{0}
\end{tabular}
\caption{4V Chain (No Noise)}
\label{tab:4V_Chain_noN}
\end{table}

\begin{table}[htb]
\begin{tabular}{l|c|c|c|c|c|c}
Method    & Ground Truth      & Predicted & Precision    & Recall       & F1           & SHD        \\ \hline
tsFCI     & \multirow{7}{*}[-0.7em]{\begin{minipage}{.17\linewidth} \centering \includegraphics[width=\linewidth]{imgs/sim_new/gt/4V_chain_gt.png} \end{minipage}} & \begin{minipage}{.17\linewidth} \centering \includegraphics[width=\linewidth]{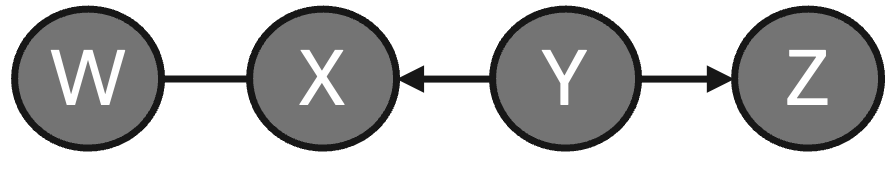} \end{minipage}          & 0.50         & 0.67         & 0.57         & 3          \\
VARLiNGAM &                   & \begin{minipage}{.17\linewidth} \centering \includegraphics[width=\linewidth]{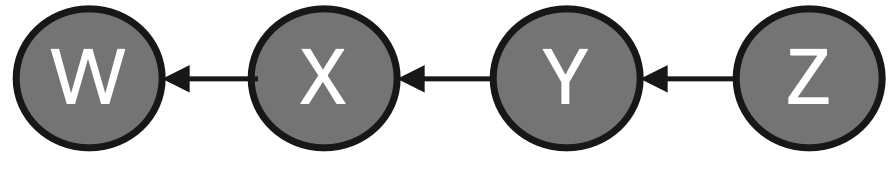} \end{minipage} & 0            & 0            & 0            & 6          \\
Granger   &                   & \begin{minipage}{.17\linewidth} \centering \includegraphics[width=\linewidth]{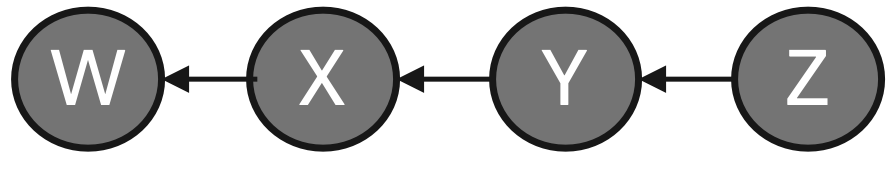} \end{minipage} & 0            & 0            & 0            & 6          \\
PCMCI     &                   & \begin{minipage}{.17\linewidth} \centering \includegraphics[width=\linewidth]{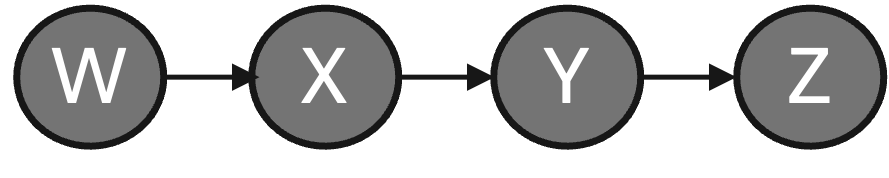} \end{minipage} & \textbf{1.0} & \textbf{1.0} & \textbf{1.0} & \textbf{0} \\
DYNOTEARS &                   & \begin{minipage}{.17\linewidth} \centering \includegraphics[width=\linewidth]{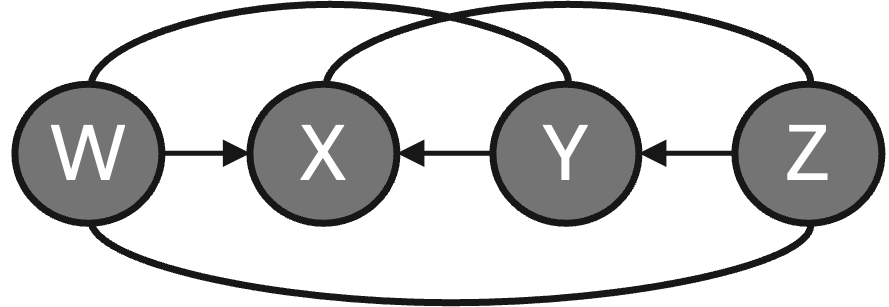} \end{minipage} & 0.11         & 0.33         & 0.17         & 10         \\
SLARAC    &                   & \begin{minipage}{.17\linewidth} \centering \includegraphics[width=\linewidth]{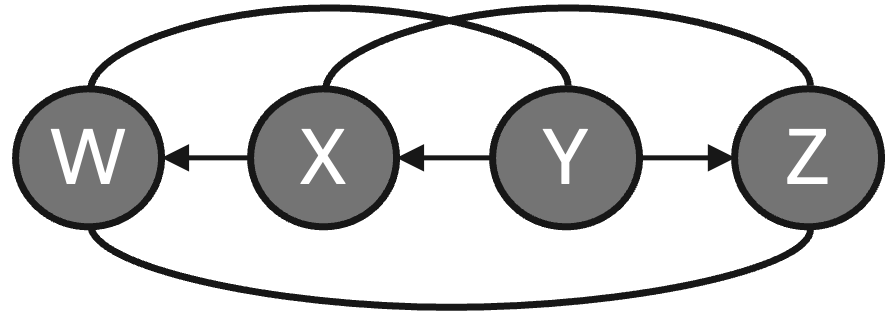} \end{minipage} & 0.11         & 0.33         & 0.17         & 10         \\ \cline{1-1} \cline{3-7} 
MXMap     &                   &  \begin{minipage}{.17\linewidth} \centering \includegraphics[width=\linewidth]{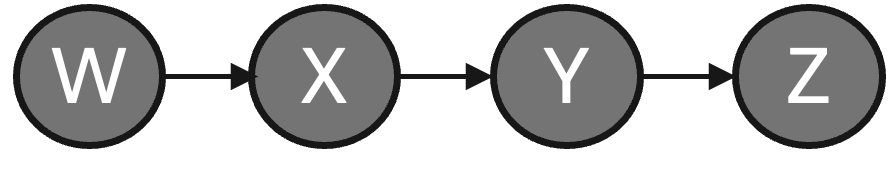} \end{minipage}   & \textbf{1.0} & \textbf{1.0} & \textbf{1.0} & \textbf{0}
\end{tabular}
\caption{4V Chain (Gaussian Additive Noise, Level 0.01)}
\label{tab:4V_Chain_gN}
\end{table}

\begin{table}[htb]
\begin{tabular}{l|c|c|c|c|c|c}
Method    & Ground Truth      & Predicted & Precision     & Recall       & F1            & SHD        \\ \hline
tsFCI     & \multirow{7}{*}[-5em]{\begin{minipage}{.17\linewidth} \centering \includegraphics[width=\linewidth]{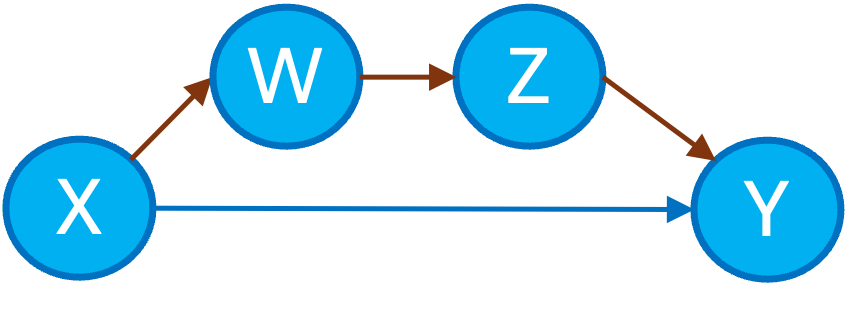} \end{minipage}} & \begin{minipage}{.17\linewidth} \centering \includegraphics[width=\linewidth]{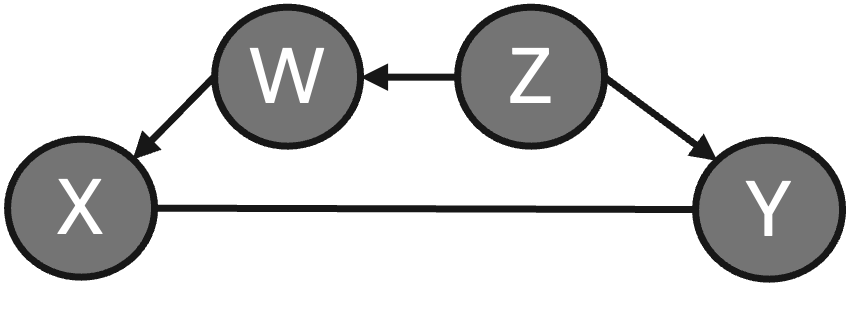} \end{minipage}   & 0.40          & 0.50         & 0.44          & 5          \\
VARLiNGAM &                   & \begin{minipage}{.17\linewidth} \centering \includegraphics[width=\linewidth]{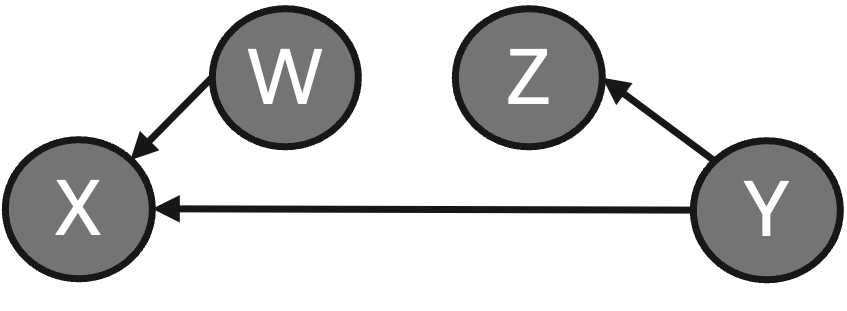} \end{minipage} & 0             & 0            & 0             & 7          \\
Granger   &                   & \begin{minipage}{.17\linewidth} \centering \includegraphics[width=\linewidth]{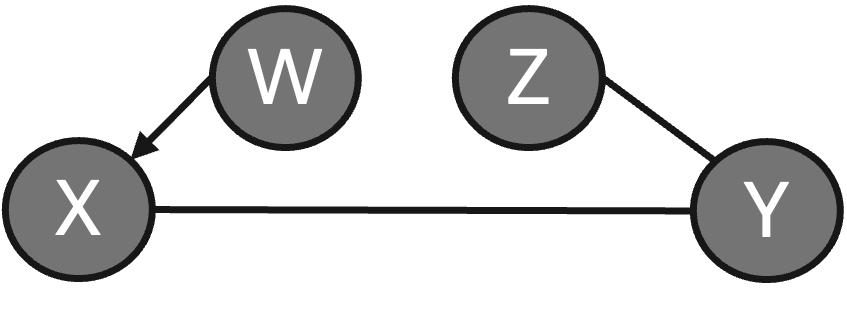} \end{minipage} & 0.40          & 0.50         & 0.44          & 5          \\
PCMCI     &                   & \begin{minipage}{.17\linewidth} \centering \includegraphics[width=\linewidth]{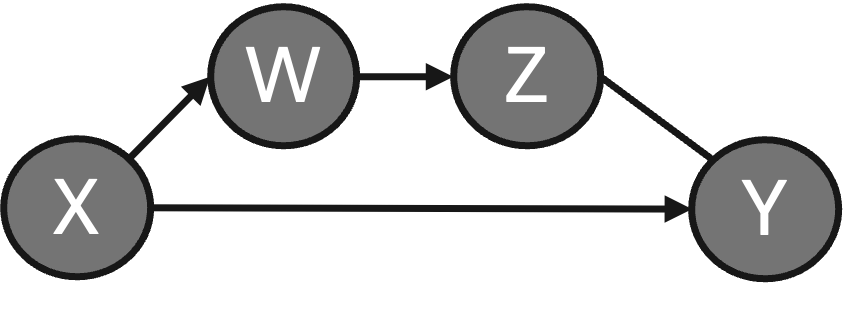} \end{minipage} & 0.80          & \textbf{1.0} & \textbf{0.89} & \textbf{1} \\
DYNOTEARS &                   & \begin{minipage}{.17\linewidth} \centering \includegraphics[width=\linewidth]{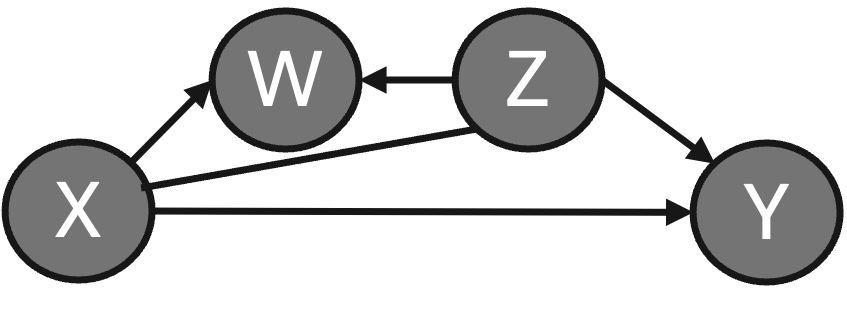} \end{minipage} & 0.60          & 0.75         & 0.67          & 3          \\
SLARAC    &                   & \begin{minipage}{.17\linewidth} \centering \includegraphics[width=\linewidth]{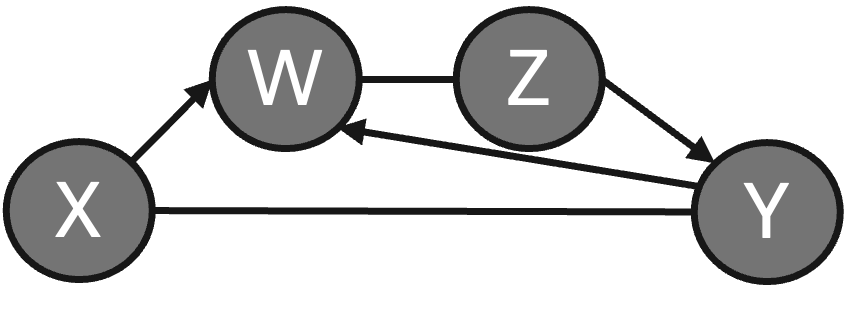} \end{minipage}  & 0.57          & \textbf{1.0} & 0.72          & 3          \\ \cline{1-1} \cline{3-7} 
MXMap     &                   &  \begin{minipage}{.17\linewidth} \centering \includegraphics[width=\linewidth]{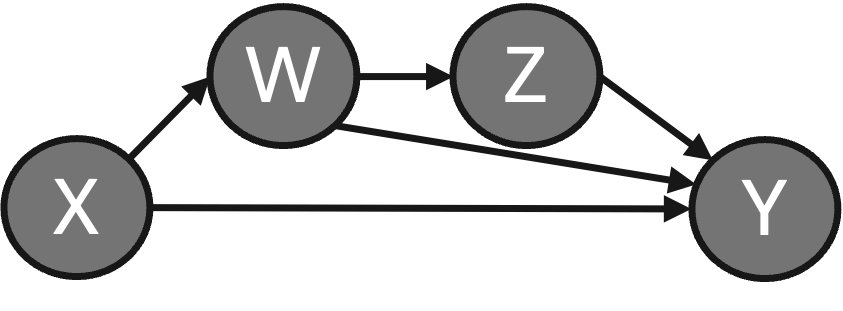} \end{minipage} & \textbf{0.80} & \textbf{1.0} & \textbf{0.89} & \textbf{1}
\end{tabular}
\caption{4V No Cycle (No Noise)}
\label{tab:4V_noCycle_noN}
\end{table}

\begin{table}[htb]
\begin{tabular}{l|c|c|c|c|c|c}
Method    & Ground Truth      & Predicted & Precision     & Recall       & F1            & SHD        \\ \hline
tsFCI     & \multirow{7}{*}[-5em]{\begin{minipage}{.17\linewidth} \centering \includegraphics[width=\linewidth]{imgs/sim_new/gt/4V_both_noCycle_gt.png} \end{minipage}} & \begin{minipage}{.17\linewidth} \centering \includegraphics[width=\linewidth]{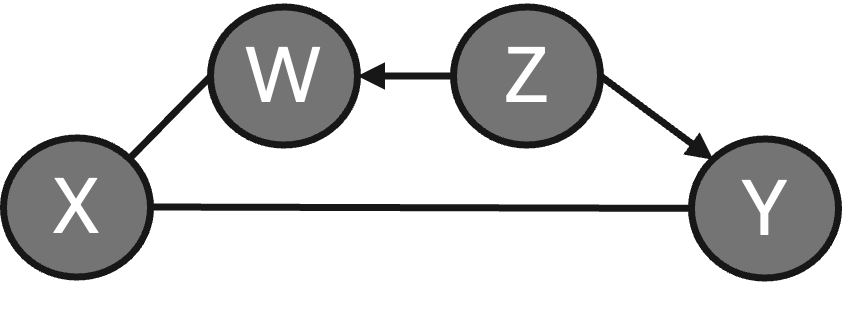} \end{minipage} & 0.50          & 0.75         & 0.60          & 4          \\
VARLiNGAM &                   & \begin{minipage}{.17\linewidth} \centering \includegraphics[width=\linewidth]{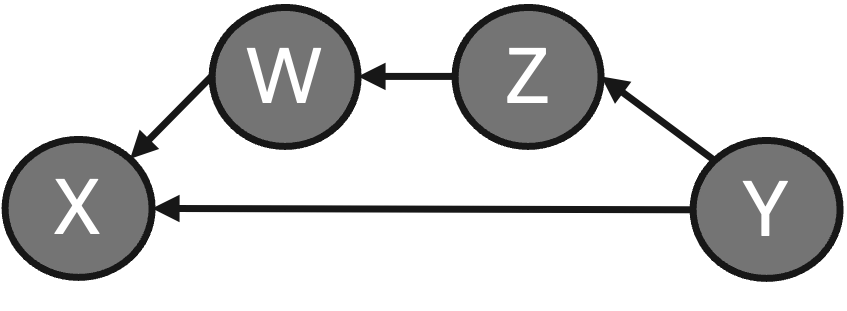} \end{minipage} & 0             & 0            & 0             & 8          \\
Granger   &                   & \begin{minipage}{.17\linewidth} \centering \includegraphics[width=\linewidth]{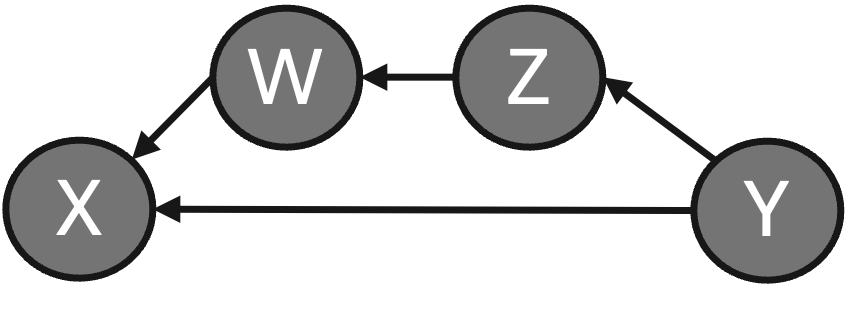} \end{minipage} & 0             & 0            & 0             & 8          \\
PCMCI     &                   & \begin{minipage}{.17\linewidth} \centering \includegraphics[width=\linewidth]{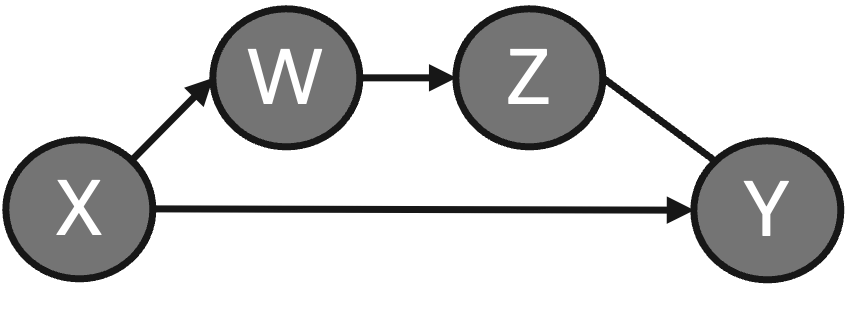} \end{minipage} & \textbf{0.80} & \textbf{1.0} & \textbf{0.89} & \textbf{1} \\
DYNOTEARS &                   & \begin{minipage}{.17\linewidth} \centering \includegraphics[width=\linewidth]{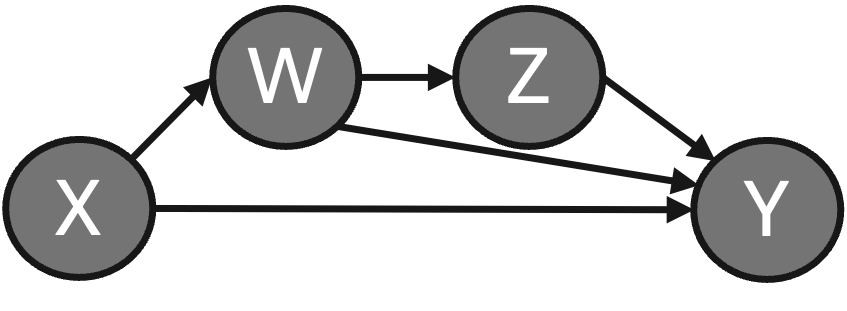} \end{minipage} & \textbf{0.80} & \textbf{1.0} & \textbf{0.89} & \textbf{1} \\
SLARAC    &                   &  \begin{minipage}{.17\linewidth} \centering \includegraphics[width=\linewidth]{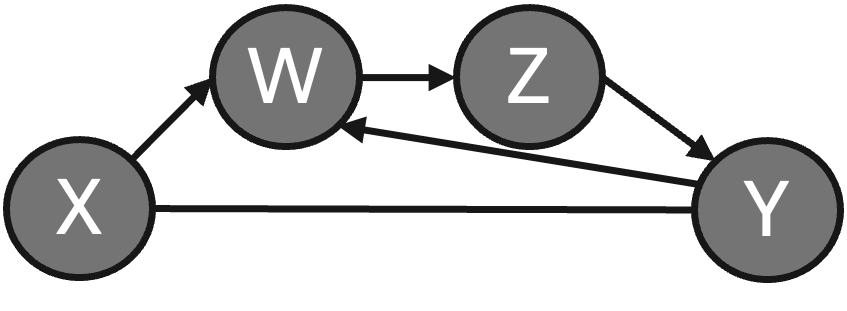} \end{minipage} & 0.67          & \textbf{1.0} & 0.8           & 2          \\ \cline{1-1} \cline{3-7} 
MXMap     &                   & \begin{minipage}{.17\linewidth} \centering \includegraphics[width=\linewidth]{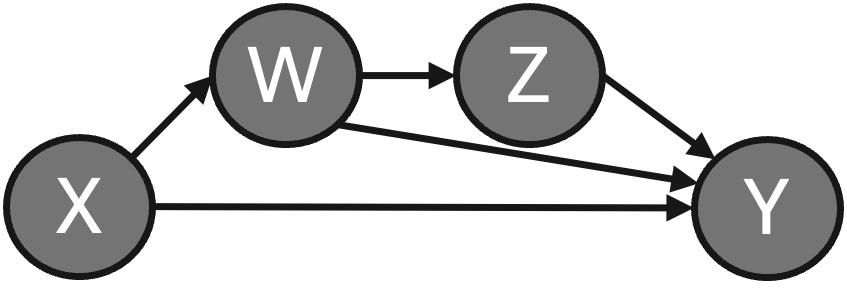} \end{minipage}  & \textbf{0.80} & \textbf{1.0} & \textbf{0.89} & \textbf{1}
\end{tabular}
\caption{4V No Cycle (Gaussian Additive Noise, Level 0.01)}
\label{tab:4V_noCycle_gN}
\end{table}

\begin{table}[htb]
\begin{tabular}{l|c|c|c|c|c|c}
Method    & Ground Truth      & Predicted & Precision    & Recall       & F1           & SHD        \\ \hline
tsFCI     & \multirow{7}{*}[-4.6em]{\begin{minipage}{.17\linewidth} \centering \includegraphics[width=\linewidth]{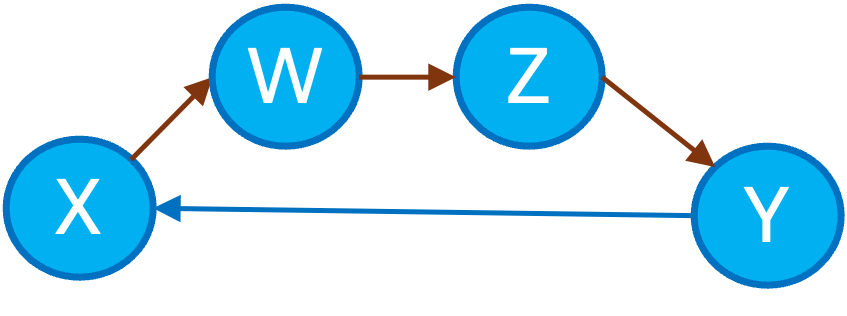} \end{minipage}} & \begin{minipage}{.17\linewidth} \centering \includegraphics[width=\linewidth]{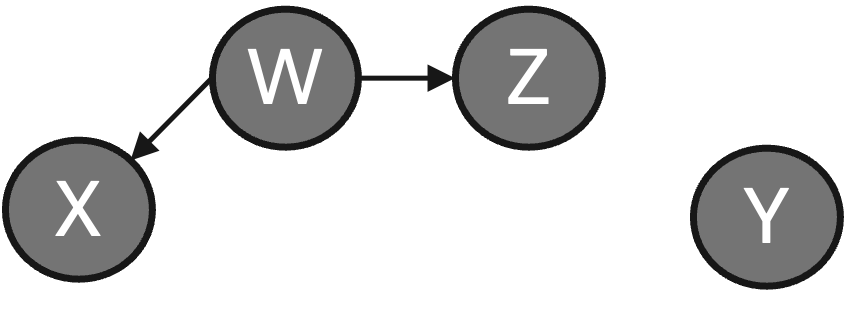} \end{minipage}    & 0.50         & 0.25         & 0.33         & 4          \\
VARLiNGAM &                   & \begin{minipage}{.17\linewidth} \centering \includegraphics[width=\linewidth]{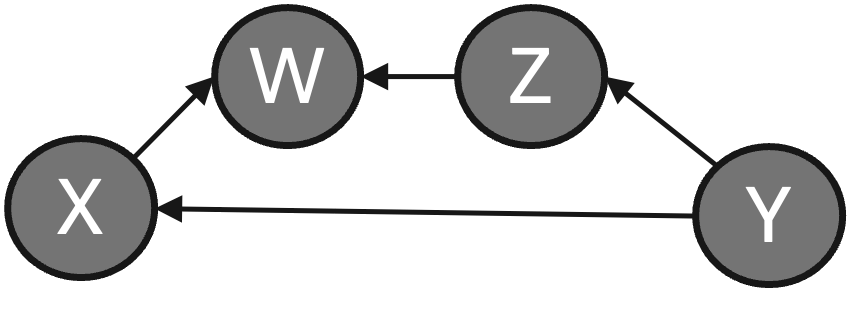} \end{minipage}  & 0.50         & 0.50         & 0.50         & 4          \\
Granger   &                   & \begin{minipage}{.17\linewidth} \centering \includegraphics[width=\linewidth]{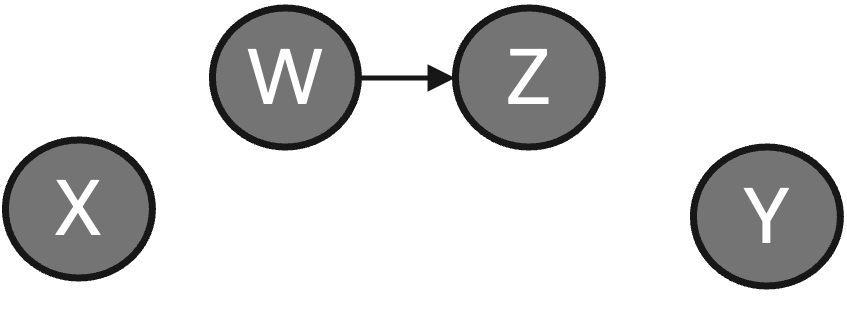} \end{minipage}  & \textbf{1.0} & 0.25         & 0.40         & 3          \\
PCMCI     &                   & \begin{minipage}{.17\linewidth} \centering \includegraphics[width=\linewidth]{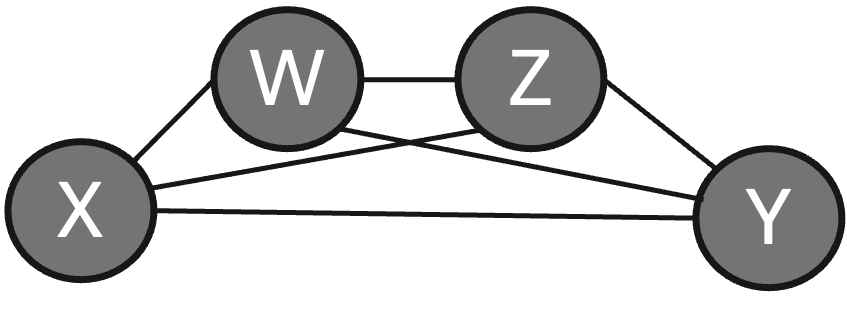} \end{minipage}  & 0.33         & \textbf{1.0} & 0.50         & 8          \\
DYNOTEARS &                   & \begin{minipage}{.17\linewidth} \centering \includegraphics[width=\linewidth]{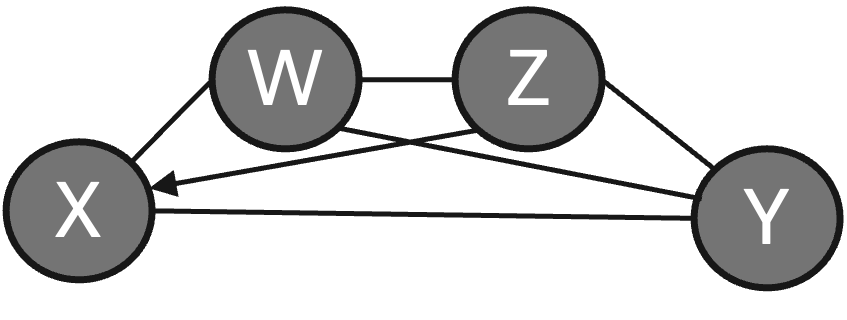} \end{minipage}  & 0.36         & \textbf{1.0} & 0.53         & 7          \\
SLARAC    &                   & \begin{minipage}{.17\linewidth} \centering \includegraphics[width=\linewidth]{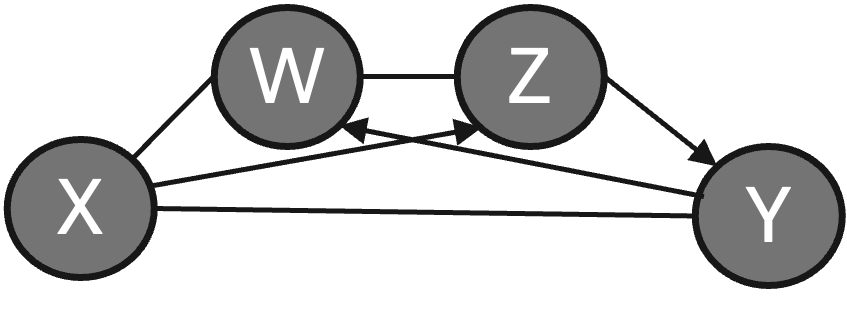} \end{minipage}   & 0.44         & \textbf{1.0} & 0.62         & 5          \\ \cline{1-1} \cline{3-7} 
MXMap     &                   & \begin{minipage}{.17\linewidth} \centering \includegraphics[width=\linewidth]{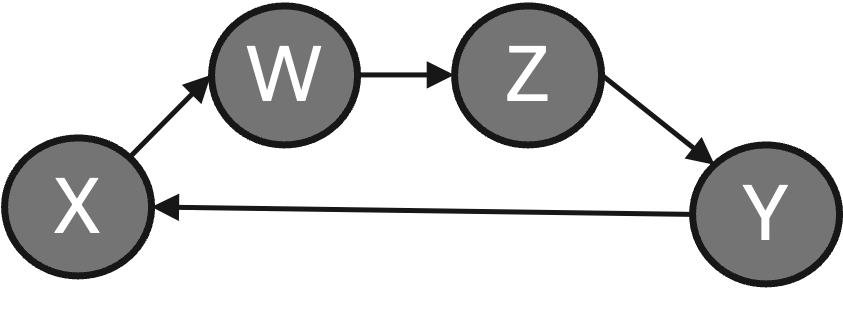} \end{minipage}  & \textbf{1.0} & \textbf{1.0} & \textbf{1.0} & \textbf{0}
\end{tabular}
\caption{4V Cycle (No Noise)}
\label{tab:4V_Cycle_noN}
\end{table}

\begin{table}[htb]
\begin{tabular}{l|c|c|c|c|c|c}
Method    & Ground Truth      & Predicted & Precision     & Recall       & F1            & SHD        \\ \hline
tsFCI     & \multirow{7}{*}[-4.6em]{\begin{minipage}{.17\linewidth} \centering \includegraphics[width=\linewidth]{imgs/sim_new/gt/4V_Cycle_gt.png} \end{minipage}} & \begin{minipage}{.17\linewidth} \centering \includegraphics[width=\linewidth]{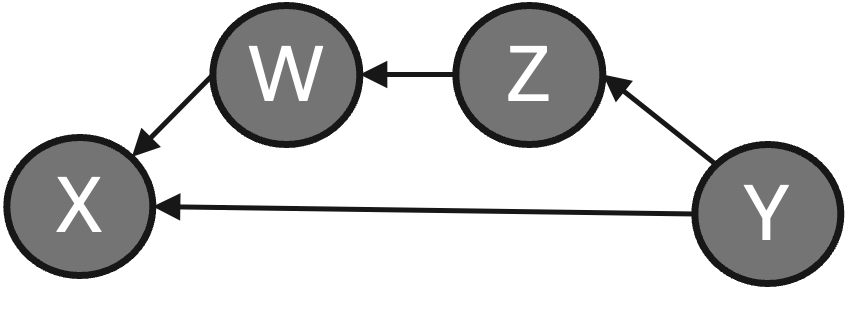} \end{minipage}   & 0.25          & 0.25         & 0.25          & 6          \\
VARLiNGAM &                   & \begin{minipage}{.17\linewidth} \centering \includegraphics[width=\linewidth]{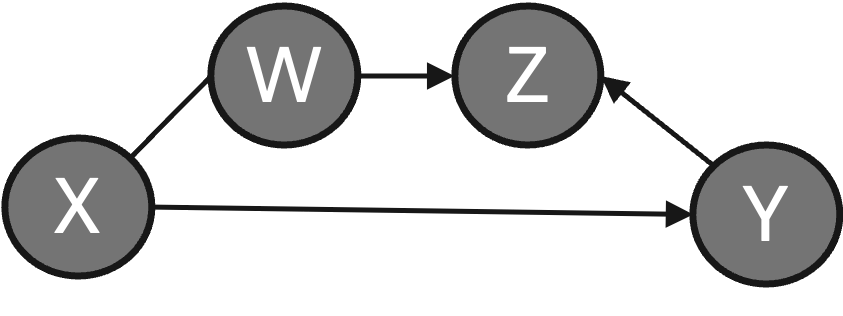} \end{minipage}   & 0.40          & 0.50         & 0.44          & 5          \\
Granger   &                   &  \begin{minipage}{.17\linewidth} \centering \includegraphics[width=\linewidth]{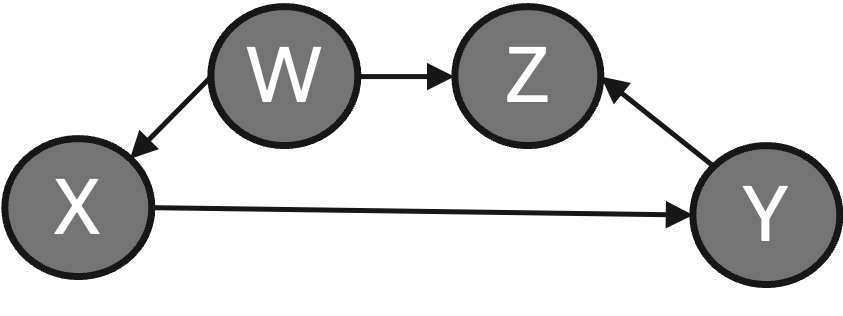} \end{minipage}  & 0.25          & 0.25         & 0.25          & 6          \\
PCMCI     &                   & \begin{minipage}{.17\linewidth} \centering \includegraphics[width=\linewidth]{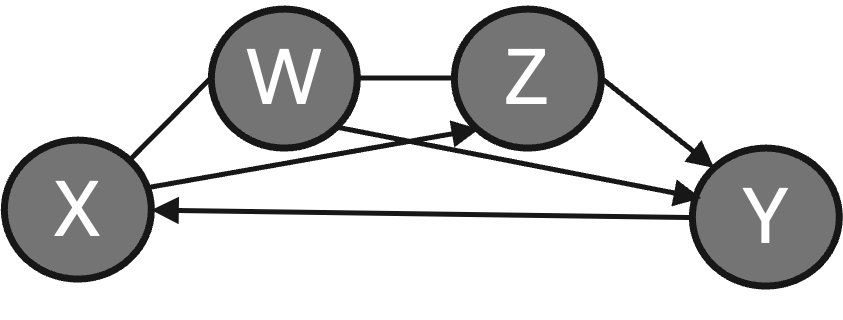} \end{minipage}   & 0.50          & \textbf{1.0} & 0.67          & 4          \\
DYNOTEARS &                   & \begin{minipage}{.17\linewidth} \centering \includegraphics[width=\linewidth]{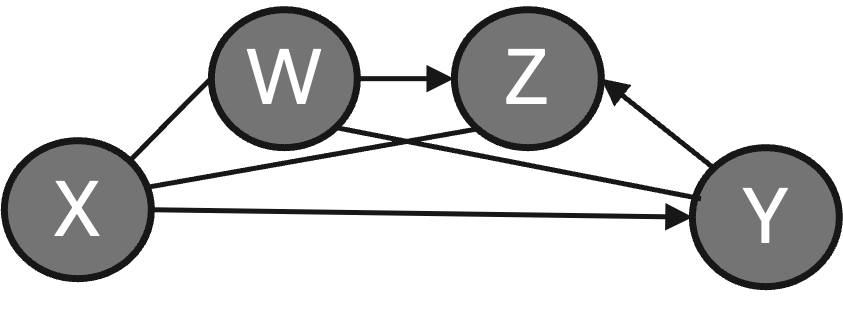} \end{minipage}    & 0.38          & 0.75         & 0.50          & 6          \\
SLARAC    &                   &   \begin{minipage}{.17\linewidth} \centering \includegraphics[width=\linewidth]{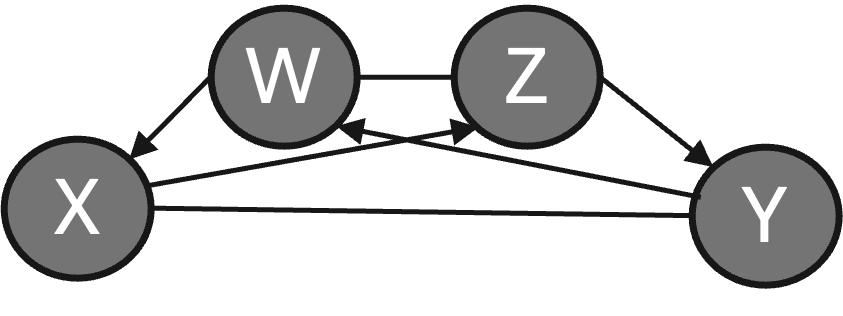} \end{minipage}    & 0.38          & 0.75         & 0.50          & 6          \\ \cline{1-1} \cline{3-7} 
MXMap     &                   & \begin{minipage}{.17\linewidth} \centering \includegraphics[width=\linewidth]{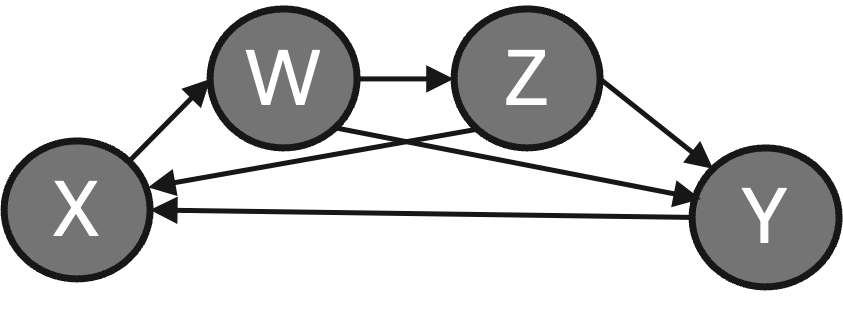} \end{minipage}    & \textbf{0.67} & \textbf{1.0} & \textbf{0.80} & \textbf{2}
\end{tabular}
\caption{4V Cycle (Gaussian Additive Noise, Level 0.01)}
\label{tab:4V_Cycle_gN}
\end{table}

\begin{table}[htb]
\begin{tabular}{l|c|c|c|c|c|c}
Method    & Ground Truth      & Predicted & Precision     & Recall       & F1            & SHD        \\ \hline
tsFCI     & \multirow{7}{*}[-9.6em]{\begin{minipage}{.17\linewidth} \centering \includegraphics[width=\linewidth]{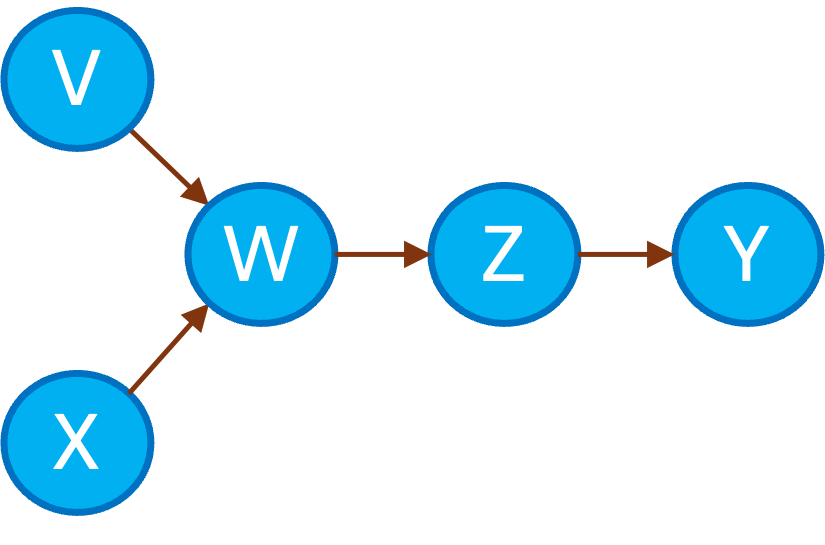} \end{minipage}} &  \begin{minipage}{.17\linewidth} \centering \includegraphics[width=\linewidth]{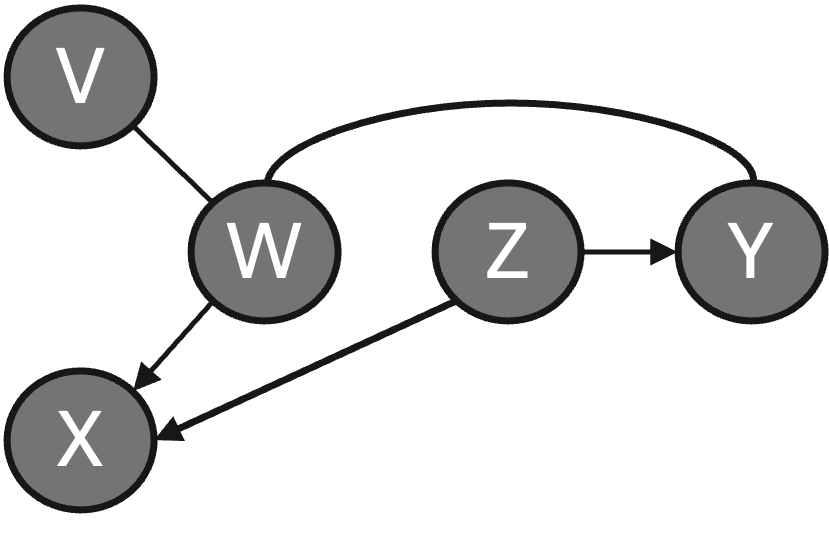} \end{minipage}& 0.29          & 0.50         & 0.36          & 7          \\
VARLiNGAM &                   & \begin{minipage}{.17\linewidth} \centering \includegraphics[width=\linewidth]{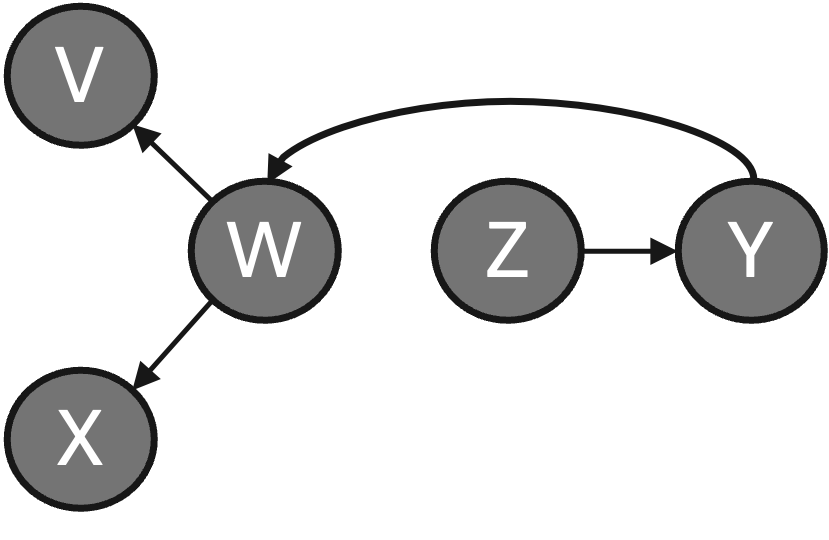} \end{minipage} & 0.25          & 0.25         & 0.25          & 6          \\
Granger   &                   &  \begin{minipage}{.17\linewidth} \centering \includegraphics[width=\linewidth]{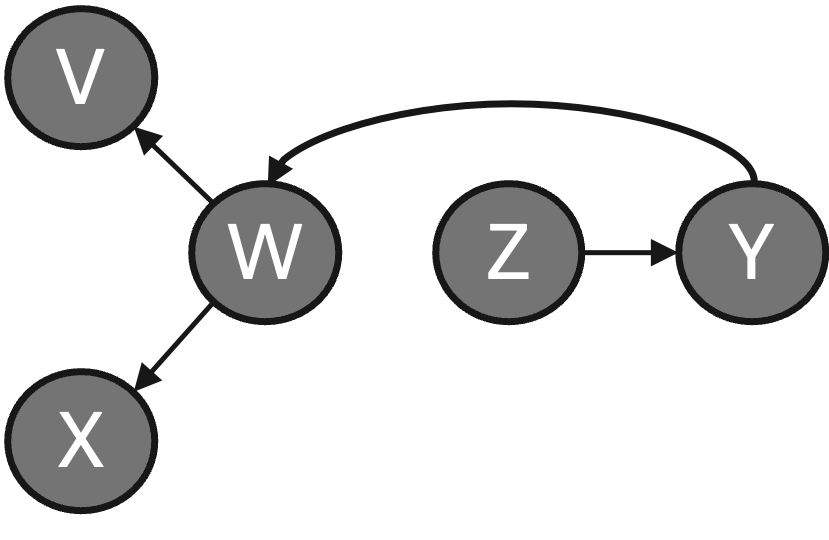} \end{minipage} & 0.25          & 0.25         & 0.25          & 6          \\
PCMCI     &                   &  \begin{minipage}{.17\linewidth} \centering \includegraphics[width=\linewidth]{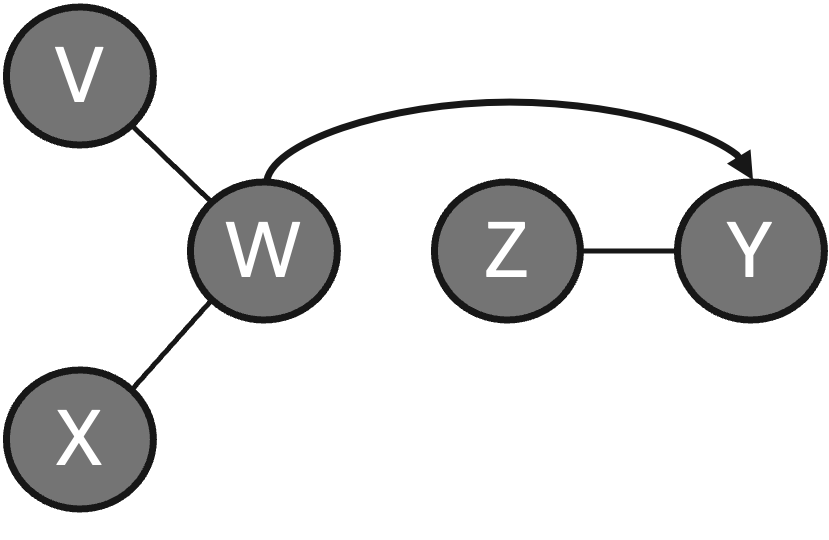} \end{minipage}   & 0.42          & 0.75         & 0.55          & 5          \\
DYNOTEARS &                   &  \begin{minipage}{.17\linewidth} \centering \includegraphics[width=\linewidth]{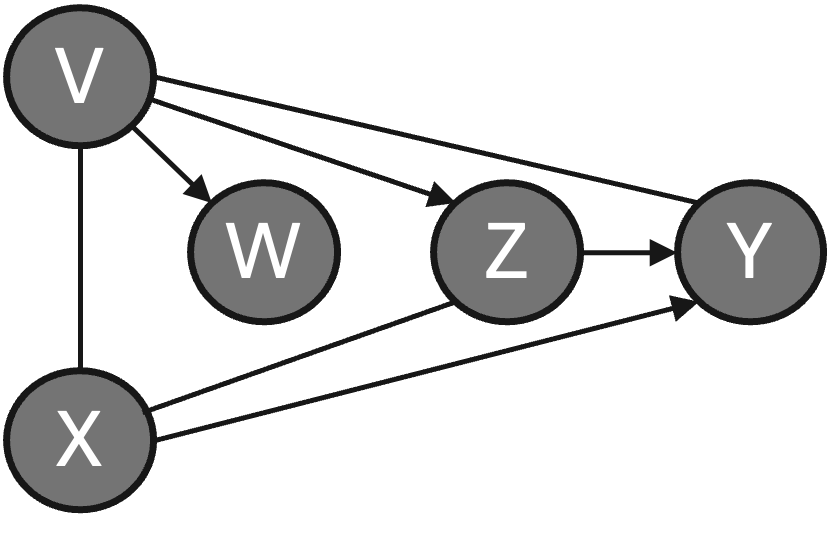} \end{minipage}  & 0.25          & 0.50         & 0.33          & 8          \\
SLARAC    &                   &  \begin{minipage}{.17\linewidth} \centering \includegraphics[width=\linewidth]{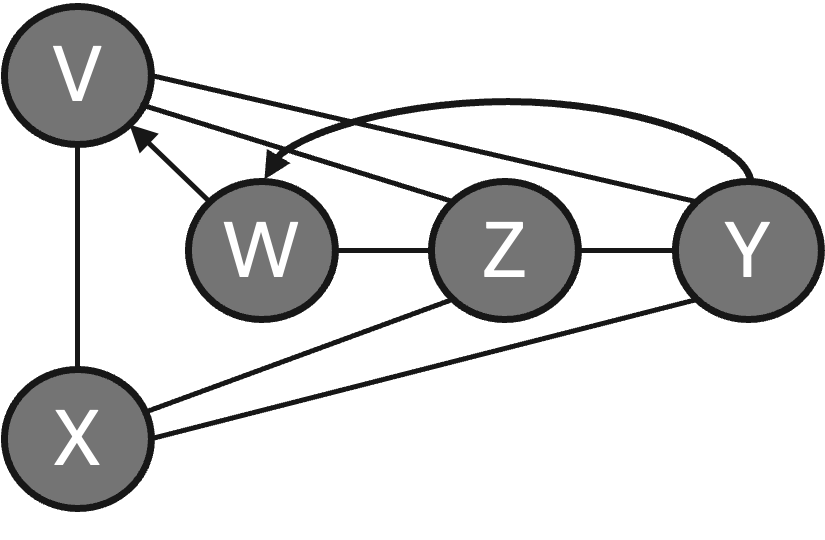} \end{minipage}  & 0.13          & 0.50         & 0.20          & 16         \\ \cline{1-1} \cline{3-7} 
MXMap     &                   &  \begin{minipage}{.17\linewidth} \centering \includegraphics[width=\linewidth]{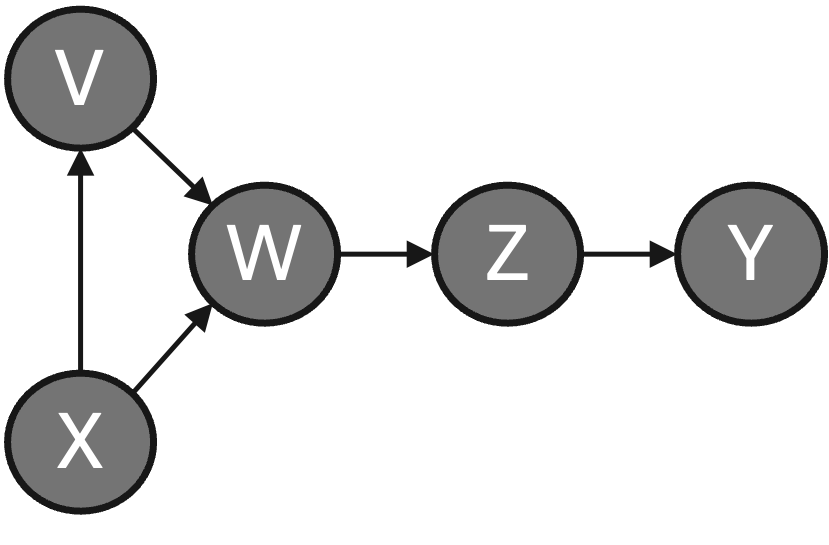} \end{minipage} & \textbf{0.80} & \textbf{1.0} & \textbf{0.89} & \textbf{1}
\end{tabular}
\caption{5V Structure 1 Without Cycle (No Noise)}
\label{tab:5V1_noN}
\end{table}

\begin{table}[htb]
\begin{tabular}{l|c|c|c|c|c|c}
Method    & Ground Truth      & Predicted & Precision     & Recall       & F1            & SHD        \\ \hline
tsFCI     & \multirow{7}{*}[-9.6em]{\begin{minipage}{.17\linewidth} \centering \includegraphics[width=\linewidth]{imgs/sim_new/gt/5V-1_gt.png} \end{minipage}} &  \begin{minipage}{.17\linewidth} \centering \includegraphics[width=\linewidth]{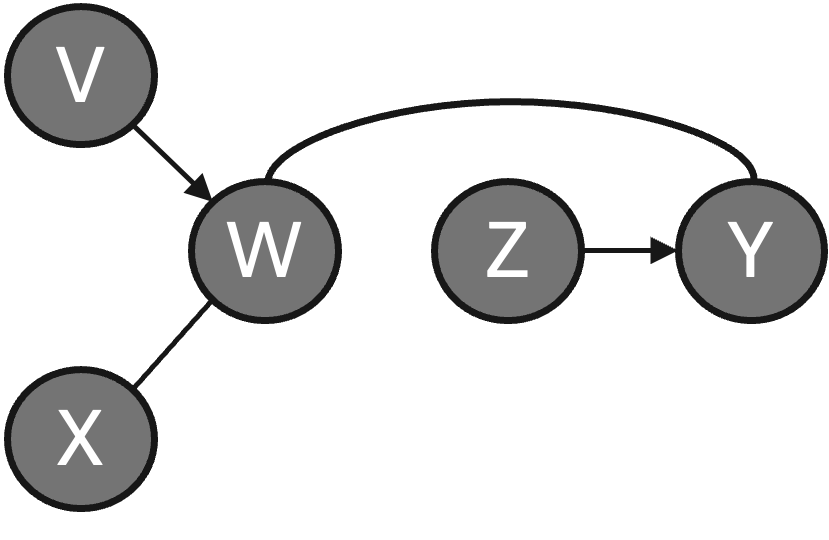} \end{minipage} & 0.50          & 0.75         & 0.60          & 4          \\
VARLiNGAM &                   & \begin{minipage}{.17\linewidth} \centering \includegraphics[width=\linewidth]{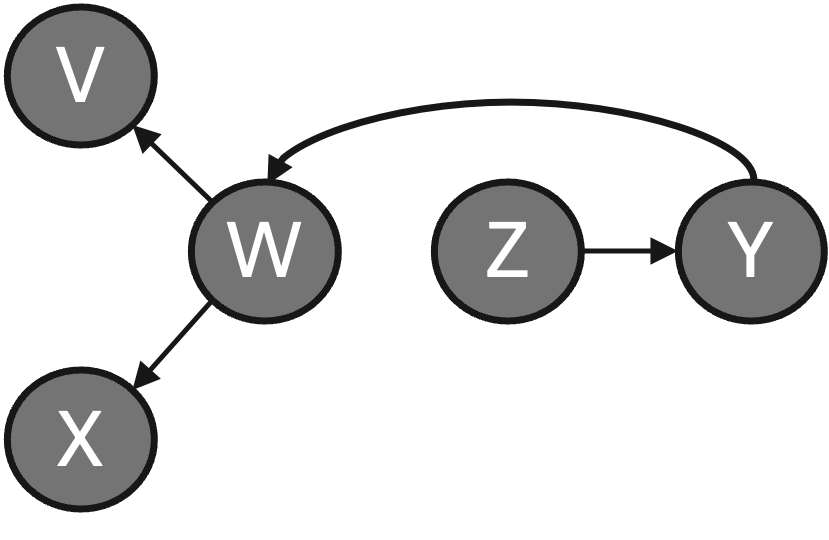} \end{minipage} & 0.25          & 0.25         & 0.25          & 6          \\
Granger   &                   & \begin{minipage}{.17\linewidth} \centering \includegraphics[width=\linewidth]{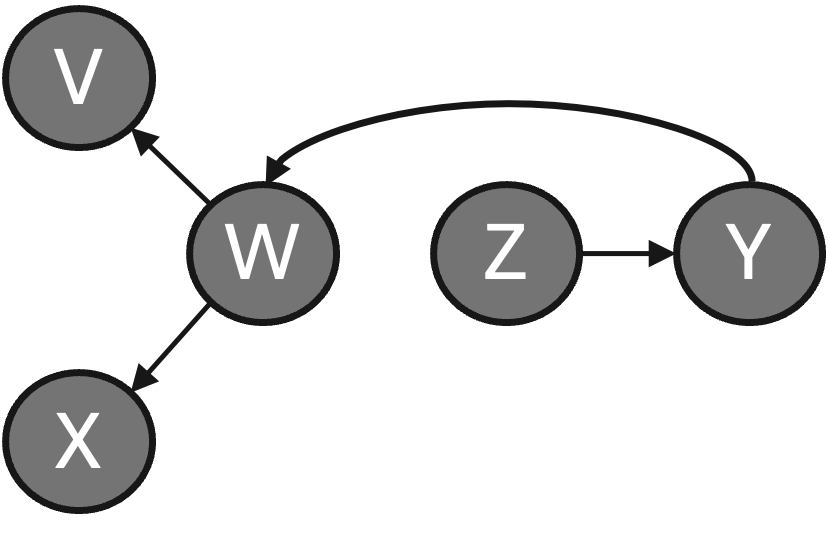} \end{minipage}  & 0.25          & 0.25         & 0.25          & 6          \\
PCMCI     &                   & \begin{minipage}{.17\linewidth} \centering \includegraphics[width=\linewidth]{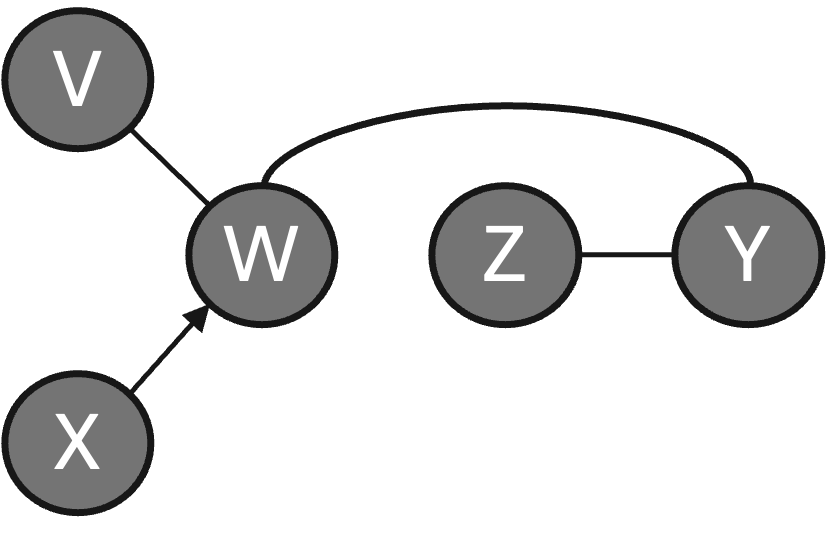} \end{minipage} & 0.43          & 0.75         & 0.55          & 5          \\
DYNOTEARS &                   & \begin{minipage}{.17\linewidth} \centering \includegraphics[width=\linewidth]{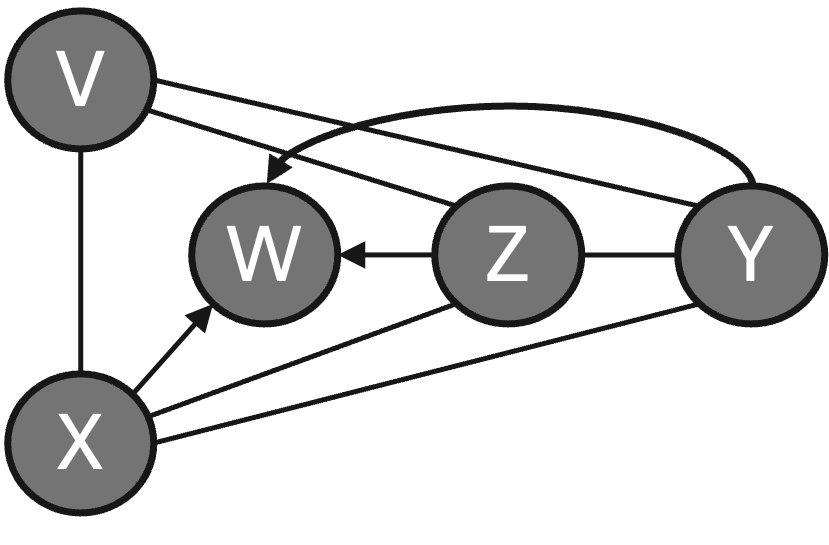} \end{minipage} & 0.13          & 0.50         & 0.21          & 15         \\
SLARAC    &                   &  \begin{minipage}{.17\linewidth} \centering \includegraphics[width=\linewidth]{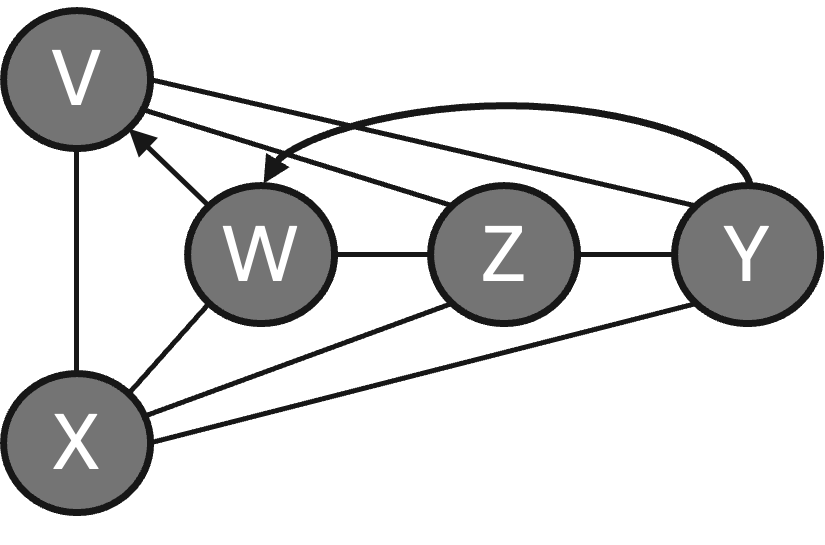} \end{minipage} & 0.17          & 0.75         & 0.27          & 16         \\ \cline{1-1} \cline{3-7} 
MXMap     &                   & \begin{minipage}{.17\linewidth} \centering \includegraphics[width=\linewidth]{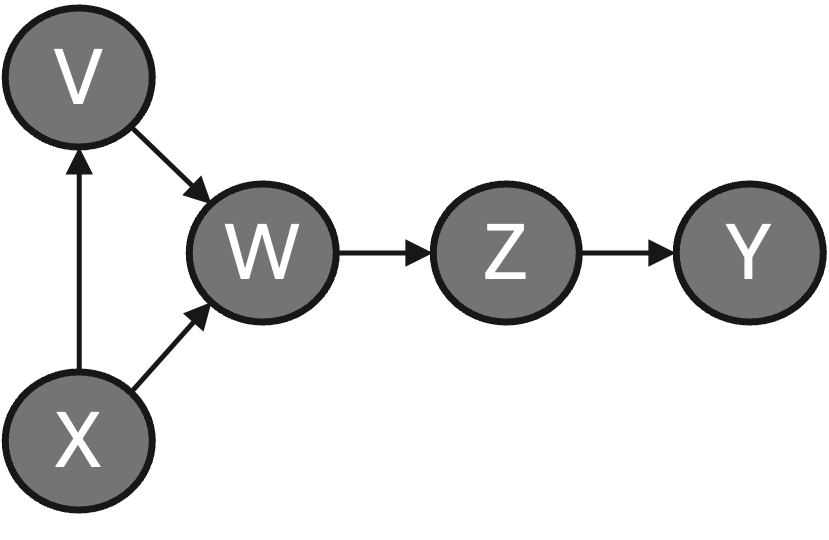} \end{minipage}  & \textbf{0.80} & \textbf{1.0} & \textbf{0.89} & \textbf{1}
\end{tabular}
\caption{5V Structure 1 Without Cycle (Gaussian Additive Noise, Level 0.01)}
\label{tab:5V1_gN}
\end{table}

\begin{table}[htb]
\begin{tabular}{l|c|c|c|c|c|c}
Method    & Ground Truth      & Predicted & Precision    & Recall       & F1           & SHD        \\ \hline
tsFCI     & \multirow{7}{*}[-8.6em]{\begin{minipage}{.17\linewidth} \centering \includegraphics[width=\linewidth]{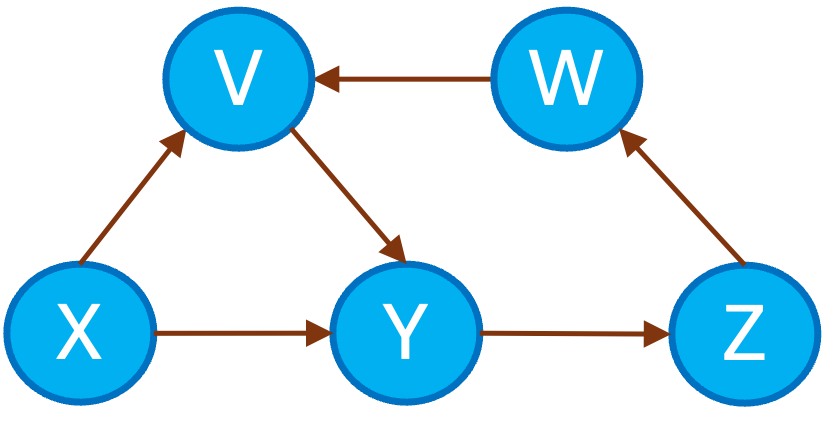} \end{minipage}} & \begin{minipage}{.17\linewidth} \centering \includegraphics[width=\linewidth]{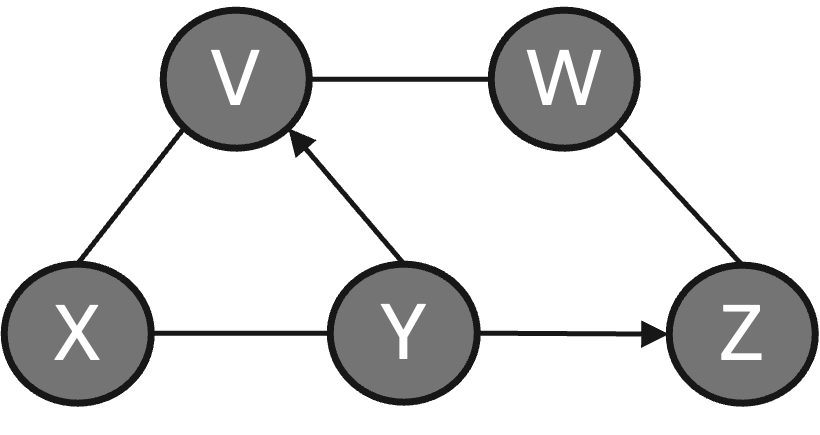} \end{minipage} & 0.50         & 0.83         & 0.64         & 6          \\
VARLiNGAM &                   & \begin{minipage}{.17\linewidth} \centering \includegraphics[width=\linewidth]{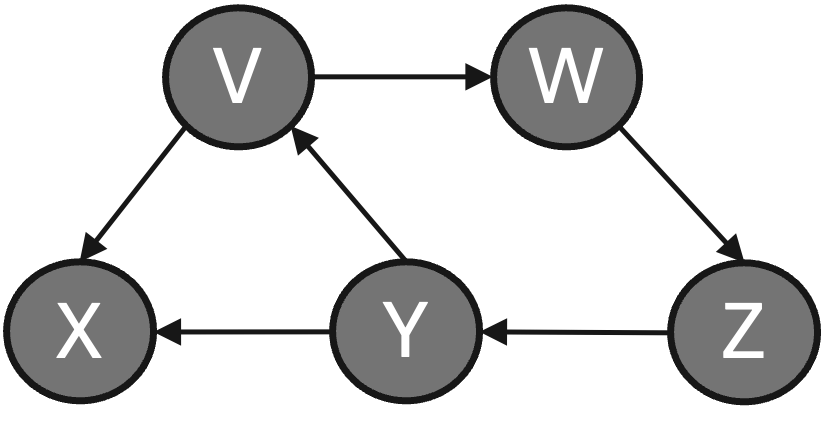} \end{minipage} & 0            & 0            & 0            & 12         \\
Granger   &                   &  \begin{minipage}{.17\linewidth} \centering \includegraphics[width=\linewidth]{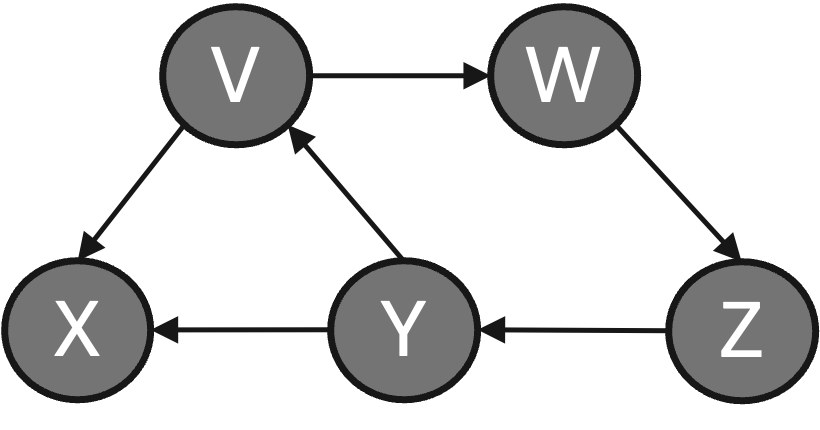} \end{minipage}  & 0            & 0            & 0            & 12         \\
PCMCI     &                   &  \begin{minipage}{.17\linewidth} \centering \includegraphics[width=\linewidth]{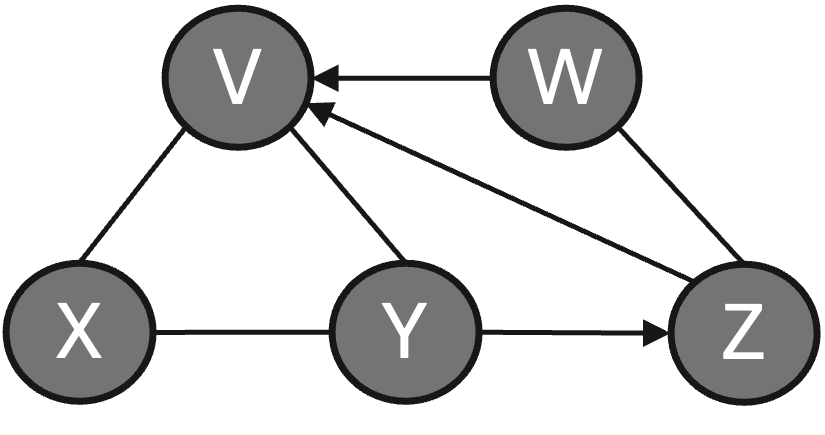} \end{minipage} & 0.55         & \textbf{1.0} & 0.71         & 5          \\
DYNOTEARS &                   &  \begin{minipage}{.17\linewidth} \centering \includegraphics[width=\linewidth]{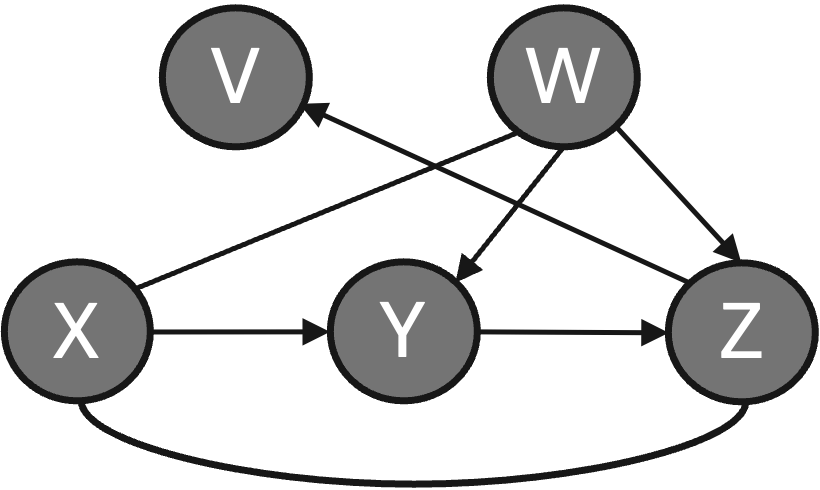} \end{minipage}  & 0.22         & 0.33         & 0.27         & 11         \\
SLARAC    &                   &  \begin{minipage}{.17\linewidth} \centering \includegraphics[width=\linewidth]{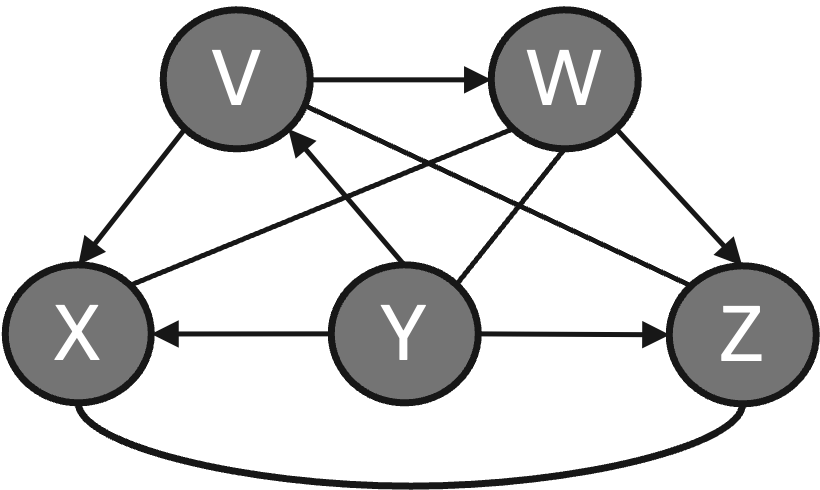} \end{minipage} & 0.07         & 0.17         & 0.10         & 18         \\ \cline{1-1} \cline{3-7} 
MXMap     &                   &  \begin{minipage}{.17\linewidth} \centering \includegraphics[width=\linewidth]{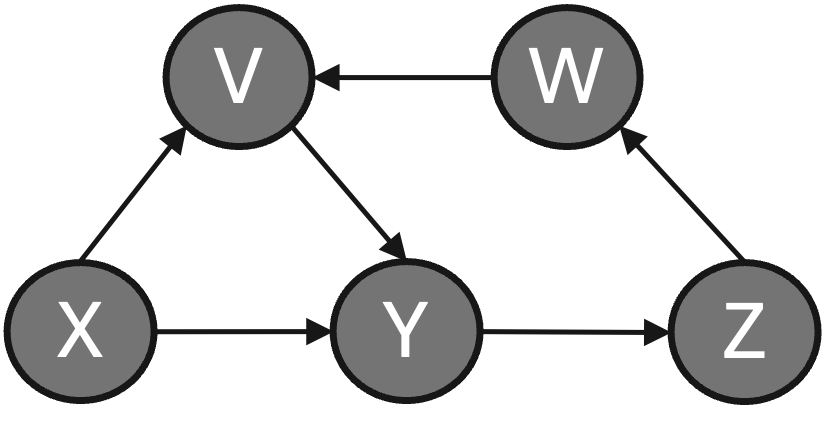} \end{minipage}  & \textbf{1.0} & \textbf{1.0} & \textbf{1.0} & \textbf{0}
\end{tabular}
\caption{5V Structure 2 With Cycle (No Noise)}
\label{tab:5V2_noN}
\end{table}

\begin{table}[htb]
\begin{tabular}{l|c|c|c|c|c|c}
Method    & Ground Truth      & Predicted & Precision     & Recall       & F1            & SHD        \\ \hline
tsFCI     & \multirow{7}{*}[-9.6em]{\begin{minipage}{.17\linewidth} \centering \includegraphics[width=\linewidth]{imgs/sim_new/gt/5V-2_gt.png} \end{minipage}} & \begin{minipage}{.17\linewidth} \centering \includegraphics[width=\linewidth]{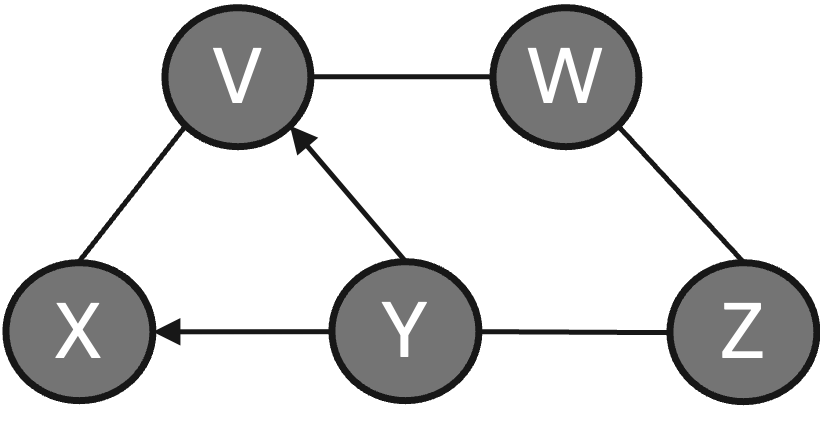} \end{minipage} & 0.40          & 0.67         & 0.50          & 8          \\
VARLiNGAM &                   &  \begin{minipage}{.17\linewidth} \centering \includegraphics[width=\linewidth]{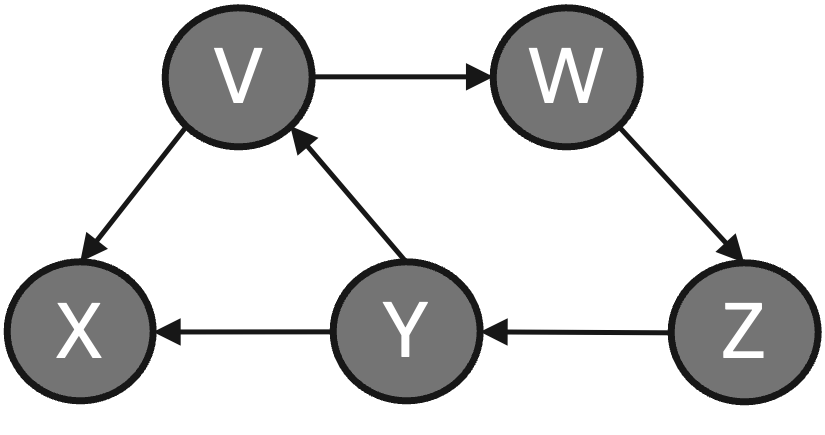} \end{minipage} & 0             & 0            & 0             & 12         \\
Granger   &                   &  \begin{minipage}{.17\linewidth} \centering \includegraphics[width=\linewidth]{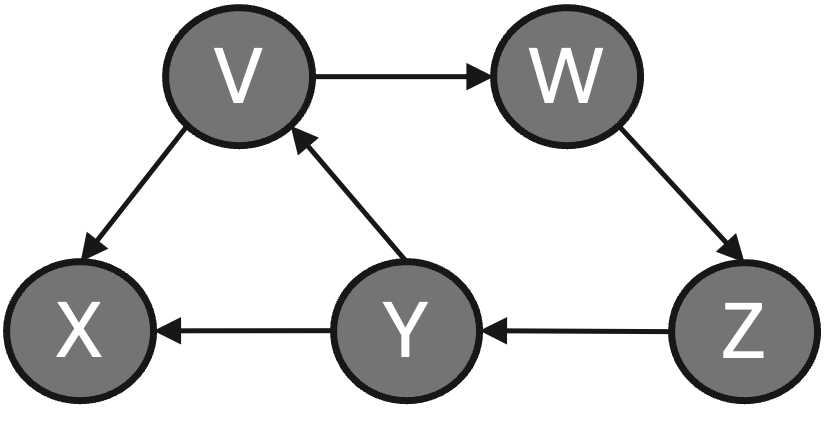} \end{minipage}   & 0             & 0            & 0             & 12         \\
PCMCI     &                   & \begin{minipage}{.17\linewidth} \centering \includegraphics[width=\linewidth]{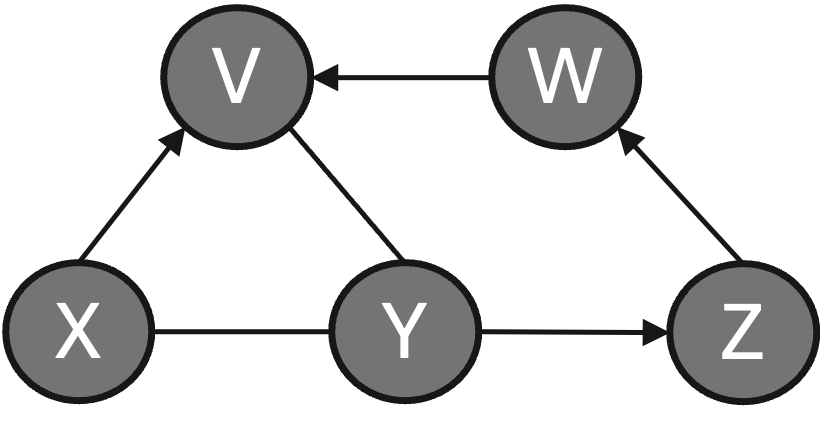} \end{minipage} & \textbf{0.75} & \textbf{1.0} & \textbf{0.86} & \textbf{2} \\
DYNOTEARS &                   &  \begin{minipage}{.17\linewidth} \centering \includegraphics[width=\linewidth]{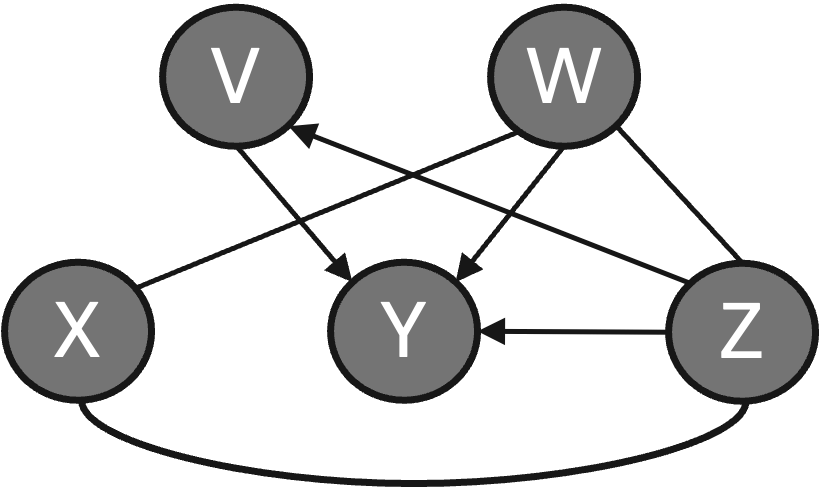} \end{minipage} & 0.20          & 0.33         & 0.25          & 12         \\
SLARAC    &                   &  \begin{minipage}{.17\linewidth} \centering \includegraphics[width=\linewidth]{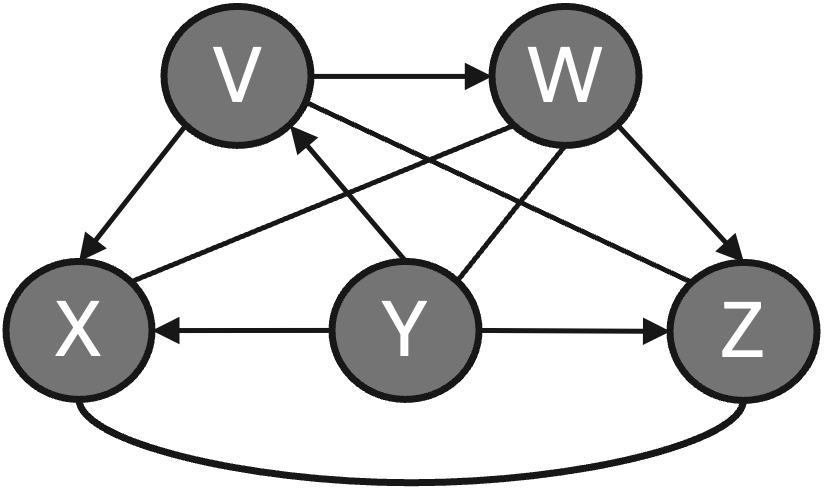} \end{minipage}  & 0.07          & 0.17         & 0.10          & 18         \\ \cline{1-1} \cline{3-7} 
MXMap     &                   & \begin{minipage}{.17\linewidth} \centering \includegraphics[width=\linewidth]{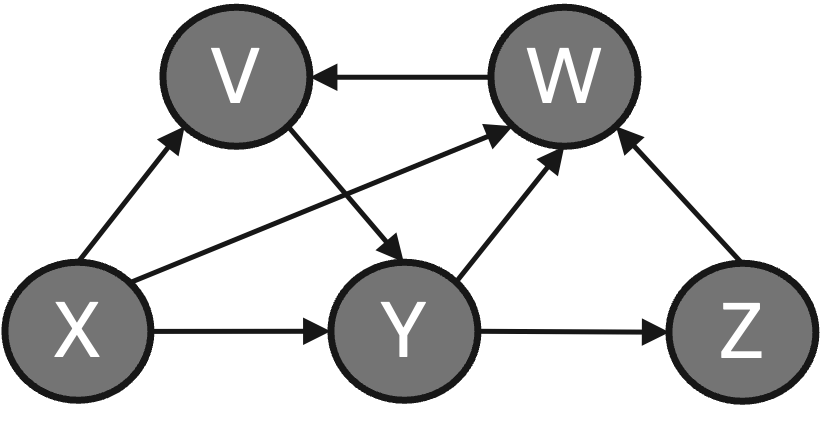} \end{minipage} & \textbf{0.75} & \textbf{1.0} & \textbf{0.86} & \textbf{2}
\end{tabular}
\caption{5V Structure 2 With Cycle (Gaussian Additive Noise, Level 0.01)}
\label{tab:5V2_gN}
\end{table}

\begin{table}[htb]
\begin{tabular}{l|c|c|c|c|c|c}
Method    & Ground Truth      & Predicted & Precision     & Recall       & F1            & SHD        \\ \hline
tsFCI     & \multirow{7}{*}[-16em]{\begin{minipage}{.17\linewidth} \centering \includegraphics[width=\linewidth]{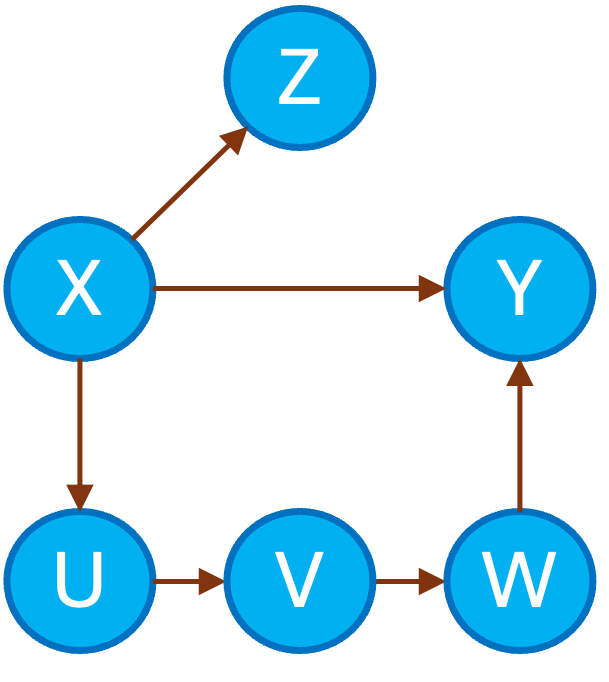} \end{minipage}} &  \begin{minipage}{.17\linewidth} \centering \includegraphics[width=\linewidth]{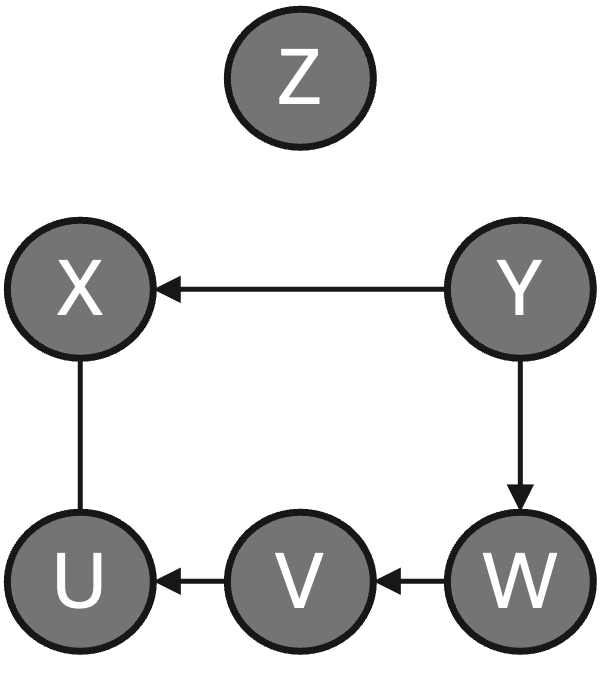} \end{minipage}  & 0.17          & 0.17         & 0.17          & 10         \\
VARLiNGAM &                   &  \begin{minipage}{.17\linewidth} \centering \includegraphics[width=\linewidth]{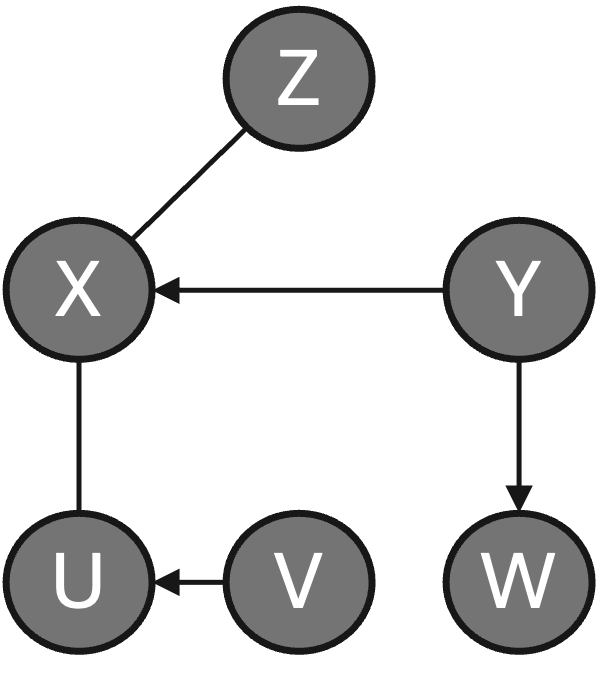} \end{minipage} & 0.29          & 0.33         & 0.31          & 9          \\
Granger   &                   &  \begin{minipage}{.17\linewidth} \centering \includegraphics[width=\linewidth]{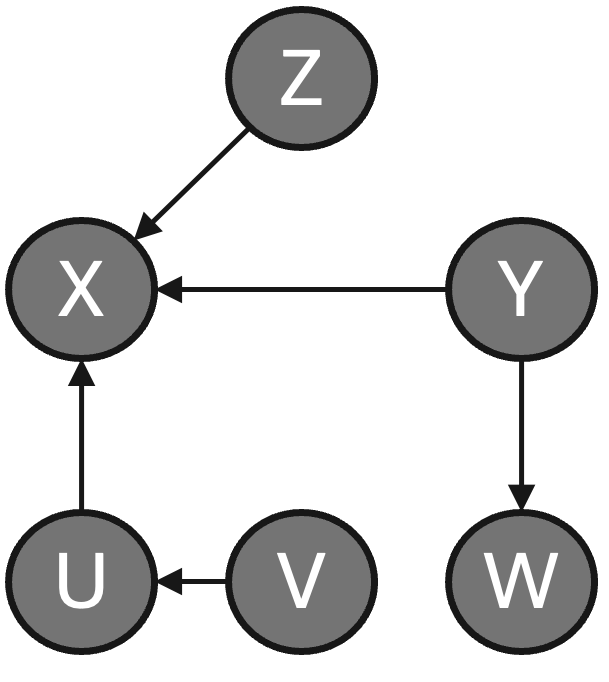} \end{minipage} & 0             & 0            & 0             & 11         \\
PCMCI     &                   &  \begin{minipage}{.17\linewidth} \centering \includegraphics[width=\linewidth]{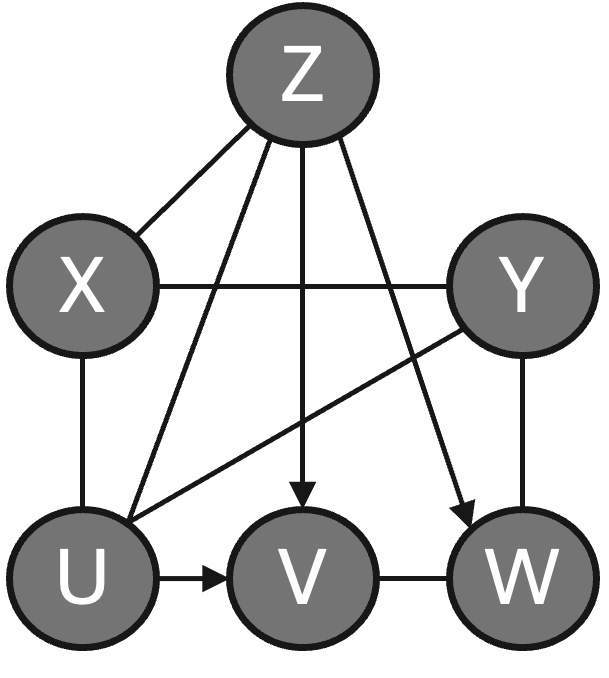} \end{minipage} & 0.35          & \textbf{1.0} & 0.52          & 11         \\
DYNOTEARS &                   &  \begin{minipage}{.17\linewidth} \centering \includegraphics[width=\linewidth]{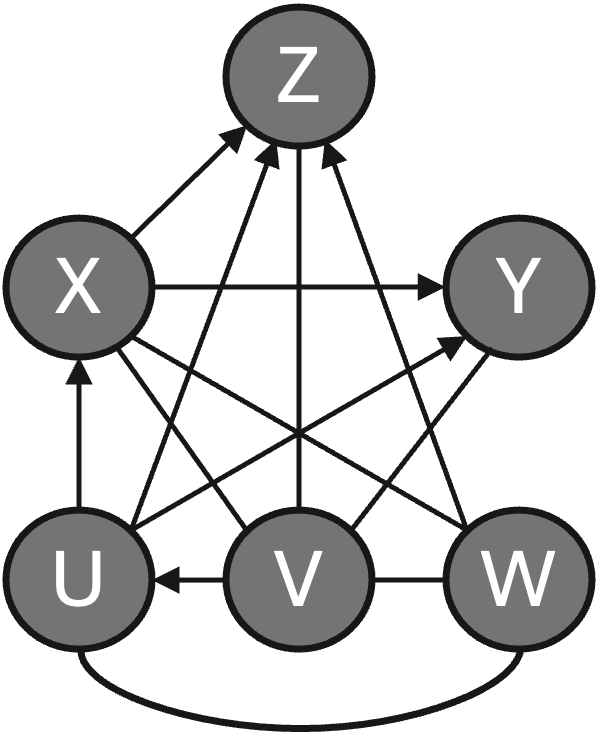} \end{minipage}  & 0.16          & 0.50         & 0.24          & 19         \\
SLARAC    &                   & \begin{minipage}{.17\linewidth} \centering \includegraphics[width=\linewidth]{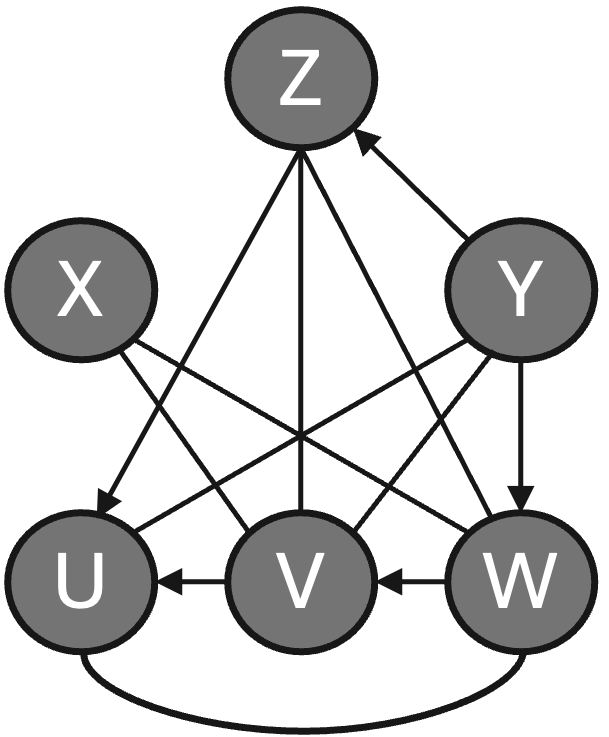} \end{minipage}   & 0             & 0            & 0             & 25         \\ \cline{1-1} \cline{3-7} 
MXMap     &                   &  \begin{minipage}{.17\linewidth} \centering \includegraphics[width=\linewidth]{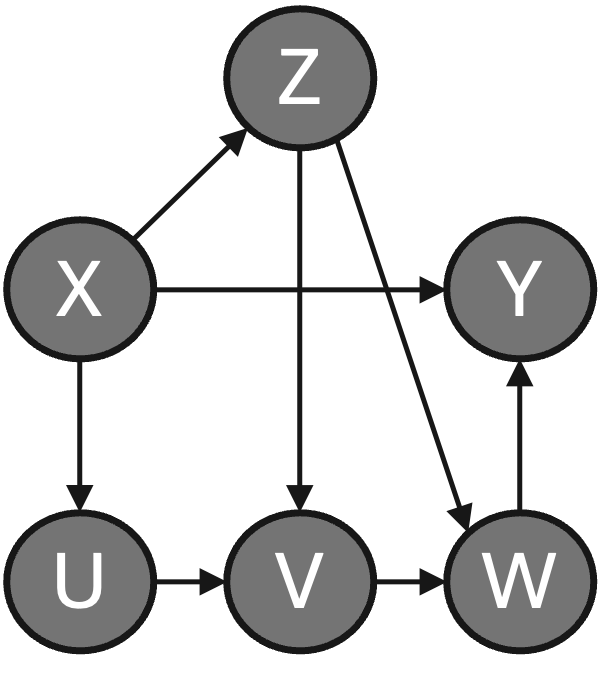} \end{minipage} & \textbf{0.75} & \textbf{1.0} & \textbf{0.86} & \textbf{2}
\end{tabular}
\caption{6V Structure Without Cycle (No Noise)}
\label{tab:6V_noN}
\end{table}

\begin{table}[htb]
\begin{tabular}{l|c|c|c|c|c|c}
Method    & Ground Truth      & Predicted & Precision     & Recall       & F1            & SHD        \\ \hline
tsFCI     & \multirow{7}{*}[-16em]{\begin{minipage}{.17\linewidth} \centering \includegraphics[width=\linewidth]{imgs/sim_new/gt/6V_gt.png} \end{minipage}} &  \begin{minipage}{.17\linewidth} \centering \includegraphics[width=\linewidth]{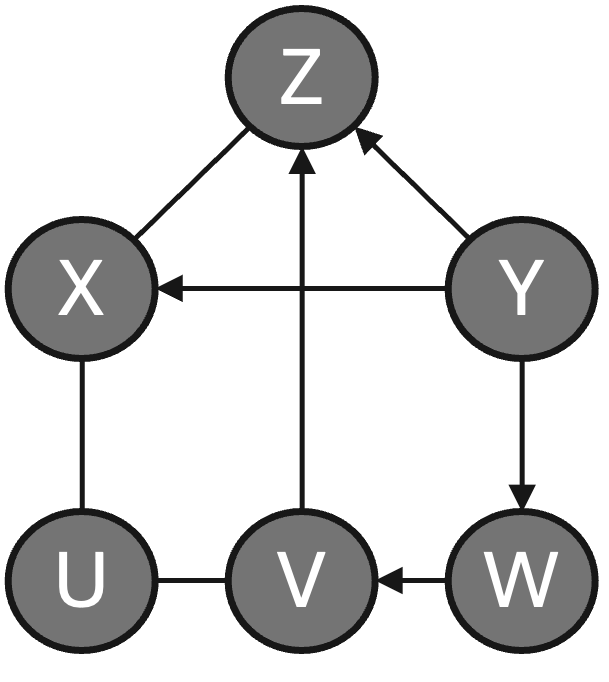} \end{minipage}   & 0.27          & 0.50         & 0.35          & 11         \\
VARLiNGAM &                   &  \begin{minipage}{.17\linewidth} \centering \includegraphics[width=\linewidth]{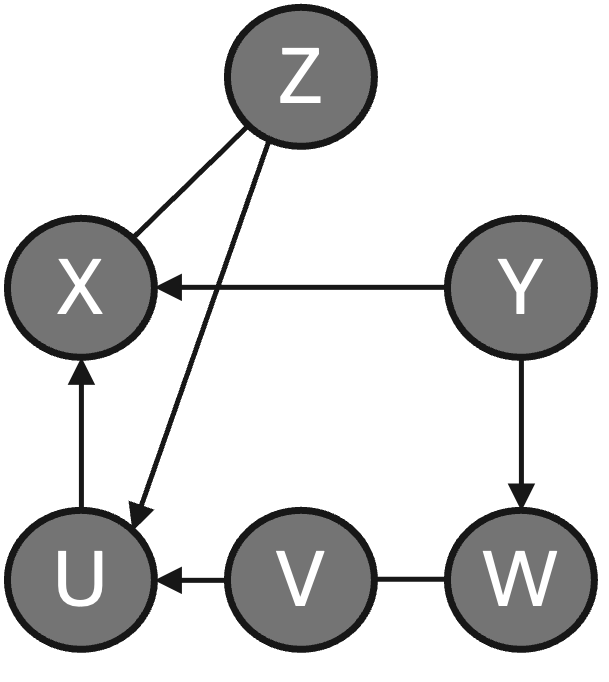} \end{minipage}    & 0.22          & 0.33         & 0.27          & 11         \\
Granger   &                   & \begin{minipage}{.17\linewidth} \centering \includegraphics[width=\linewidth]{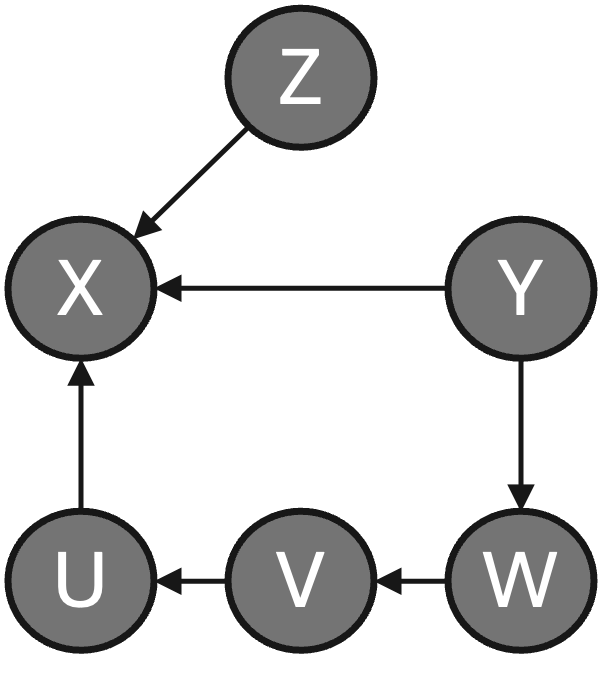} \end{minipage}   & 0             & 0            & 0             & 12         \\
PCMCI     &                   & \begin{minipage}{.17\linewidth} \centering \includegraphics[width=\linewidth]{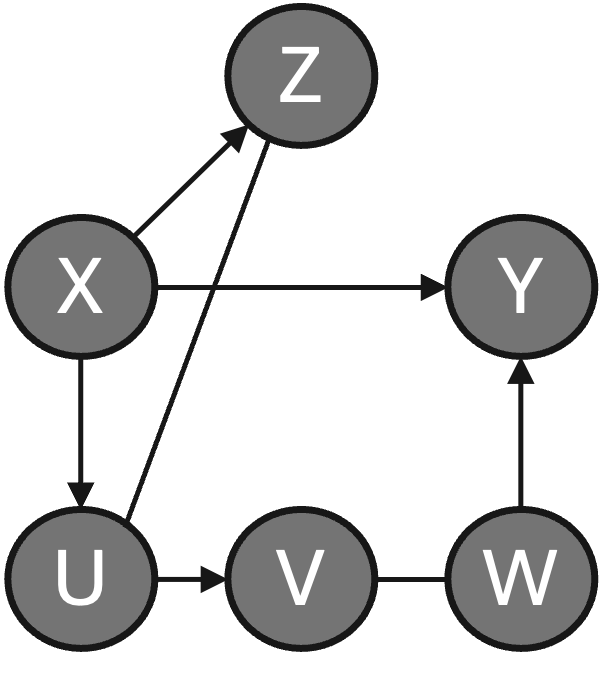} \end{minipage}   & \textbf{0.67} & \textbf{1.0} & \textbf{0.80} & \textbf{3} \\
DYNOTEARS &                   & \begin{minipage}{.17\linewidth} \centering \includegraphics[width=\linewidth]{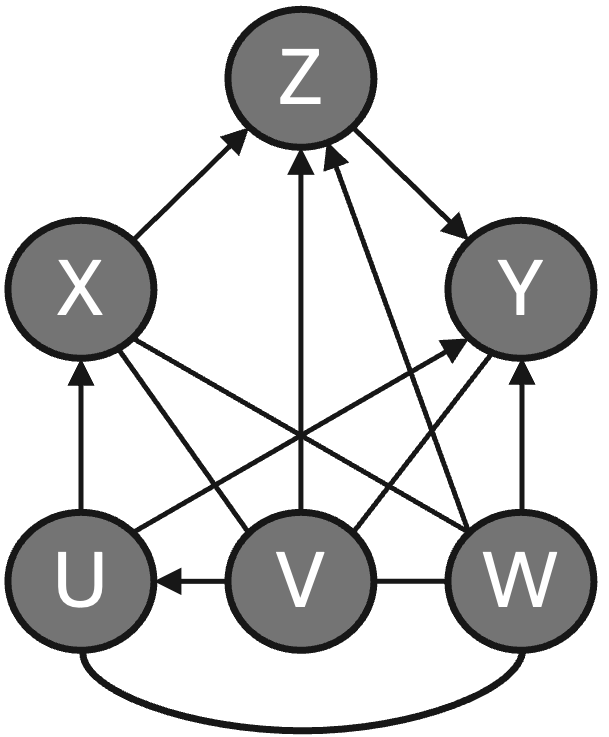} \end{minipage}   & 0.17          & 0.50         & 0.25          & 18         \\
SLARAC    &                   & \begin{minipage}{.17\linewidth} \centering \includegraphics[width=\linewidth]{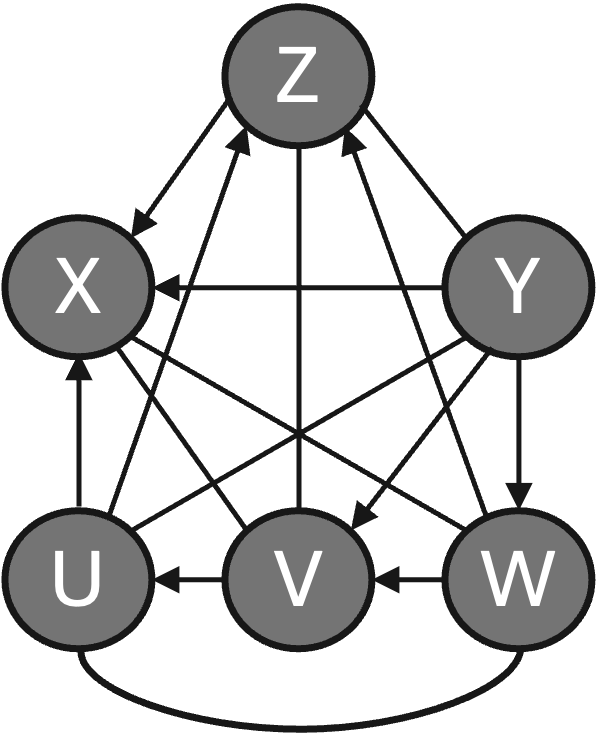} \end{minipage}     & 0             & 0            & 0             & 27         \\ \cline{1-1} \cline{3-7} 
MXMap     &                   &  \begin{minipage}{.17\linewidth} \centering \includegraphics[width=\linewidth]{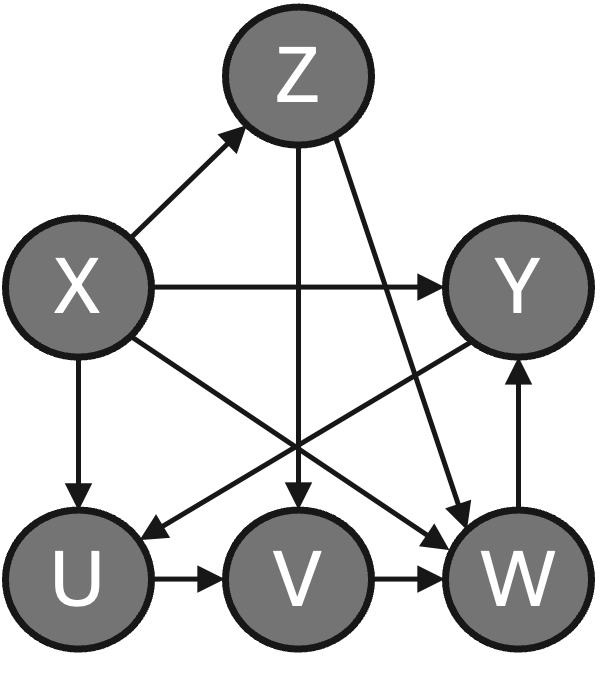} \end{minipage}   & 0.60          & \textbf{1.0} & 0.75          & 4         
\end{tabular}
\caption{6V Structure Without Cycle (Gaussian Additive Noise, Level 0.01)}
\label{tab:6V_gN}
\end{table}

\begin{table}[htb]
\begin{tabular}{l|c|c|c|c|c|c}
Method    & Ground Truth      & Predicted & Precision     & Recall       & F1            & SHD        \\ \hline
tsFCI     & \multirow{7}{*}[-9em]{\begin{minipage}{.17\linewidth} \centering \includegraphics[width=\linewidth]{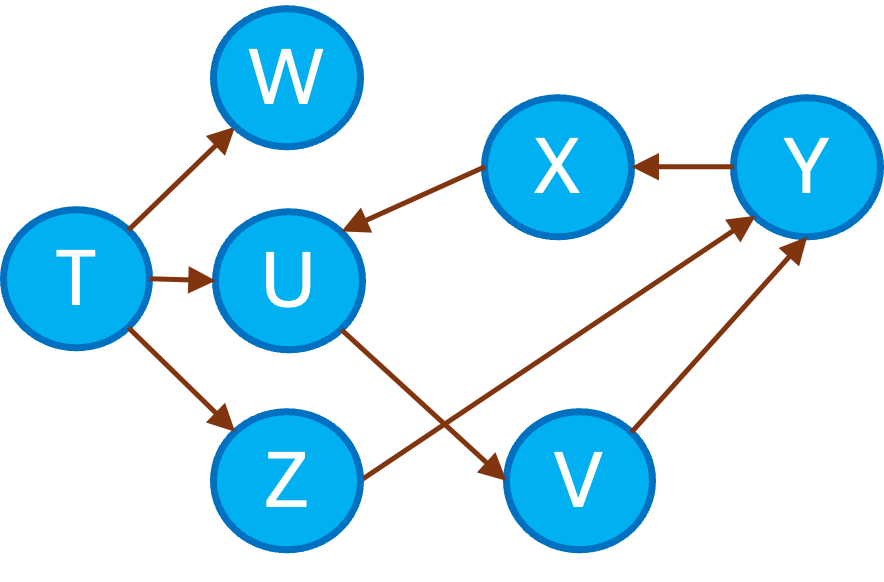} \end{minipage}} & \begin{minipage}{.17\linewidth} \centering \includegraphics[width=\linewidth]{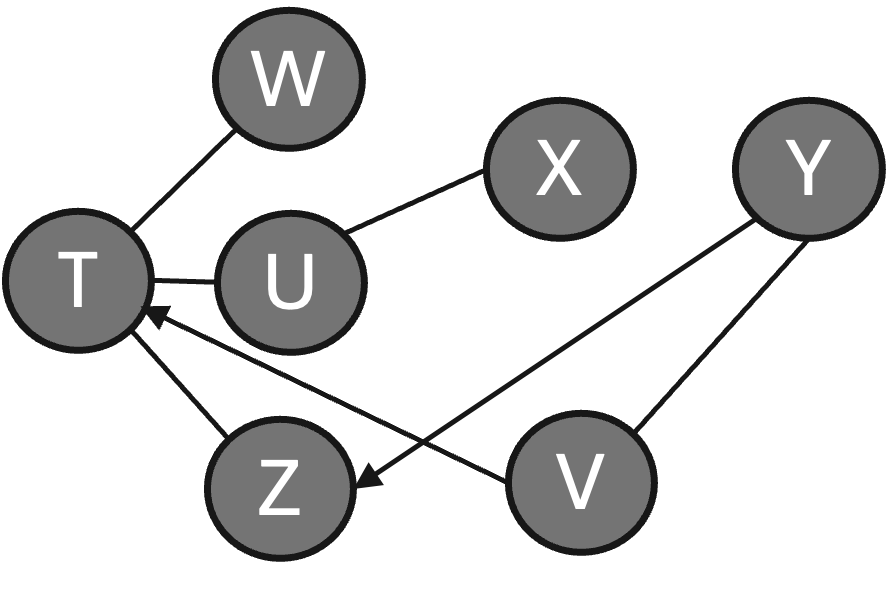} \end{minipage}   & 0.43          & 0.63         & 0.50          & 10         \\
VARLiNGAM &                   &  \begin{minipage}{.17\linewidth} \centering \includegraphics[width=\linewidth]{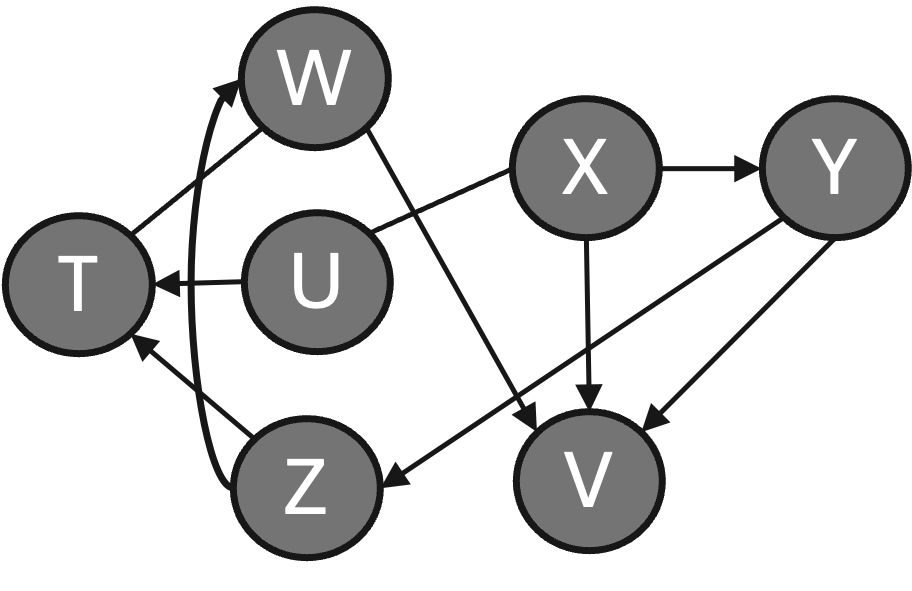} \end{minipage}   & 0.17          & 0.25         & 0.20          & 16         \\
Granger   &                   &  \begin{minipage}{.17\linewidth} \centering \includegraphics[width=\linewidth]{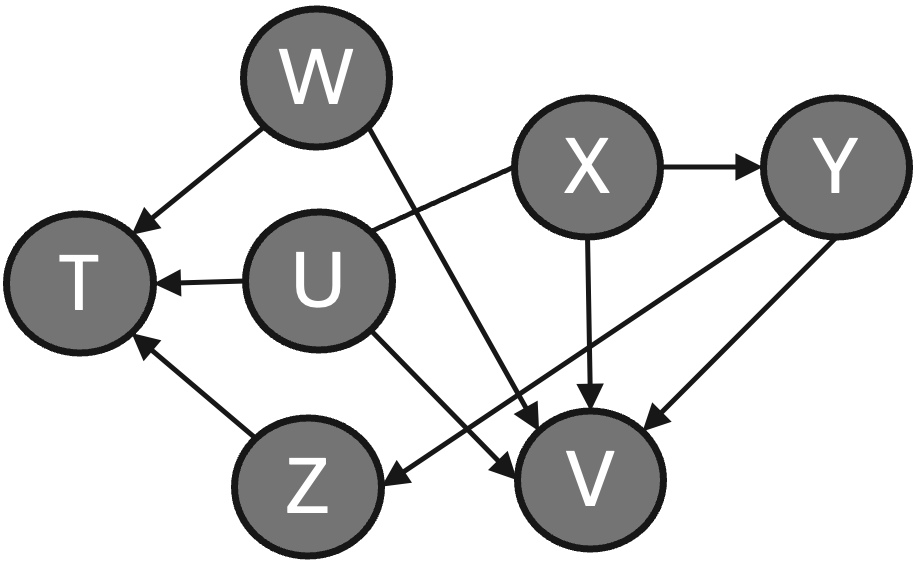} \end{minipage}   & 0.18          & 0.25         & 0.21          & 15         \\
PCMCI     &                   &  \begin{minipage}{.17\linewidth} \centering \includegraphics[width=\linewidth]{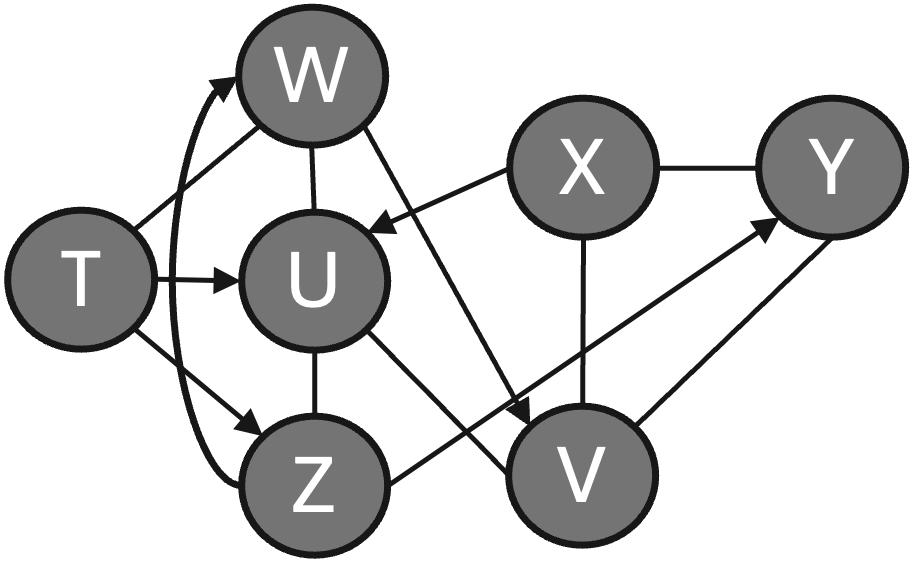} \end{minipage}   & 0.42          & \textbf{1.0} & 0.59          & 11         \\
DYNOTEARS &                   & \begin{minipage}{.17\linewidth} \centering \includegraphics[width=\linewidth]{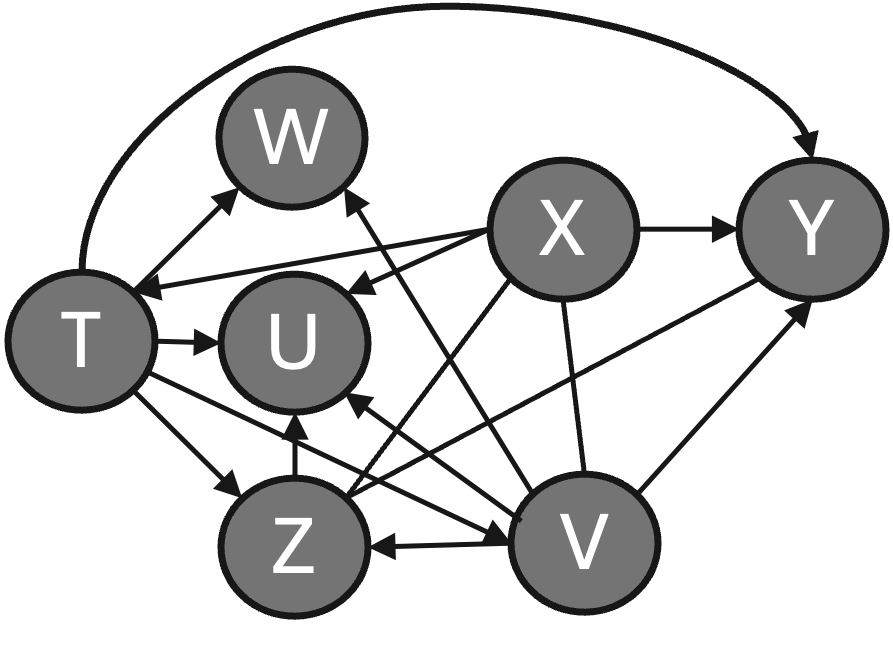} \end{minipage}   & 0.30          & 0.75         & 0.43          & 16         \\
SLARAC    &                   &  \begin{minipage}{.17\linewidth} \centering \includegraphics[width=\linewidth]{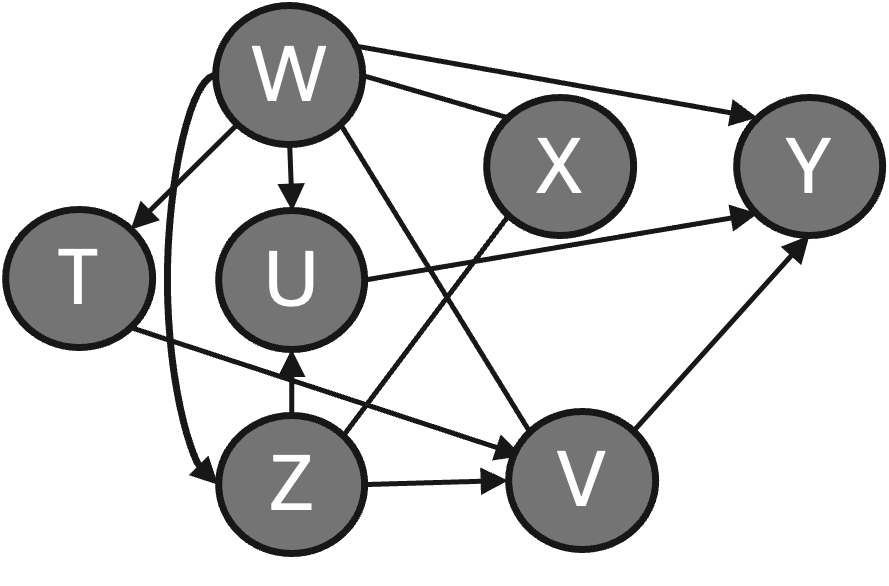} \end{minipage}   & 0.06          & 0.13         & 0.09          & 21         \\ \cline{1-1} \cline{3-7} 
MXMap     &                   &  \begin{minipage}{.17\linewidth} \centering \includegraphics[width=\linewidth]{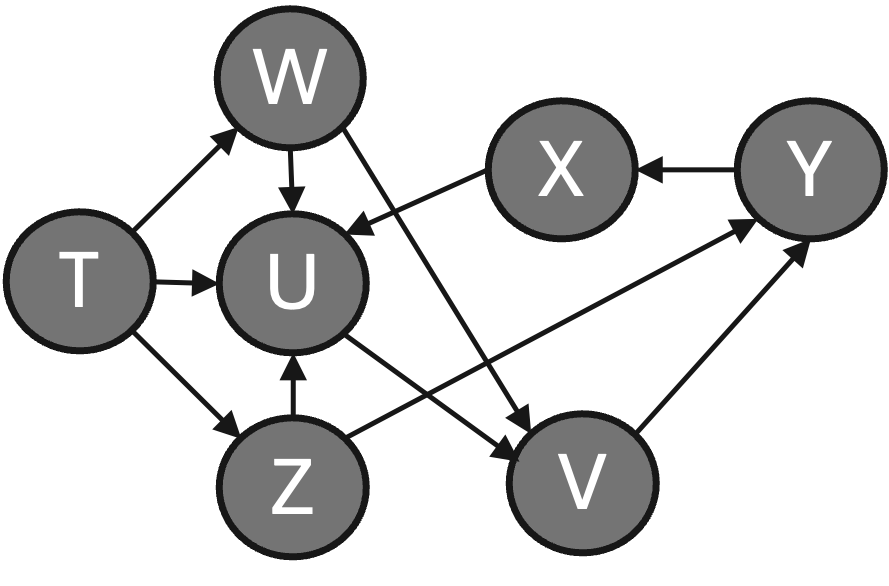} \end{minipage}   & \textbf{0.67} & \textbf{1.0} & \textbf{0.80} & \textbf{4}
\end{tabular}
\caption{7V Structure With Cycle (No Noise)}
\label{tab:7V_noN}
\end{table}

\begin{table}[htb]
\begin{tabular}{l|c|c|c|c|c|c}
Method    & Ground Truth      & Predicted & Precision     & Recall       & F1            & SHD        \\ \hline
tsFCI     & \multirow{7}{*}[-9em]{\begin{minipage}{.17\linewidth} \centering \includegraphics[width=\linewidth]{imgs/sim_new/gt/7V_gt.png} \end{minipage}} &  \begin{minipage}{.17\linewidth} \centering \includegraphics[width=\linewidth]{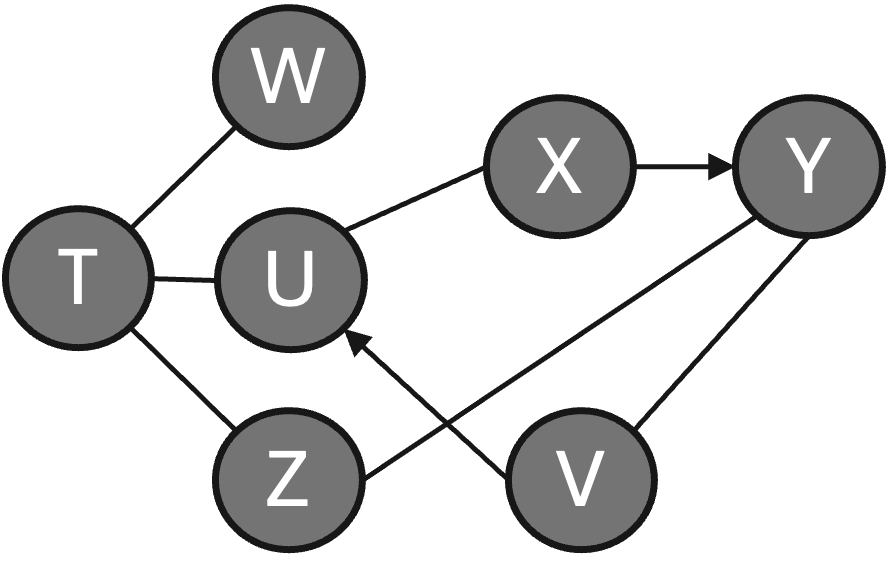} \end{minipage}  & 0.43          & 0.75         & 0.55          & 10         \\
VARLiNGAM &                   &  \begin{minipage}{.17\linewidth} \centering \includegraphics[width=\linewidth]{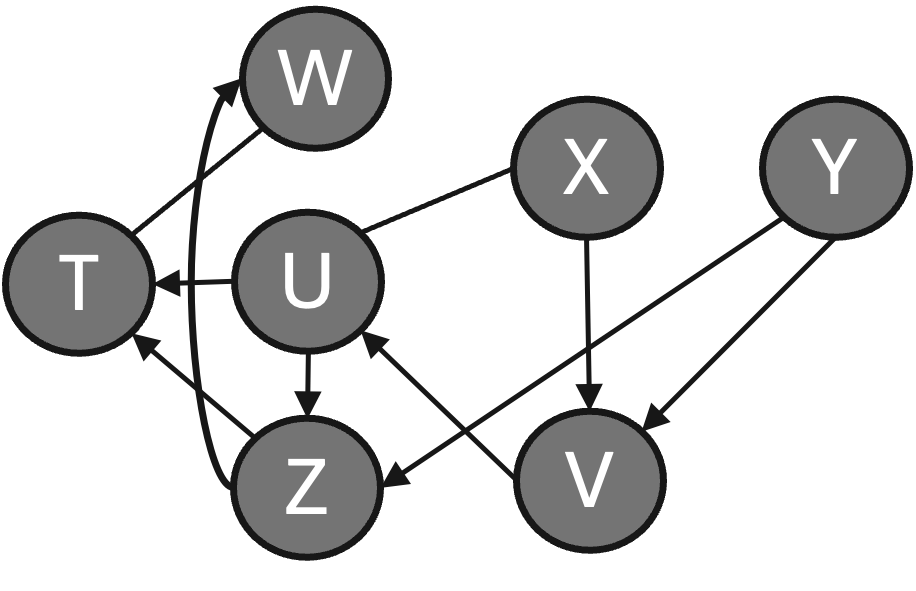} \end{minipage}   & 0.18          & 0.25         & 0.21          & 15         \\
Granger   &                   & \begin{minipage}{.17\linewidth} \centering \includegraphics[width=\linewidth]{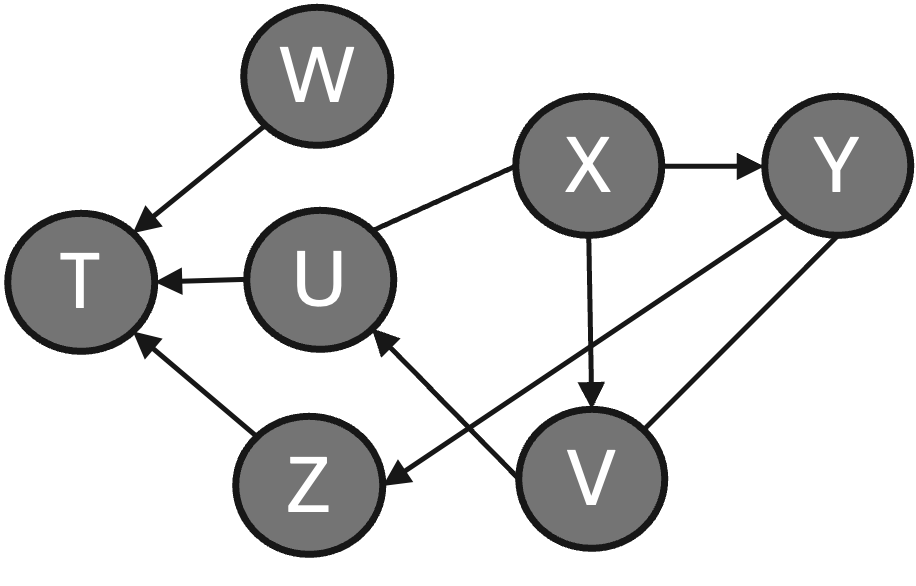} \end{minipage}  & 0.18          & 0.25         & 0.21          & 15         \\
PCMCI     &                   &  \begin{minipage}{.17\linewidth} \centering \includegraphics[width=\linewidth]{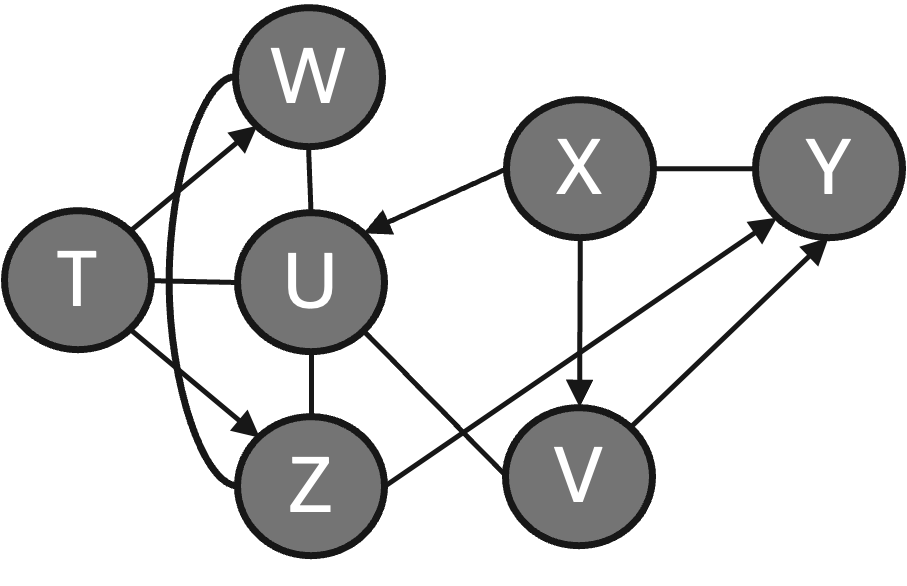} \end{minipage}   & 0.44          & \textbf{1.0} & 0.62          & 10         \\
DYNOTEARS &                   & \begin{minipage}{.17\linewidth} \centering \includegraphics[width=\linewidth]{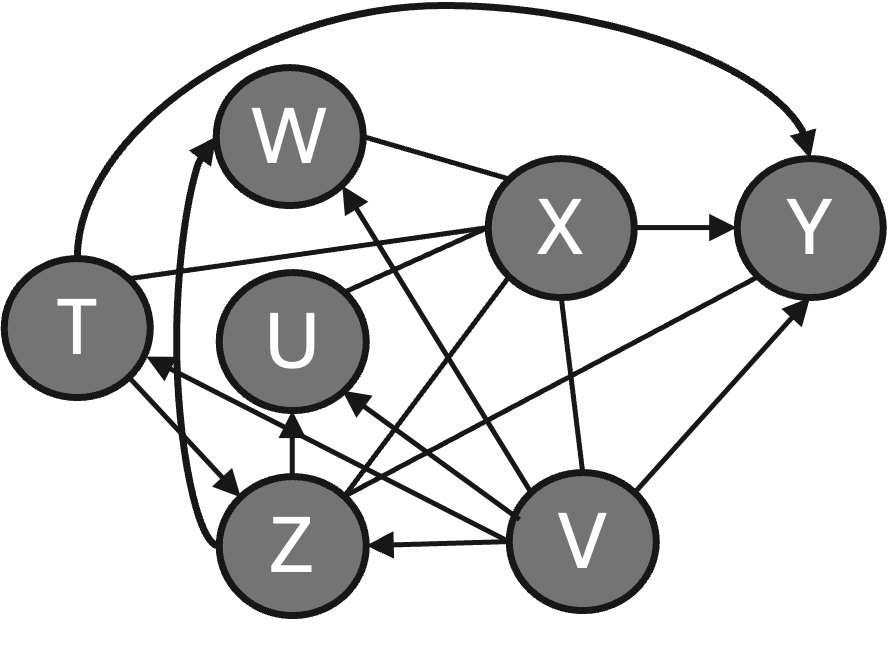} \end{minipage}  & 0.18          & 0.50         & 0.27          & 22         \\
SLARAC    &                   &  \begin{minipage}{.17\linewidth} \centering \includegraphics[width=\linewidth]{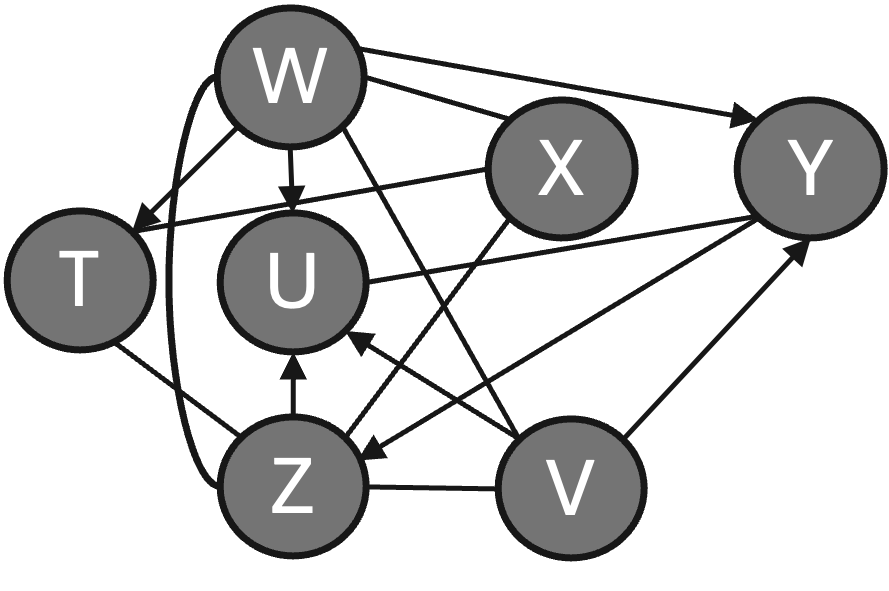} \end{minipage}  & 0.10          & 0.25         & 0.14          & 25         \\ \cline{1-1} \cline{3-7} 
MXMap     &                   &   \begin{minipage}{.17\linewidth} \centering \includegraphics[width=\linewidth]{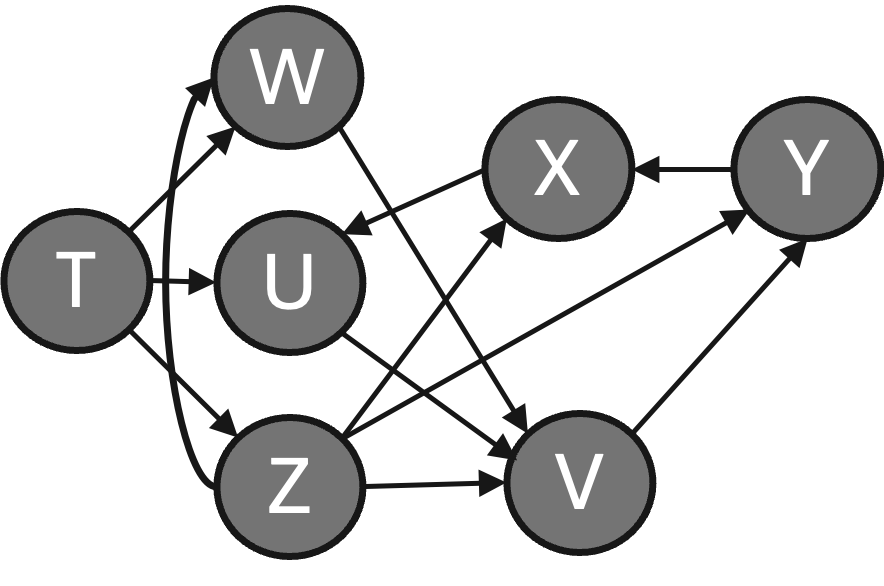} \end{minipage}    & \textbf{0.58} & 0.88         & \textbf{0.70} & \textbf{6}
\end{tabular}
\caption{7V Structure With Cycle (Gaussian Additive Noise, Level 0.01)}
\label{tab:7V_gN}
\end{table}

\end{document}